%% file: main.tex
\documentclass[10pt,letterpaper]{technicalreport}
\input{preamble}

\input{metadata}

\begin{document}
\begin{reportfrontmatter}
\maketitle

\input{sec/00_abstract}
\reportlinks
\end{reportfrontmatter}
\input{sec/01_introduction}
\input{sec/02_related_work}
\input{sec/03_published_comparisons}
\input{sec/04_benchmark_scope}
\input{sec/05_empirical_comparison}
\input{sec/06_training_progress_stability}
\input{sec/07_cost_performance}
\input{sec/08_discussion}
\input{sec/09_conclusion}
\input{sec/10_limitations}
\input{sec/11_ethics}

\FloatBarrier
\bibliography{main}

\clearpage
\appendix
\setcounter{figure}{0}
\setcounter{table}{0}
\setcounter{algorithm}{0}
\renewcommand{\thefigure}{S\arabic{figure}}
\renewcommand{\thetable}{S\arabic{table}}
\renewcommand{\thealgorithm}{S\arabic{algorithm}}
\renewcommand{\theHfigure}{appendix.\arabic{figure}}
\renewcommand{\theHtable}{appendix.\arabic{table}}
\providecommand{\theHalgorithm}{}
\renewcommand{\theHalgorithm}{appendix.\arabic{algorithm}}
\input{sec/12_appendix}

\end{document}

%% file: preamble.tex
\usepackage{latexsym}
\usepackage[T1]{fontenc}
\usepackage[utf8]{inputenc}
\usepackage{microtype}
\usepackage{inconsolata}
\usepackage{graphicx}
\usepackage{amsmath}
\usepackage{amssymb}
\usepackage{booktabs}
\usepackage{colortbl}
\usepackage{algorithm}
\usepackage{algpseudocode}

\graphicspath{{figure/}}

\input{numbers/arxiv_counts}
\input{numbers/generated_paper_numbers}


\input{numbers/generated_searchqa_numbers}

\input{numbers/generated_scienceworld_numbers}
\input{numbers/generated_full_compute_numbers}
\input{numbers/generated_full_training_numbers}

%% file: numbers/arxiv_counts.tex
\newcommand{\SurveyPaperCount}{47}
\newcommand{\SurveyDayCount}{90}

\newcommand{\SurveyDailyRatePrecise}{0.5222}

\newcommand{\SurveyPrivilegedCount}{32}

%% file: numbers/generated_paper_numbers.tex
\newcommand{\PaperAlfInstances}{140}
\newcommand{\PaperWebInstances}{128}

%% file: numbers/generated_searchqa_numbers.tex
\newcommand{\PaperCompletedRuns}{99}
\newcommand{\PaperCompletedSettings}{9}
\newcommand{\PaperCompletedValidations}{2,079}
\newcommand{\PaperCompletedEvalSweeps}{864}

%% file: numbers/generated_scienceworld_numbers.tex
\renewcommand{\PaperCompletedRuns}{132}
\renewcommand{\PaperCompletedSettings}{12}
\renewcommand{\PaperCompletedValidations}{2,772}
\renewcommand{\PaperCompletedEvalSweeps}{1,152}
\newcommand{\PaperTwelveGRPOMean}{49.68}

\newcommand{\PaperTwelveGRPOOPSDMean}{43.05}

\newcommand{\PaperTwelveOCSDMean}{49.50}

\newcommand{\PaperTwelveAgentOPSDMean}{51.34}
\newcommand{\PaperTwelveAgentOPSDGain}{1.66}

\newcommand{\PaperTwelveICSDMean}{44.75}

\newcommand{\PaperTwelvePCSDMean}{54.26}
\newcommand{\PaperTwelvePCSDGain}{4.58}

%% file: numbers/generated_full_compute_numbers.tex
\newcommand{\PaperAllRecordedGPUHours}{12,124.4}
\newcommand{\PaperAllTrainingGPUHours}{10,580.7}
\newcommand{\PaperAllValidationGPUHours}{1,515.3}
\newcommand{\PaperAllCheckpointGPUHours}{28.4}
\newcommand{\PaperCompletedTrainSteps}{13,200}

%% file: numbers/generated_full_training_numbers.tex
\newcommand{\PaperAllLowValidationRuns}{16}
\newcommand{\PaperAllLowFinalRuns}{15}
\newcommand{\PaperAllDropRuns}{9}
\newcommand{\PaperAllNonfiniteRuns}{21}
\newcommand{\PaperAllLowAndNonfiniteRuns}{3}
\newcommand{\PaperAllLowWithoutNonfiniteRuns}{13}
\newcommand{\PaperAllNonfiniteWithoutLowRuns}{18}
\newcommand{\PaperAllFiniteDropRuns}{9}
\newcommand{\PaperAllGRPOAUC}{36.89}

\newcommand{\PaperAllPCSDAUC}{38.45}

\newcommand{\PaperAllOCSDAUC}{39.88}

\newcommand{\PaperAllAgentAUC}{34.41}

%% file: metadata.tex
\title{\textcolor{metablue}{More Methods, Better Agents?}\\
A Reality Check on Self-Distillation in Agentic Reinforcement Learning}

\author{%
Linhao~Zhong\textsuperscript{1,2,*},
Bozhen~Fang\textsuperscript{1,*},
Xuehao~Wang\textsuperscript{1},
Duochao~Shi\textsuperscript{1},
Zekai~Luo\textsuperscript{1},
Guangming~Yao\textsuperscript{2},
Yi~Yuan\textsuperscript{2},
Hao~Chen\textsuperscript{1},
Jingdong~Chen\textsuperscript{2},
Chunhua~Shen\textsuperscript{1,2}}
\affiliations{%
\textsuperscript{1}Zhejiang University, State Lab of CAD \& CG\quad
\textsuperscript{2}Ant Group}
\contributions{\textsuperscript{*}Equal contribution.}
\date{October 2026}
\projectpage{https://aim-uofa.github.io/Self-Distillation-AgentRL}
\githubrepo{https://github.com/aim-uofa/Self-Distillation-AgentRL}
\hypersetup{pdftitle={More Methods, Better Agents? A Reality Check on Self-Distillation in Agentic Reinforcement Learning},pdfauthor={Linhao Zhong; Bozhen Fang; Xuehao Wang; Duochao Shi; Zekai Luo; Guangming Yao; Yi Yuan; Hao Chen; Jingdong Chen; Chunhua Shen}}

%% file: sec/00_abstract.tex
\begin{abstract}
Self-distillation in agentic reinforcement learning has produced a growing number of methods.
However, existing evaluations often reuse published baseline scores under differing experimental conditions, provide few direct comparisons among contemporaneous methods, and give limited attention to training stability and computational cost.
These limitations obscure the source of reported gains and the practical strengths and weaknesses of competing methods.
Costly agent training further motivates a common basis for comparison.
To address these gaps, we compare ten representative methods against GRPO on ALFWorld, ScienceWorld, Search-QA, and WebShop across three model sizes.
Our study comprises \PaperCompletedRuns{} training runs totaling \PaperAllRecordedGPUHours{} recorded H800 GPU-hours, with repeated evaluations over eight rollout seeds for the pretrained backbones and all eleven trained methods.
Beyond final task performance, we examine training progress, stability, and measured computational cost.
Our evaluation yields three key findings.
First, gains over GRPO remain limited in scope, with no method consistently outperforming it across settings.
Second, strong final performance can mask unstable training.
Third, additional training cost does not consistently translate into better performance.
We establish this benchmark to support fair, comprehensive comparison and guide future research toward reliable and cost-effective improvements in agent training.
\end{abstract}

%% file: sec/01_introduction.tex
\section{Introduction}
\label{sec:introduction}

\input{figure/overview}

Large language model (LLM) agents interact with environments through sequences of actions and observations~\citep{shridhar2021alfworld,yao2022webshop}.
Reinforcement learning (RL) with sparse outcome rewards trains these agents from task-level feedback, but provides limited guidance for individual decisions within a trajectory.
On-policy self-distillation (OPSD) provides denser supervision for student-generated trajectories using a teacher branch conditioned on training-only privileged information~\citep{lu2026sdar,wang2026skillsd}.
Recent work integrates this teacher guidance into RL through auxiliary losses, advantage modulation, or reward shaping~\citep{lu2026sdar,yang2026ocsd,zhang2026adrs,wang2026agentopsd}, with related preprints appearing at an average rate of roughly one every two days.\footnote{Survey criteria and publication inventory appear in Appendices~\ref{app:survey-method} and~\ref{app:survey-statistics}.}

However, current evaluations of self-distillation in agentic reinforcement learning provide an incomplete basis for judging progress.
Individual gains over established baselines do not establish which contemporaneous methods are stronger, yet systematic comparisons among these methods remain scarce.
Moreover, most comparisons reuse baseline scores from earlier papers rather than reproducing the baselines under a common protocol, allowing differences in training and evaluation conditions to be mistaken for methodological progress.
Many studies also emphasize final scores, leaving training instability and the computational cost of improvement less systematically examined.
As Figure~\ref{fig:overview} illustrates, the proliferation of methods has outpaced the comparable evidence needed to decide which ones are worth adopting.
Closing this gap requires substantial resources: reproducing agent training is costly, making a shared, systematic comparison difficult to assemble within each new method paper.
We therefore argue that \emph{independent evaluation should take priority over proposing another method}: the immediate need is to determine what is worth building on.

To address these limitations, we present an independent benchmark for self-distillation in agentic reinforcement learning.
We compare ten representative methods against GRPO~\citep{shao2024deepseekmath} across four benchmarks---ALFWorld~\citep{shridhar2021alfworld}, ScienceWorld~\citep{wang-etal-2022-scienceworld}, Search-QA, and WebShop~\citep{yao2022webshop}---and three scales (2B, 4B, and 9B).
This comparison spans \PaperCompletedSettings{} model--task settings, \PaperCompletedRuns{} training runs, \PaperCompletedValidations{} training-time validation sweeps, and \PaperCompletedEvalSweeps{} evaluation sweeps over eight rollout seeds for pretrained and trained models.
The training runs total \emph{\PaperAllRecordedGPUHours{} recorded H800 GPU-hours} across all four benchmarks.
Shared environments and evaluation procedures provide a common basis for comparison while retaining method-specific training settings informed by the source papers and implementations.
We examine not only whether gains hold across settings, but also whether they are sustained during training and justify the additional computation.
Our analysis yields three key observations:

\textbf{(i)} Gains over GRPO remain limited in scope, with no method consistently outperforming it across settings.

\textbf{(ii)} Strong final performance does not imply stable training.

\textbf{(iii)} Additional training cost does not consistently translate into better agent performance.

%% file: figure/overview.tex
\begin{figure*}[!t]
\centering
\includegraphics[width=\textwidth]{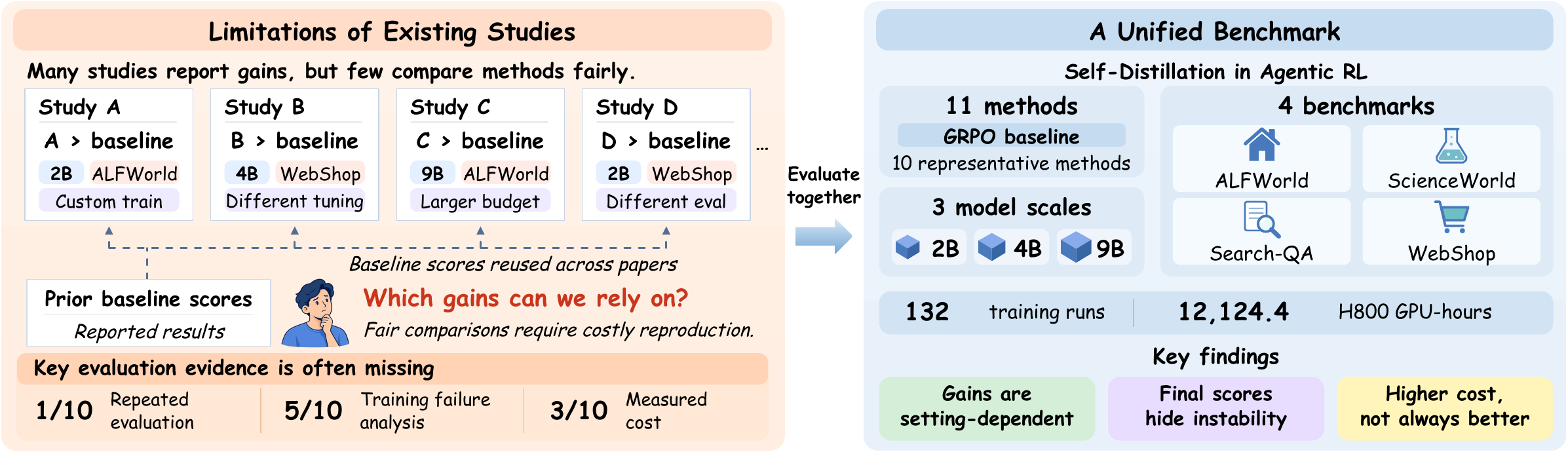}
\caption{Motivation and overview of our benchmark for self-distillation in agentic reinforcement learning. Differences in experimental conditions, reused baseline scores, and incomplete evaluation motivate a common comparison of ten representative methods against GRPO across four benchmarks and three model sizes. We examine the breadth of gains, training stability, and computational cost.}
\label{fig:overview}
\end{figure*}

%% file: sec/02_related_work.tex
\section{Related Work}
\label{sec:related-work}

\input{table/reporting_audit}

\noindent \textbf{Interactive agents and learning signals.}
Language agents solve tasks through sequential actions and observations~\citep{shridhar2021alfworld,yao2022webshop,xi2025agentgym}.
Inference-time methods interleave reasoning and action~\citep{yao2023react} or refine decisions and outputs through verbal feedback~\citep{shinn2023reflexion,madaan2023selfrefine}, without parameter updates.
Training methods learn from filtered tool-use annotations~\citep{schick2023toolformer} and curated interaction trajectories~\citep{qin2024toolllm,chen2023fireact,zeng2024agenttuning}.
Reward-based optimization learns from sampled behavior~\citep{shao2024deepseekmath}, while online curricula support continued policy improvement~\citep{qi2025webrl}.
Multi-turn RL can nevertheless exhibit unstable updates and degenerate behavior, motivating careful reward design and rollout sampling~\citep{wang2025ragen}.
Our benchmark focuses on capabilities acquired through parameter updates, rather than improvements obtained by modifying an agent's inference procedure.

\noindent \textbf{Distillation and privileged supervision.}
Distillation transfers teacher predictions, generated sequences, or policy behavior to a student~\citep{hinton2015distilling,kim2016seqkd,rusu2016policy}.
Training-only privileged information~\citep{lopezpaz2016privileged} and reasoning rationales~\citep{hsieh2023stepbystep} enrich supervision beyond target labels.
Privileged supervision therefore seeks to improve the student's policy without requiring access to the teacher's additional information at deployment.
Supervising learner-visited states or student-generated sequences helps reduce distribution mismatch~\citep{ross2011dagger,lin2020imitkd,agarwal2024gkd}.
MiniLLM and DistiLLM study divergence objectives and sampling strategies~\citep{gu2024minillm,ko2024distillm}, while iterative self-training learns from selected, self-generated reasoning or interaction traces~\citep{zelikman2022star,aksitov2023restreact}.
Supervision content, sampling distribution, and optimization objective offer complementary perspectives for comparing distillation-based agent training.

\noindent \textbf{Self-distillation in agentic reinforcement learning.}
Agentic self-distillation derives teacher supervision from privileged context available only during training.
The evaluated self-distillation methods use four mechanisms.
Skill-SD, GRPO+OPSD, SDAR, ICSD, and PCSD add auxiliary objectives~\citep{wang2026skillsd,lu2026sdar,lan2026icsd,lv2026pcsd}, whereas RLSD, OCSD, and AgentOPSD modulate policy advantages through teacher--student comparisons~\citep{yang2026rlsd,yang2026ocsd,wang2026agentopsd}.
ADRS shapes rewards and advantages~\citep{zhang2026adrs}, while OVCSD adds environment-verified teacher continuations~\citep{xia2026ovcsd}.
Recent analyses identify supervision inconsistency and privileged-information bias as potential failure modes~\citep{yang2026feedbackfailure,harne2026privilegedbias}.
Our benchmark compares these mechanisms' performance, training stability, and computational cost under a common evaluation protocol.
Appendices~\ref{app:alg-auxiliary}, \ref{app:alg-reweighting}, \ref{app:alg-reward-shaping}, and~\ref{app:alg-continuation} detail the four mechanisms, respectively.

%% file: table/reporting_audit.tex
\begin{table*}[t]
\centering
\small
\setlength{\tabcolsep}{5pt}
\begin{tabular*}{\textwidth}{@{\extracolsep{\fill}}lccc@{}}
\toprule
Method (source) & Repeated evaluation & Training failure analysis & Measured cost \\
\midrule
GRPO+OPSD \citep{lu2026sdar} & $\times$ & $\checkmark$ & $\times$ \\
SDAR \citep{lu2026sdar} & $\times$ & $\checkmark$ & $\times$ \\
Skill-SD \citep{wang2026skillsd} & $\times$ & $\checkmark$ & $\times$ \\
RLSD \citep{yang2026rlsd} & $\times$ & $\checkmark$ & $\times$ \\
OCSD \citep{yang2026ocsd} & $\checkmark$ & $\times$ & $\checkmark$ \\
OVCSD \citep{xia2026ovcsd} & $\times$ & $\times$ & $\checkmark$ \\
AgentOPSD \citep{wang2026agentopsd} & $\times$ & $\times$ & $\times$ \\
ADRS \citep{zhang2026adrs} & $\times$ & $\checkmark$ & $\times$ \\
ICSD \citep{lan2026icsd} & $\times$ & $\times$ & $\checkmark$ \\
PCSD \citep{lv2026pcsd} & $\times$ & $\times$ & $\times$ \\
\midrule
\textbf{Methods with reported evidence} & \textbf{1 / 10} & \textbf{5 / 10} & \textbf{3 / 10} \\
\bottomrule
\end{tabular*}
\caption{Coverage of key evaluation dimensions in prior work.}
\label{tab:reporting-audit}
\end{table*}

%% file: sec/03_published_comparisons.tex
\section{The Limits of Published Comparisons}
\label{sec:reporting-audit}

Table~\ref{tab:reporting-audit} summarizes how the source papers of the ten compared methods address repeated evaluation, training-failure analysis, and measured computational cost.
Gaps in this coverage limit assessments of method reliability and practical value beyond reported performance gains.

\noindent \textbf{Repeated evaluation.}
Performance differences must be interpreted relative to variation across evaluations.
Yet repeated evaluation is rarely reported in the surveyed sources, making it difficult to distinguish consistent improvements from favorable sampling outcomes.
Reporting variability alongside mean performance is therefore essential to judging the strength of a claimed advantage.

\noindent \textbf{Training failure analysis.}
A strong final score does not establish that training is stable.
Although some studies investigate collapse or performance degradation, such analyses are not consistently available across competing methods.
Examining training trajectories is necessary to reveal regressions and failure risks that final scores can conceal.

\noindent \textbf{Measured cost.}
Reporting computational cost alongside performance is essential for assessing the efficiency and practical feasibility of competing methods.
Measured cost remains sparsely reported, limiting comparisons of the resources required to obtain competing methods' gains.
Assessing performance together with training cost is necessary to determine whether an improvement offers practical value.

None of the surveyed source papers covers all three dimensions.
Our benchmark brings them into a common comparison so that method selection rests on the consistency of measured gains, training stability, and computational cost rather than a favorable final score alone.

%% file: sec/04_benchmark_scope.tex
\section{Benchmark Design}
\label{sec:benchmark-scope}
\label{sec:protocol}

\input{table/generated_main_results}
\input{table/generated_main_results_search_webshop}

Our benchmark examines the value of self-distillation in multi-turn text-agent training, using outcome-based GRPO as the reference.
It combines common model--task settings and evaluation procedures with method-specific objectives and hyperparameters informed by the source papers and released implementations.

\noindent \textbf{Tasks and model backbones.}
ALFWorld~\citep{shridhar2021alfworld} requires agents to complete household tasks through textual actions and observations.
We evaluate \PaperAlfInstances{} games, reporting overall and task-specific success rates.
ScienceWorld~\citep{wang-etal-2022-scienceworld} requires agents to conduct experiments and solve scientific tasks in a text-based environment.
We evaluate 200 tasks across Measure, Test-Cond., Find, Chem-Mix, and Lifespan, reporting task score.
WebShop~\citep{yao2022webshop} requires agents to find and select products that satisfy a shopping goal.
We evaluate a fixed set of \PaperWebInstances{} goals, reporting task success and a graded score for partial goal satisfaction.
Search-QA requires retrieval-assisted answers to questions from seven sources.
We use a fixed, seed-0 subset: 500 questions each from NQ, TriviaQA, PopQA, HotpotQA, 2WikiMultiHopQA, and MuSiQue, plus all 125 Bamboogle questions.
We report exact-match accuracy after answer normalization; Overall gives equal weight to all 3,125 questions rather than to sources or evaluation batches.
Combining these four benchmarks with Qwen3.5-2B, 4B, and 9B yields twelve model--task settings.

\noindent \textbf{Methods.}
GRPO~\citep{shao2024deepseekmath} is the outcome-only reference.
GRPO+OPSD and SDAR~\citep{lu2026sdar}, together with Skill-SD~\citep{wang2026skillsd}, ICSD~\citep{lan2026icsd}, and PCSD~\citep{lv2026pcsd}, add auxiliary objectives.
RLSD~\citep{yang2026rlsd}, OCSD~\citep{yang2026ocsd}, AgentOPSD~\citep{wang2026agentopsd}, and ADRS~\citep{zhang2026adrs} use privileged signals to shape credit or rewards; OVCSD~\citep{xia2026ovcsd} adds verified teacher continuations.
All methods are evaluated without privileged teacher context.
For Skill-SD and RLSD, we use the skill-bank adaptations; RLSD was originally proposed for verifiable reasoning.
GRPO+OPSD follows SDAR's constant-gate implementation, whose auxiliary gradient is independent of teacher scores.
Appendices~\ref{app:method-scope}, \ref{app:teacher-contexts}, and~\ref{app:implemented-algorithms} detail method adaptations, teacher contexts, and algorithms, respectively.

\noindent \textbf{Training and evaluation protocol.}
We train each method in every model--task setting for 100 global steps, yielding \PaperCompletedRuns{} runs, with validation every five steps and method-specific hyperparameters guided by the source papers and released implementations.
Using the same fixed task sets as validation, we evaluate all eleven trained methods and the pretrained backbones at temperature $T=0.4$ over eight rollout seeds, reporting the mean and sample standard deviation.
Final-performance averages assign equal weight to all twelve settings, using each benchmark's reported metric; computational-cost comparisons cover the same settings.
Across all four benchmarks, we assess training progress and stability through validation-curve AUC, peak-to-final declines, and low-performing outcomes, with equal weight on the twelve settings in aggregate comparisons.
Appendices~\ref{app:training-configuration}, \ref{app:numerical-settings}, and~\ref{app:alg-shared} specify hyperparameters, numerical settings, and the shared training procedure.
Appendices~\ref{app:evaluation-settings} and~\ref{app:reproducibility} detail the evaluation protocol and experimental records, respectively.

%% file: table/generated_main_results.tex
\begin{table*}[!t]
\begingroup
\input{styles/main_results_style}
\centering
\footnotesize
\setlength{\tabcolsep}{1.8pt}
\renewcommand{\arraystretch}{1.08}
\resizebox{\textwidth}{!}{%
\arrayrulecolor{MainTableInk}
\begin{tabular}{lcccccc>{\columncolor{MainTableFocus}}cccccc>{\columncolor{MainTableFocus}}c}
\toprule
\rowcolor{MainTableInk}
 & \multicolumn{7}{c}{\color{white}\bfseries ALFWorld: success rate (\%)} & \multicolumn{6}{c}{\color{white}\bfseries ScienceWorld: task score} \\
\rowcolor{MainTableHeader}
\textbf{Method} & Pick & Look & Clean & Heat & Cool & Pick2 & \textbf{Overall} & Measure & Test-Cond. & Find & Chem-Mix & Lifespan & \textbf{Overall} \\
\midrule
\rowcolor{MainTableBand}
\multicolumn{14}{l}{\MainTableModel{Qwen3.5-2B}} \\
\rowcolor{MainTableHeader}
\textit{Pretrained} & \MainTableValue{16.43}{2.53} & \MainTableValue{23.08}{5.81} & \MainTableValue{0.93}{1.71} & \MainTableValue{1.56}{2.89} & \MainTableValue{1.50}{2.98} & \MainTableValue{2.60}{3.10} & \MainTableValue{7.32}{1.19} & \MainTableValue{-10.32}{2.82} & \MainTableValue{2.94}{0.75} & \MainTableValue{-7.26}{3.44} & \MainTableValue{1.50}{7.07} & \MainTableValue{3.33}{0.00} & \MainTableValue{-2.83}{1.31} \\
\rowcolor{MainTableBaseline}
\textbf{GRPO} & \MainTableValue{36.79}{3.22} & \MainTableValue{\underline{64.42}}{7.05} & \MainTableValue{0.00}{0.00} & \MainTableValue{13.28}{2.21} & \MainTableValue{33.50}{5.63} & \MainTableValue{33.85}{3.48} & \MainTableValue{28.48}{1.23} & \MainTableValue{0.42}{0.00} & \MainTableValue{0.60}{0.00} & \MainTableValue{0.55}{0.00} & \MainTableValue{4.00}{0.00} & \MainTableValue{3.33}{0.00} & \MainTableValue{0.83}{0.00} \\
GRPO+OPSD & \MainTableValue{14.64}{4.69} & \MainTableValue{31.73}{8.66} & \MainTableValue{27.78}{4.43} & \MainTableValue{37.50}{7.47} & \MainTableValue{33.00}{6.68} & \MainTableValue{0.52}{1.47} & \MainTableValue{22.23}{1.73} & \MainTableValue{0.42}{0.00} & \MainTableValue{0.60}{0.00} & \MainTableValue{0.55}{0.00} & \MainTableValue{4.00}{0.00} & \MainTableValue{3.33}{0.00} & \MainTableValue{0.83}{0.00} \\
SDAR & \MainTableValue{32.86}{1.53} & \MainTableValue{15.38}{0.00} & \MainTableValue{0.46}{1.31} & \MainTableValue{0.00}{0.00} & \MainTableValue{17.50}{5.63} & \MainTableValue{9.38}{4.31} & \MainTableValue{14.46}{1.13} & \MainTableValue{0.42}{0.00} & \MainTableValue{0.60}{0.00} & \MainTableValue{0.55}{0.00} & \MainTableValue{4.00}{0.00} & \MainTableValue{3.33}{0.00} & \MainTableValue{0.83}{0.00} \\
Skill-SD & \MainTableValue{69.29}{5.01} & \MainTableValue{59.62}{3.56} & \MainTableValue{\underline{85.65}}{3.09} & \MainTableValue{\MainTableBest{93.75}}{0.00} & \MainTableValue{\underline{71.50}}{4.50} & \MainTableValue{\MainTableBest{66.67}}{3.15} & \cellcolor{MainTableBestFill}\MainTableValue{\MainTableBest{74.29}}{2.44} & \MainTableValue{0.42}{0.00} & \MainTableValue{0.60}{0.00} & \MainTableValue{0.55}{0.00} & \MainTableValue{4.00}{0.00} & \MainTableValue{3.33}{0.00} & \MainTableValue{0.83}{0.00} \\
RLSD & \MainTableValue{58.57}{3.41} & \MainTableValue{39.42}{7.62} & \MainTableValue{59.26}{3.96} & \MainTableValue{57.03}{5.22} & \MainTableValue{58.00}{7.09} & \MainTableValue{0.00}{0.00} & \MainTableValue{46.61}{1.70} & \MainTableValue{0.48}{0.05} & \MainTableValue{0.74}{0.08} & \MainTableValue{4.94}{0.26} & \MainTableValue{4.00}{0.00} & \MainTableValue{3.54}{0.59} & \MainTableValue{1.89}{0.07} \\
OCSD & \MainTableValue{36.79}{3.22} & \MainTableValue{59.62}{5.44} & \MainTableValue{49.54}{3.93} & \MainTableValue{38.28}{2.21} & \MainTableValue{42.50}{2.98} & \MainTableValue{35.42}{4.45} & \MainTableValue{42.32}{2.15} & \MainTableValue{\MainTableBest{7.08}}{0.04} & \MainTableValue{\MainTableBest{9.47}}{0.08} & \MainTableValue{\MainTableBest{24.43}}{0.00} & \MainTableValue{\MainTableBest{29.00}}{0.00} & \MainTableValue{\underline{25.00}}{0.00} & \MainTableValue{\underline{13.78}}{0.03} \\
OVCSD & \MainTableValue{56.07}{5.70} & \MainTableValue{63.46}{7.96} & \MainTableValue{73.61}{2.37} & \MainTableValue{74.22}{5.22} & \MainTableValue{\MainTableBest{75.00}}{2.83} & \MainTableValue{5.73}{2.16} & \MainTableValue{56.96}{1.52} & \MainTableValue{\underline{6.02}}{0.38} & \MainTableValue{\underline{8.52}}{0.27} & \MainTableValue{\underline{22.31}}{0.41} & \MainTableValue{\underline{15.15}}{2.18} & \MainTableValue{\MainTableBest{45.83}}{3.56} & \cellcolor{MainTableBestFill}\MainTableValue{\MainTableBest{13.85}}{0.37} \\
AgentOPSD & \MainTableValue{\underline{71.43}}{3.41} & \MainTableValue{52.88}{6.42} & \MainTableValue{1.39}{1.92} & \MainTableValue{65.62}{3.34} & \MainTableValue{67.50}{3.34} & \MainTableValue{28.65}{5.65} & \MainTableValue{47.50}{1.75} & \MainTableValue{0.42}{0.00} & \MainTableValue{0.60}{0.00} & \MainTableValue{0.55}{0.00} & \MainTableValue{4.00}{0.00} & \MainTableValue{3.33}{0.00} & \MainTableValue{0.83}{0.00} \\
ADRS & \MainTableValue{31.43}{3.05} & \MainTableValue{32.69}{5.44} & \MainTableValue{0.00}{0.00} & \MainTableValue{0.00}{0.00} & \MainTableValue{1.50}{2.98} & \MainTableValue{10.94}{4.42} & \MainTableValue{13.04}{1.65} & \MainTableValue{0.42}{0.00} & \MainTableValue{0.60}{0.00} & \MainTableValue{0.55}{0.00} & \MainTableValue{4.00}{0.00} & \MainTableValue{3.33}{0.00} & \MainTableValue{0.83}{0.00} \\
ICSD & \MainTableValue{60.00}{3.05} & \MainTableValue{53.85}{8.22} & \MainTableValue{62.96}{3.96} & \MainTableValue{55.47}{6.19} & \MainTableValue{51.50}{3.96} & \MainTableValue{\underline{41.15}}{4.13} & \MainTableValue{54.73}{2.05} & \MainTableValue{0.42}{0.00} & \MainTableValue{0.60}{0.00} & \MainTableValue{0.55}{0.00} & \MainTableValue{4.00}{0.00} & \MainTableValue{3.33}{0.00} & \MainTableValue{0.83}{0.00} \\
PCSD & \MainTableValue{\MainTableBest{72.86}}{2.16} & \MainTableValue{\MainTableBest{73.08}}{4.11} & \MainTableValue{\MainTableBest{89.35}}{2.37} & \MainTableValue{\underline{75.78}}{2.21} & \MainTableValue{56.00}{3.02} & \MainTableValue{\MainTableBest{66.67}}{0.00} & \MainTableValue{\underline{72.32}}{0.92} & \MainTableValue{0.42}{0.00} & \MainTableValue{0.60}{0.00} & \MainTableValue{0.55}{0.00} & \MainTableValue{4.00}{0.00} & \MainTableValue{3.33}{0.00} & \MainTableValue{0.83}{0.00} \\
\midrule
\rowcolor{MainTableBand}
\multicolumn{14}{l}{\MainTableModel{Qwen3.5-4B}} \\
\rowcolor{MainTableHeader}
\textit{Pretrained} & \MainTableValue{68.57}{3.05} & \MainTableValue{45.19}{4.93} & \MainTableValue{20.83}{5.92} & \MainTableValue{3.12}{3.34} & \MainTableValue{12.00}{4.78} & \MainTableValue{36.46}{4.31} & \MainTableValue{34.11}{1.82} & \MainTableValue{-20.88}{10.62} & \MainTableValue{30.44}{3.86} & \MainTableValue{-30.62}{3.47} & \MainTableValue{-6.35}{14.94} & \MainTableValue{15.83}{4.45} & \MainTableValue{1.39}{3.20} \\
\rowcolor{MainTableBaseline}
\textbf{GRPO} & \MainTableValue{86.07}{2.38} & \MainTableValue{\underline{74.04}}{3.98} & \MainTableValue{\MainTableBest{93.98}}{1.92} & \MainTableValue{\MainTableBest{79.69}}{2.89} & \MainTableValue{\MainTableBest{87.50}}{1.41} & \MainTableValue{63.02}{3.48} & \MainTableValue{82.05}{1.50} & \MainTableValue{6.60}{0.33} & \MainTableValue{54.80}{0.71} & \MainTableValue{24.43}{0.14} & \MainTableValue{16.57}{1.20} & \MainTableValue{25.00}{0.00} & \MainTableValue{32.16}{0.30} \\
GRPO+OPSD & \MainTableValue{0.00}{0.00} & \MainTableValue{13.46}{5.44} & \MainTableValue{0.46}{1.31} & \MainTableValue{0.00}{0.00} & \MainTableValue{1.00}{1.85} & \MainTableValue{0.00}{0.00} & \MainTableValue{1.52}{0.71} & \MainTableValue{6.91}{0.16} & \MainTableValue{9.33}{0.16} & \MainTableValue{24.36}{0.09} & \MainTableValue{28.50}{1.41} & \MainTableValue{25.00}{0.00} & \MainTableValue{13.65}{0.06} \\
SDAR & \MainTableValue{89.64}{1.48} & \MainTableValue{47.12}{7.62} & \MainTableValue{78.70}{7.34} & \MainTableValue{52.34}{6.63} & \MainTableValue{74.50}{4.24} & \MainTableValue{\underline{86.98}}{4.13} & \MainTableValue{76.16}{2.29} & \MainTableValue{7.09}{0.00} & \MainTableValue{9.64}{0.00} & \MainTableValue{24.43}{0.00} & \MainTableValue{\MainTableBest{29.00}}{0.00} & \MainTableValue{25.00}{0.00} & \MainTableValue{13.86}{0.00} \\
Skill-SD & \MainTableValue{81.79}{4.57} & \MainTableValue{43.27}{11.58} & \MainTableValue{56.94}{7.12} & \MainTableValue{39.84}{5.73} & \MainTableValue{42.00}{6.76} & \MainTableValue{45.31}{6.84} & \MainTableValue{55.27}{2.59} & \MainTableValue{\MainTableBest{24.38}}{0.55} & \MainTableValue{\underline{56.66}}{0.46} & \MainTableValue{41.98}{0.27} & \MainTableValue{23.35}{1.83} & \MainTableValue{25.00}{0.00} & \MainTableValue{\underline{41.67}}{0.17} \\
RLSD & \MainTableValue{\underline{94.64}}{2.38} & \MainTableValue{\MainTableBest{80.77}}{4.11} & \MainTableValue{\underline{93.52}}{3.28} & \MainTableValue{\underline{77.34}}{10.53} & \MainTableValue{70.50}{4.75} & \MainTableValue{80.21}{2.95} & \cellcolor{MainTableBestFill}\MainTableValue{\MainTableBest{84.38}}{1.29} & \MainTableValue{10.07}{0.03} & \MainTableValue{9.46}{0.00} & \MainTableValue{\MainTableBest{52.82}}{0.00} & \MainTableValue{17.85}{1.04} & \MainTableValue{\MainTableBest{50.00}}{0.00} & \MainTableValue{22.41}{0.03} \\
OCSD & \MainTableValue{75.71}{3.05} & \MainTableValue{32.69}{7.96} & \MainTableValue{47.22}{3.83} & \MainTableValue{47.66}{8.80} & \MainTableValue{41.50}{6.02} & \MainTableValue{48.96}{3.69} & \MainTableValue{52.32}{2.47} & \MainTableValue{\underline{24.24}}{0.08} & \MainTableValue{\MainTableBest{57.24}}{0.25} & \MainTableValue{\underline{43.16}}{0.94} & \MainTableValue{14.10}{0.93} & \MainTableValue{23.33}{1.54} & \cellcolor{MainTableBestFill}\MainTableValue{\MainTableBest{41.77}}{0.34} \\
OVCSD & \MainTableValue{87.50}{3.39} & \MainTableValue{62.50}{8.66} & \MainTableValue{79.63}{3.96} & \MainTableValue{69.53}{5.22} & \MainTableValue{66.50}{5.63} & \MainTableValue{79.17}{4.98} & \MainTableValue{76.43}{2.19} & \MainTableValue{14.80}{2.83} & \MainTableValue{19.54}{2.31} & \MainTableValue{7.99}{1.41} & \MainTableValue{10.50}{5.63} & \MainTableValue{11.67}{2.52} & \MainTableValue{14.93}{1.57} \\
AgentOPSD & \MainTableValue{74.64}{4.16} & \MainTableValue{51.92}{5.44} & \MainTableValue{87.96}{4.31} & \MainTableValue{75.78}{7.04} & \MainTableValue{\underline{81.00}}{2.83} & \MainTableValue{66.67}{7.04} & \MainTableValue{75.00}{2.19} & \MainTableValue{5.50}{3.54} & \MainTableValue{55.34}{0.89} & \MainTableValue{25.51}{0.86} & \MainTableValue{25.10}{3.30} & \MainTableValue{28.54}{2.59} & \MainTableValue{32.80}{0.98} \\
ADRS & \MainTableValue{85.36}{5.39} & \MainTableValue{64.42}{3.98} & \MainTableValue{67.13}{5.02} & \MainTableValue{67.19}{7.28} & \MainTableValue{65.50}{6.39} & \MainTableValue{64.58}{8.03} & \MainTableValue{70.71}{3.10} & \MainTableValue{7.71}{0.26} & \MainTableValue{55.88}{0.93} & \MainTableValue{41.38}{2.15} & \MainTableValue{23.38}{3.01} & \MainTableValue{24.17}{0.89} & \MainTableValue{36.73}{0.46} \\
ICSD & \MainTableValue{\underline{94.64}}{3.22} & \MainTableValue{70.19}{2.72} & \MainTableValue{90.74}{3.43} & \MainTableValue{64.84}{5.73} & \MainTableValue{75.50}{1.41} & \MainTableValue{81.25}{3.86} & \MainTableValue{\underline{82.50}}{1.62} & \MainTableValue{7.68}{0.11} & \MainTableValue{9.87}{0.13} & \MainTableValue{24.91}{0.14} & \MainTableValue{\underline{28.75}}{0.71} & \MainTableValue{\underline{48.75}}{1.73} & \MainTableValue{15.99}{0.19} \\
PCSD & \MainTableValue{\MainTableBest{96.79}}{1.83} & \MainTableValue{48.08}{6.82} & \MainTableValue{77.78}{3.43} & \MainTableValue{69.53}{7.79} & \MainTableValue{73.50}{2.98} & \MainTableValue{\MainTableBest{88.02}}{4.13} & \MainTableValue{79.82}{1.86} & \MainTableValue{6.80}{0.15} & \MainTableValue{38.79}{0.61} & \MainTableValue{17.14}{2.49} & \MainTableValue{17.10}{0.93} & \MainTableValue{23.33}{2.82} & \MainTableValue{23.85}{0.58} \\
\midrule
\rowcolor{MainTableBand}
\multicolumn{14}{l}{\MainTableModel{Qwen3.5-9B}} \\
\rowcolor{MainTableHeader}
\textit{Pretrained} & \MainTableValue{71.07}{3.22} & \MainTableValue{41.35}{7.05} & \MainTableValue{20.83}{7.65} & \MainTableValue{8.59}{6.63} & \MainTableValue{4.50}{3.96} & \MainTableValue{41.67}{5.89} & \MainTableValue{34.55}{1.43} & \MainTableValue{-12.74}{2.80} & \MainTableValue{29.29}{3.01} & \MainTableValue{-25.52}{4.79} & \MainTableValue{13.85}{0.71} & \MainTableValue{-22.08}{15.29} & \MainTableValue{1.86}{1.28} \\
\rowcolor{MainTableBaseline}
\textbf{GRPO} & \MainTableValue{92.50}{2.13} & \MainTableValue{76.92}{8.22} & \MainTableValue{46.30}{5.94} & \MainTableValue{59.38}{5.79} & \MainTableValue{18.50}{2.98} & \MainTableValue{78.12}{5.79} & \MainTableValue{62.68}{1.19} & \MainTableValue{33.70}{0.63} & \MainTableValue{\MainTableBest{60.00}}{0.00} & \MainTableValue{76.78}{0.49} & \MainTableValue{\MainTableBest{30.60}}{0.00} & \MainTableValue{\MainTableBest{100.00}}{0.00} & \cellcolor{MainTableBestFill}\MainTableValue{\MainTableBest{58.99}}{0.27} \\
GRPO+OPSD & \MainTableValue{93.21}{3.03} & \MainTableValue{66.35}{7.05} & \MainTableValue{83.33}{4.85} & \MainTableValue{49.22}{8.48} & \MainTableValue{38.50}{4.75} & \MainTableValue{80.21}{6.20} & \MainTableValue{71.79}{2.53} & \MainTableValue{\underline{39.84}}{0.03} & \MainTableValue{58.67}{0.63} & \MainTableValue{40.34}{1.10} & \MainTableValue{28.00}{1.85} & \MainTableValue{\MainTableBest{100.00}}{0.00} & \MainTableValue{51.98}{0.38} \\
SDAR & \MainTableValue{\MainTableBest{98.93}}{1.48} & \MainTableValue{\underline{89.42}}{9.14} & \MainTableValue{69.91}{5.02} & \MainTableValue{59.38}{8.84} & \MainTableValue{46.00}{3.02} & \MainTableValue{82.29}{3.69} & \MainTableValue{75.62}{2.04} & \MainTableValue{26.01}{3.06} & \MainTableValue{\underline{59.99}}{0.02} & \MainTableValue{\MainTableBest{81.55}}{1.16} & \MainTableValue{\MainTableBest{30.60}}{0.00} & \MainTableValue{\MainTableBest{100.00}}{0.00} & \MainTableValue{58.00}{0.88} \\
Skill-SD & \MainTableValue{93.93}{3.22} & \MainTableValue{67.31}{6.82} & \MainTableValue{70.37}{4.43} & \MainTableValue{54.69}{4.42} & \MainTableValue{61.00}{4.66} & \MainTableValue{82.29}{4.85} & \MainTableValue{74.55}{1.08} & \MainTableValue{6.72}{0.31} & \MainTableValue{40.65}{1.33} & \MainTableValue{30.20}{1.14} & \MainTableValue{28.90}{1.73} & \MainTableValue{40.83}{2.52} & \MainTableValue{29.08}{0.39} \\
RLSD & \MainTableValue{97.14}{2.16} & \MainTableValue{85.58}{4.93} & \MainTableValue{\underline{93.98}}{2.76} & \MainTableValue{71.88}{4.72} & \MainTableValue{\MainTableBest{87.00}}{1.85} & \MainTableValue{88.02}{2.67} & \MainTableValue{\underline{89.20}}{1.40} & \MainTableValue{31.07}{0.74} & \MainTableValue{57.29}{0.00} & \MainTableValue{71.01}{0.56} & \MainTableValue{\MainTableBest{30.60}}{0.00} & \MainTableValue{25.00}{0.00} & \MainTableValue{50.27}{0.27} \\
OCSD & \MainTableValue{98.21}{2.13} & \MainTableValue{76.92}{7.12} & \MainTableValue{84.72}{3.67} & \MainTableValue{71.09}{8.80} & \MainTableValue{\underline{83.00}}{5.13} & \MainTableValue{\underline{88.54}}{6.20} & \MainTableValue{86.16}{1.75} & \MainTableValue{-10.32}{2.11} & \MainTableValue{\MainTableBest{60.00}}{0.00} & \MainTableValue{63.22}{2.53} & \MainTableValue{\underline{29.00}}{0.00} & \MainTableValue{\MainTableBest{100.00}}{0.00} & \MainTableValue{44.30}{0.86} \\
OVCSD & \MainTableValue{97.50}{2.83} & \MainTableValue{77.88}{4.93} & \MainTableValue{76.85}{2.62} & \MainTableValue{69.53}{5.22} & \MainTableValue{81.50}{4.75} & \MainTableValue{84.38}{2.95} & \MainTableValue{83.39}{1.36} & \MainTableValue{-15.07}{2.73} & \MainTableValue{59.47}{0.02} & \MainTableValue{65.92}{0.71} & \MainTableValue{\MainTableBest{30.60}}{0.00} & \MainTableValue{\MainTableBest{100.00}}{0.00} & \MainTableValue{43.45}{0.80} \\
AgentOPSD & \MainTableValue{\underline{98.57}}{1.53} & \MainTableValue{67.31}{6.82} & \MainTableValue{64.81}{5.94} & \MainTableValue{\underline{75.00}}{7.47} & \MainTableValue{66.50}{6.74} & \MainTableValue{82.81}{4.13} & \MainTableValue{78.04}{3.32} & \MainTableValue{-29.13}{2.73} & \MainTableValue{\underline{59.99}}{0.02} & \MainTableValue{\underline{78.83}}{0.84} & \MainTableValue{18.40}{0.57} & \MainTableValue{\MainTableBest{100.00}}{0.00} & \MainTableValue{42.48}{0.78} \\
ADRS & \MainTableValue{82.50}{2.83} & \MainTableValue{71.15}{5.44} & \MainTableValue{91.67}{3.28} & \MainTableValue{69.53}{4.01} & \MainTableValue{76.00}{6.41} & \MainTableValue{72.92}{3.86} & \MainTableValue{78.93}{1.15} & \MainTableValue{-45.33}{4.90} & \MainTableValue{\MainTableBest{60.00}}{0.00} & \MainTableValue{65.80}{1.00} & \MainTableValue{\MainTableBest{30.60}}{0.00} & \MainTableValue{\MainTableBest{100.00}}{0.00} & \MainTableValue{35.63}{1.18} \\
ICSD & \MainTableValue{95.00}{2.02} & \MainTableValue{86.54}{9.86} & \MainTableValue{\MainTableBest{98.15}}{2.80} & \MainTableValue{\MainTableBest{90.62}}{7.47} & \MainTableValue{73.50}{4.24} & \MainTableValue{\MainTableBest{90.10}}{4.42} & \cellcolor{MainTableBestFill}\MainTableValue{\MainTableBest{89.64}}{2.62} & \MainTableValue{-18.86}{2.09} & \MainTableValue{28.45}{0.31} & \MainTableValue{31.22}{0.39} & \MainTableValue{27.00}{0.00} & \MainTableValue{52.08}{1.73} & \MainTableValue{18.26}{0.59} \\
PCSD & \MainTableValue{95.36}{4.02} & \MainTableValue{\MainTableBest{90.38}}{3.56} & \MainTableValue{82.41}{5.14} & \MainTableValue{55.47}{7.04} & \MainTableValue{29.50}{5.21} & \MainTableValue{76.04}{6.20} & \MainTableValue{72.77}{1.77} & \MainTableValue{\MainTableBest{51.54}}{0.12} & \MainTableValue{\MainTableBest{60.00}}{0.00} & \MainTableValue{57.73}{0.13} & \MainTableValue{28.60}{0.00} & \MainTableValue{\underline{88.33}}{4.71} & \MainTableValue{\underline{58.60}}{0.38} \\
\bottomrule
\end{tabular}%
}
\arrayrulecolor{black}
\endgroup
\caption{Results for pretrained models and eleven training methods on ALFWorld and ScienceWorld. Entries report mean$_{\pm\mathrm{SD}}$ over eight rollout seeds at $T=0.4$; trained models are evaluated after 100 global steps. ALFWorld reports success (\%), and ScienceWorld reports task score. Teal bold and underlining mark the highest and second-highest means within each model and metric, including ties. Overall columns are shaded.}
\label{tab:main_results}
\end{table*}

%% file: styles/main_results_style.tex
\definecolor{MainTableInk}{HTML}{162B40}
\definecolor{MainTableTeal}{HTML}{167A83}
\definecolor{MainTableMuted}{HTML}{5E6F80}
\definecolor{MainTableBand}{HTML}{E3EDF3}
\definecolor{MainTableHeader}{HTML}{F0F4F7}
\definecolor{MainTableFocus}{HTML}{F1F7F8}
\definecolor{MainTableBestFill}{HTML}{D9EDEB}
\definecolor{MainTableBaseline}{HTML}{F3F4F5}

\newcommand{\MainTableValue}[2]{$#1_{\textcolor{MainTableMuted}{\pm #2}}$}
\newcommand{\MainTableBest}[1]{\textcolor{MainTableTeal}{\mathbf{#1}}}
\newcommand{\MainTableModel}[1]{\textcolor{MainTableInk}{\textbf{#1}}}

%% file: table/generated_main_results_search_webshop.tex
\begin{table*}[!t]
\begingroup
\input{styles/main_results_style}
\centering
\footnotesize
\setlength{\tabcolsep}{1.8pt}
\renewcommand{\arraystretch}{1.08}
\resizebox{\textwidth}{!}{%
\arrayrulecolor{MainTableInk}
\begin{tabular}{lccccccc>{\columncolor{MainTableFocus}}cc>{\columncolor{MainTableFocus}}c}
\toprule
\rowcolor{MainTableInk}
 & \multicolumn{8}{c}{\color{white}\bfseries Search-QA: exact-match accuracy (\%)} & \multicolumn{2}{c}{\color{white}\bfseries WebShop} \\
\rowcolor{MainTableHeader}
\textbf{Method} & NQ & Triv & Pop & Hotp & 2Wk & MuS & Bam & \textbf{Overall} & Score & \textbf{Acc. (\%)} \\
\midrule
\rowcolor{MainTableBand}
\multicolumn{11}{l}{\MainTableModel{Qwen3.5-2B}} \\
\rowcolor{MainTableHeader}
\textit{Pretrained} & \MainTableValue{9.53}{0.83} & \MainTableValue{24.20}{0.95} & \MainTableValue{13.00}{0.96} & \MainTableValue{12.78}{1.00} & \MainTableValue{23.70}{1.46} & \MainTableValue{2.43}{0.57} & \MainTableValue{12.30}{1.48} & \MainTableValue{14.19}{0.43} & \MainTableValue{6.04}{1.31} & \MainTableValue{1.86}{0.72} \\
\rowcolor{MainTableBaseline}
\textbf{GRPO} & \MainTableValue{37.93}{0.40} & \MainTableValue{56.03}{0.36} & \MainTableValue{47.93}{0.71} & \MainTableValue{27.10}{0.61} & \MainTableValue{23.48}{0.43} & \MainTableValue{7.10}{0.32} & \MainTableValue{11.40}{0.93} & \MainTableValue{32.38}{0.28} & \MainTableValue{85.01}{1.22} & \MainTableValue{70.80}{0.93} \\
GRPO+OPSD & \MainTableValue{39.68}{0.59} & \MainTableValue{59.03}{0.57} & \MainTableValue{48.00}{0.72} & \MainTableValue{31.78}{0.84} & \MainTableValue{26.23}{0.65} & \MainTableValue{6.38}{0.52} & \MainTableValue{11.20}{0.74} & \MainTableValue{34.22}{0.24} & \MainTableValue{\MainTableBest{86.76}}{0.75} & \cellcolor{MainTableBestFill}\MainTableValue{\MainTableBest{77.05}}{0.93} \\
SDAR & \MainTableValue{38.75}{0.66} & \MainTableValue{57.18}{0.65} & \MainTableValue{48.00}{0.39} & \MainTableValue{39.15}{1.28} & \MainTableValue{31.23}{1.08} & \MainTableValue{\MainTableBest{16.05}}{0.81} & \MainTableValue{\underline{32.30}}{2.09} & \MainTableValue{\underline{38.15}}{0.55} & \MainTableValue{80.21}{1.11} & \MainTableValue{63.67}{1.10} \\
Skill-SD & \MainTableValue{\MainTableBest{40.65}}{0.54} & \MainTableValue{58.73}{0.61} & \MainTableValue{48.25}{0.33} & \MainTableValue{36.93}{0.76} & \MainTableValue{\MainTableBest{35.08}}{0.83} & \MainTableValue{10.15}{0.93} & \MainTableValue{19.80}{1.90} & \MainTableValue{37.56}{0.26} & \MainTableValue{81.33}{0.58} & \MainTableValue{71.19}{0.97} \\
RLSD & \MainTableValue{40.03}{1.40} & \MainTableValue{\underline{60.30}}{1.12} & \MainTableValue{43.70}{1.07} & \MainTableValue{\underline{39.93}}{0.87} & \MainTableValue{30.38}{1.35} & \MainTableValue{13.98}{0.57} & \MainTableValue{\MainTableBest{35.00}}{1.95} & \MainTableValue{37.93}{0.13} & \MainTableValue{79.86}{1.42} & \MainTableValue{69.73}{1.49} \\
OCSD & \MainTableValue{8.45}{0.32} & \MainTableValue{20.63}{0.89} & \MainTableValue{11.58}{0.56} & \MainTableValue{12.63}{0.23} & \MainTableValue{24.73}{0.37} & \MainTableValue{2.30}{0.26} & \MainTableValue{2.90}{0.60} & \MainTableValue{12.96}{0.15} & \MainTableValue{\underline{86.59}}{1.01} & \MainTableValue{\underline{75.59}}{1.60} \\
OVCSD & \MainTableValue{32.85}{0.45} & \MainTableValue{51.68}{0.72} & \MainTableValue{40.98}{1.25} & \MainTableValue{34.33}{1.25} & \MainTableValue{29.68}{0.82} & \MainTableValue{11.40}{0.89} & \MainTableValue{26.70}{1.71} & \MainTableValue{33.21}{0.33} & \MainTableValue{3.39}{1.09} & \MainTableValue{1.07}{0.83} \\
AgentOPSD & \MainTableValue{39.98}{0.70} & \MainTableValue{57.18}{0.96} & \MainTableValue{48.53}{0.50} & \MainTableValue{32.50}{1.02} & \MainTableValue{24.70}{1.49} & \MainTableValue{7.33}{0.71} & \MainTableValue{21.50}{2.55} & \MainTableValue{34.49}{0.36} & \MainTableValue{81.45}{1.61} & \MainTableValue{70.51}{1.95} \\
ADRS & \MainTableValue{9.30}{0.60} & \MainTableValue{25.18}{0.64} & \MainTableValue{12.98}{0.48} & \MainTableValue{13.70}{0.43} & \MainTableValue{23.85}{0.33} & \MainTableValue{1.93}{0.28} & \MainTableValue{4.90}{0.79} & \MainTableValue{14.10}{0.21} & \MainTableValue{74.21}{1.37} & \MainTableValue{63.09}{1.30} \\
ICSD & \MainTableValue{\underline{40.40}}{0.79} & \MainTableValue{\MainTableBest{60.55}}{0.75} & \MainTableValue{\underline{48.83}}{1.22} & \MainTableValue{\MainTableBest{41.68}}{0.69} & \MainTableValue{\underline{32.78}}{1.15} & \MainTableValue{\underline{15.53}}{0.76} & \MainTableValue{25.70}{1.73} & \cellcolor{MainTableBestFill}\MainTableValue{\MainTableBest{39.39}}{0.19} & \MainTableValue{81.79}{0.32} & \MainTableValue{70.51}{1.00} \\
PCSD & \MainTableValue{39.83}{0.41} & \MainTableValue{56.25}{0.60} & \MainTableValue{\MainTableBest{49.53}}{0.28} & \MainTableValue{29.98}{0.54} & \MainTableValue{23.30}{0.57} & \MainTableValue{6.65}{0.33} & \MainTableValue{7.30}{1.17} & \MainTableValue{33.18}{0.27} & \MainTableValue{84.96}{0.84} & \MainTableValue{74.71}{0.83} \\
\midrule
\rowcolor{MainTableBand}
\multicolumn{11}{l}{\MainTableModel{Qwen3.5-4B}} \\
\rowcolor{MainTableHeader}
\textit{Pretrained} & \MainTableValue{18.05}{0.82} & \MainTableValue{47.45}{1.13} & \MainTableValue{20.50}{1.26} & \MainTableValue{21.55}{1.02} & \MainTableValue{28.18}{1.28} & \MainTableValue{5.50}{0.73} & \MainTableValue{32.90}{2.88} & \MainTableValue{23.91}{0.61} & \MainTableValue{9.19}{1.73} & \MainTableValue{5.57}{1.35} \\
\rowcolor{MainTableBaseline}
\textbf{GRPO} & \MainTableValue{41.78}{0.73} & \MainTableValue{66.68}{0.63} & \MainTableValue{49.43}{0.38} & \MainTableValue{47.00}{1.12} & \MainTableValue{\MainTableBest{46.83}}{1.52} & \MainTableValue{19.35}{1.11} & \MainTableValue{44.50}{1.86} & \MainTableValue{\underline{45.15}}{0.29} & \MainTableValue{74.21}{1.55} & \MainTableValue{66.02}{1.56} \\
GRPO+OPSD & \MainTableValue{43.48}{0.80} & \MainTableValue{65.08}{0.55} & \MainTableValue{48.43}{0.59} & \MainTableValue{45.78}{0.58} & \MainTableValue{40.70}{0.60} & \MainTableValue{21.10}{0.52} & \MainTableValue{45.80}{2.13} & \MainTableValue{44.16}{0.21} & \MainTableValue{\MainTableBest{88.75}}{1.06} & \cellcolor{MainTableBestFill}\MainTableValue{\MainTableBest{77.15}}{1.30} \\
SDAR & \MainTableValue{\underline{45.38}}{0.71} & \MainTableValue{65.75}{0.78} & \MainTableValue{49.60}{0.81} & \MainTableValue{\MainTableBest{48.83}}{1.37} & \MainTableValue{41.08}{0.95} & \MainTableValue{20.58}{1.13} & \MainTableValue{39.10}{2.03} & \MainTableValue{44.96}{0.37} & \MainTableValue{61.14}{1.86} & \MainTableValue{51.17}{1.45} \\
Skill-SD & \MainTableValue{44.93}{1.04} & \MainTableValue{63.95}{0.85} & \MainTableValue{46.78}{0.58} & \MainTableValue{44.10}{0.94} & \MainTableValue{37.30}{1.25} & \MainTableValue{19.43}{1.54} & \MainTableValue{44.40}{0.96} & \MainTableValue{42.81}{0.22} & \MainTableValue{0.00}{0.00} & \MainTableValue{0.00}{0.00} \\
RLSD & \MainTableValue{45.13}{0.78} & \MainTableValue{65.00}{0.81} & \MainTableValue{46.95}{0.91} & \MainTableValue{45.10}{1.05} & \MainTableValue{39.43}{1.68} & \MainTableValue{19.98}{1.02} & \MainTableValue{43.60}{3.05} & \MainTableValue{43.60}{0.51} & \MainTableValue{80.27}{1.54} & \MainTableValue{68.16}{1.60} \\
OCSD & \MainTableValue{44.50}{0.68} & \MainTableValue{\underline{67.05}}{0.83} & \MainTableValue{\underline{49.63}}{0.63} & \MainTableValue{45.95}{0.81} & \MainTableValue{39.13}{1.19} & \MainTableValue{\MainTableBest{21.58}}{0.87} & \MainTableValue{47.00}{1.26} & \MainTableValue{44.73}{0.35} & \MainTableValue{70.32}{1.06} & \MainTableValue{64.16}{0.77} \\
OVCSD & \MainTableValue{43.45}{0.59} & \MainTableValue{64.90}{1.02} & \MainTableValue{45.58}{0.76} & \MainTableValue{43.78}{0.66} & \MainTableValue{41.15}{0.87} & \MainTableValue{21.33}{0.88} & \MainTableValue{\underline{47.10}}{2.03} & \MainTableValue{43.51}{0.36} & \MainTableValue{\underline{83.79}}{1.46} & \MainTableValue{\underline{73.73}}{1.18} \\
AgentOPSD & \MainTableValue{44.95}{0.67} & \MainTableValue{64.65}{0.82} & \MainTableValue{45.83}{1.02} & \MainTableValue{46.10}{1.20} & \MainTableValue{41.65}{1.33} & \MainTableValue{\underline{21.43}}{0.92} & \MainTableValue{\MainTableBest{47.90}}{1.98} & \MainTableValue{44.25}{0.53} & \MainTableValue{80.33}{1.98} & \MainTableValue{72.07}{2.32} \\
ADRS & \MainTableValue{41.75}{0.54} & \MainTableValue{60.85}{0.46} & \MainTableValue{44.75}{0.73} & \MainTableValue{44.78}{0.67} & \MainTableValue{39.75}{0.81} & \MainTableValue{\underline{21.43}}{1.13} & \MainTableValue{44.50}{1.65} & \MainTableValue{42.31}{0.33} & \MainTableValue{82.70}{1.21} & \MainTableValue{65.72}{1.53} \\
ICSD & \MainTableValue{\MainTableBest{45.98}}{0.47} & \MainTableValue{\MainTableBest{68.03}}{0.95} & \MainTableValue{\MainTableBest{50.50}}{0.41} & \MainTableValue{\underline{47.65}}{0.66} & \MainTableValue{\underline{43.73}}{0.97} & \MainTableValue{20.18}{1.02} & \MainTableValue{42.60}{1.03} & \cellcolor{MainTableBestFill}\MainTableValue{\MainTableBest{45.87}}{0.36} & \MainTableValue{0.00}{0.00} & \MainTableValue{0.00}{0.00} \\
PCSD & \MainTableValue{44.23}{0.77} & \MainTableValue{65.78}{0.98} & \MainTableValue{46.65}{1.12} & \MainTableValue{43.88}{0.86} & \MainTableValue{40.25}{0.81} & \MainTableValue{17.53}{0.66} & \MainTableValue{44.60}{2.21} & \MainTableValue{43.11}{0.35} & \MainTableValue{83.06}{1.28} & \MainTableValue{72.17}{1.50} \\
\midrule
\rowcolor{MainTableBand}
\multicolumn{11}{l}{\MainTableModel{Qwen3.5-9B}} \\
\rowcolor{MainTableHeader}
\textit{Pretrained} & \MainTableValue{19.13}{0.83} & \MainTableValue{55.30}{1.22} & \MainTableValue{23.50}{1.22} & \MainTableValue{21.20}{0.93} & \MainTableValue{30.08}{1.48} & \MainTableValue{6.60}{0.60} & \MainTableValue{43.00}{1.99} & \MainTableValue{26.65}{0.23} & \MainTableValue{9.89}{1.01} & \MainTableValue{5.86}{1.10} \\
\rowcolor{MainTableBaseline}
\textbf{GRPO} & \MainTableValue{45.68}{0.55} & \MainTableValue{66.73}{0.98} & \MainTableValue{48.15}{0.52} & \MainTableValue{48.45}{1.14} & \MainTableValue{\underline{47.53}}{1.84} & \MainTableValue{21.90}{1.11} & \MainTableValue{51.00}{1.80} & \MainTableValue{46.59}{0.53} & \MainTableValue{83.30}{1.16} & \MainTableValue{70.02}{1.38} \\
GRPO+OPSD & \MainTableValue{46.50}{0.88} & \MainTableValue{67.33}{0.49} & \MainTableValue{\underline{50.68}}{0.76} & \MainTableValue{47.58}{1.08} & \MainTableValue{46.48}{1.44} & \MainTableValue{22.55}{0.81} & \MainTableValue{\MainTableBest{52.70}}{3.04} & \MainTableValue{\underline{47.08}}{0.35} & \MainTableValue{\MainTableBest{85.31}}{1.34} & \cellcolor{MainTableBestFill}\MainTableValue{\MainTableBest{74.90}}{0.77} \\
SDAR & \MainTableValue{42.65}{0.41} & \MainTableValue{\underline{68.38}}{0.95} & \MainTableValue{47.00}{1.60} & \MainTableValue{48.28}{1.22} & \MainTableValue{39.93}{0.79} & \MainTableValue{21.40}{0.65} & \MainTableValue{48.20}{1.75} & \MainTableValue{44.75}{0.55} & \MainTableValue{\underline{85.19}}{1.61} & \MainTableValue{\underline{74.32}}{2.19} \\
Skill-SD & \MainTableValue{46.10}{0.51} & \MainTableValue{67.53}{0.70} & \MainTableValue{48.68}{0.80} & \MainTableValue{\underline{48.75}}{0.52} & \MainTableValue{44.35}{1.63} & \MainTableValue{21.75}{0.83} & \MainTableValue{46.10}{1.48} & \MainTableValue{46.19}{0.28} & \MainTableValue{0.00}{0.00} & \MainTableValue{0.00}{0.00} \\
RLSD & \MainTableValue{43.63}{0.39} & \MainTableValue{67.20}{0.60} & \MainTableValue{47.20}{1.46} & \MainTableValue{47.35}{0.73} & \MainTableValue{38.15}{1.18} & \MainTableValue{21.05}{0.42} & \MainTableValue{49.50}{2.03} & \MainTableValue{44.31}{0.29} & \MainTableValue{0.00}{0.00} & \MainTableValue{0.00}{0.00} \\
OCSD & \MainTableValue{\MainTableBest{47.88}}{0.91} & \MainTableValue{\MainTableBest{68.53}}{0.81} & \MainTableValue{\MainTableBest{52.98}}{0.79} & \MainTableValue{46.93}{0.70} & \MainTableValue{\MainTableBest{49.43}}{1.28} & \MainTableValue{\MainTableBest{23.80}}{0.62} & \MainTableValue{45.80}{2.13} & \cellcolor{MainTableBestFill}\MainTableValue{\MainTableBest{48.16}}{0.44} & \MainTableValue{77.03}{2.08} & \MainTableValue{67.77}{2.20} \\
OVCSD & \MainTableValue{44.40}{0.57} & \MainTableValue{67.30}{0.99} & \MainTableValue{47.00}{1.17} & \MainTableValue{47.55}{0.82} & \MainTableValue{43.55}{1.35} & \MainTableValue{21.28}{0.85} & \MainTableValue{51.40}{2.45} & \MainTableValue{45.43}{0.47} & \MainTableValue{80.36}{1.83} & \MainTableValue{68.07}{1.28} \\
AgentOPSD & \MainTableValue{46.15}{0.75} & \MainTableValue{67.30}{1.12} & \MainTableValue{47.33}{0.63} & \MainTableValue{\MainTableBest{49.05}}{0.52} & \MainTableValue{43.68}{1.17} & \MainTableValue{20.50}{0.93} & \MainTableValue{49.20}{1.91} & \MainTableValue{45.81}{0.50} & \MainTableValue{83.72}{1.80} & \MainTableValue{72.27}{2.09} \\
ADRS & \MainTableValue{40.58}{0.58} & \MainTableValue{66.38}{0.46} & \MainTableValue{43.48}{0.79} & \MainTableValue{45.93}{0.84} & \MainTableValue{43.08}{1.01} & \MainTableValue{22.55}{0.48} & \MainTableValue{50.30}{0.79} & \MainTableValue{43.93}{0.37} & \MainTableValue{81.20}{1.62} & \MainTableValue{70.31}{1.10} \\
ICSD & \MainTableValue{\underline{46.55}}{0.98} & \MainTableValue{66.88}{0.60} & \MainTableValue{50.00}{0.59} & \MainTableValue{48.40}{0.66} & \MainTableValue{45.08}{1.23} & \MainTableValue{\underline{23.55}}{0.87} & \MainTableValue{49.30}{1.65} & \MainTableValue{46.84}{0.41} & \MainTableValue{81.89}{2.44} & \MainTableValue{72.46}{1.37} \\
PCSD & \MainTableValue{44.68}{1.00} & \MainTableValue{67.68}{0.79} & \MainTableValue{47.50}{0.80} & \MainTableValue{47.03}{0.60} & \MainTableValue{42.18}{1.57} & \MainTableValue{21.95}{0.67} & \MainTableValue{\underline{51.50}}{1.13} & \MainTableValue{45.42}{0.55} & \MainTableValue{83.67}{0.52} & \MainTableValue{\underline{74.32}}{0.65} \\
\bottomrule
\end{tabular}%
}
\arrayrulecolor{black}
\endgroup
\caption{Results on Search-QA and WebShop: mean$_{\pm\mathrm{SD}}$ over eight rollout seeds at $T=0.4$, after 100 steps for trained models. We report exact-match accuracy (\%) for Search-QA and task score (Score) and success rate (\%, Acc.) for WebShop. Highlighting follows Table~\ref{tab:main_results}.}
\label{tab:main_results_search_webshop}
\end{table*}

%% file: sec/05_empirical_comparison.tex
\section{Comparative Performance under Common Evaluation}
\label{sec:results}

\input{figure/validation_curves}
\input{figure/cost_performance_settings}

The central question is whether additional training mechanisms deliver gains beyond a few favorable settings.
We compare each alternative with GRPO across all twelve model--task settings, examining both the frequency and magnitude of improvements.
Tables~\ref{tab:main_results} and~\ref{tab:main_results_search_webshop} report the results for all methods and their pretrained backbones under the common evaluation protocol.
Appendix~\ref{app:searchqa-decoding} analyzes the discrepancy between step-100 validation and repeated final evaluation on Search-QA.

\noindent \textbf{More methods do not imply broad progress.}
All ten alternatives improve on GRPO in some settings, but only a minority retain an advantage in aggregate.
Only PCSD (\PaperTwelvePCSDMean{}, +\PaperTwelvePCSDGain{} points) and AgentOPSD (\PaperTwelveAgentOPSDMean{}, +\PaperTwelveAgentOPSDGain{} points) exceed GRPO's twelve-setting mean of \PaperTwelveGRPOMean{}; the other eight alternatives fall below it.
Since all ten alternatives yield gains somewhere, isolated positive results do little to establish progress beyond a strong outcome-only baseline.
A broadly applicable training improvement must be assessed across the range of tasks and model sizes it is expected to support.

\noindent \textbf{No method dominates across settings.}
The leading method depends on both the benchmark and model size.
ALFWorld has a different leader at each scale---Skill-SD, RLSD, and ICSD at 2B, 4B, and 9B---whereas GRPO+OPSD leads WebShop at all three scales.
Search-QA is led by ICSD at 2B and 4B and OCSD at 9B; ScienceWorld's leaders at 2B, 4B, and 9B are OVCSD, OCSD, and GRPO, respectively.
Even PCSD, which has the highest twelve-setting mean, leads no individual setting and beats GRPO in only six, trailing it in five and tying in one.
Selecting the best aggregate performer and choosing a winner for a particular setting are therefore different method-selection problems.
Neither a collection of first-place results nor the highest aggregate score establishes a method as uniformly superior; both must be interpreted alongside its performance across the full set of settings.

\noindent \textbf{Win counts can conceal severe regressions.}
ICSD outperforms GRPO in seven of twelve settings, yet its mean of \PaperTwelveICSDMean{} remains below GRPO's \PaperTwelveGRPOMean{}.
Its 4B WebShop run ends with 0.00\% success, illustrating a severe failure that a win count does not distinguish from a small regression.
GRPO+OPSD likewise wins six settings but averages only \PaperTwelveGRPOOPSDMean{}, with 1.52\% success on 4B ALFWorld.
Counting each loss equally discards whether a method narrowly trails the baseline or loses nearly all task-solving ability.
Severe failures can outweigh gains elsewhere, so assessing progress requires considering the magnitude of losses as well as the frequency of improvements.
Appendix~\ref{app:endpoint-moments} reports the twelve-setting summary and examines low-performance outcomes across all four benchmarks.

%% file: figure/validation_curves.tex
\begin{figure*}[t]
\centering
\includegraphics[width=\textwidth]{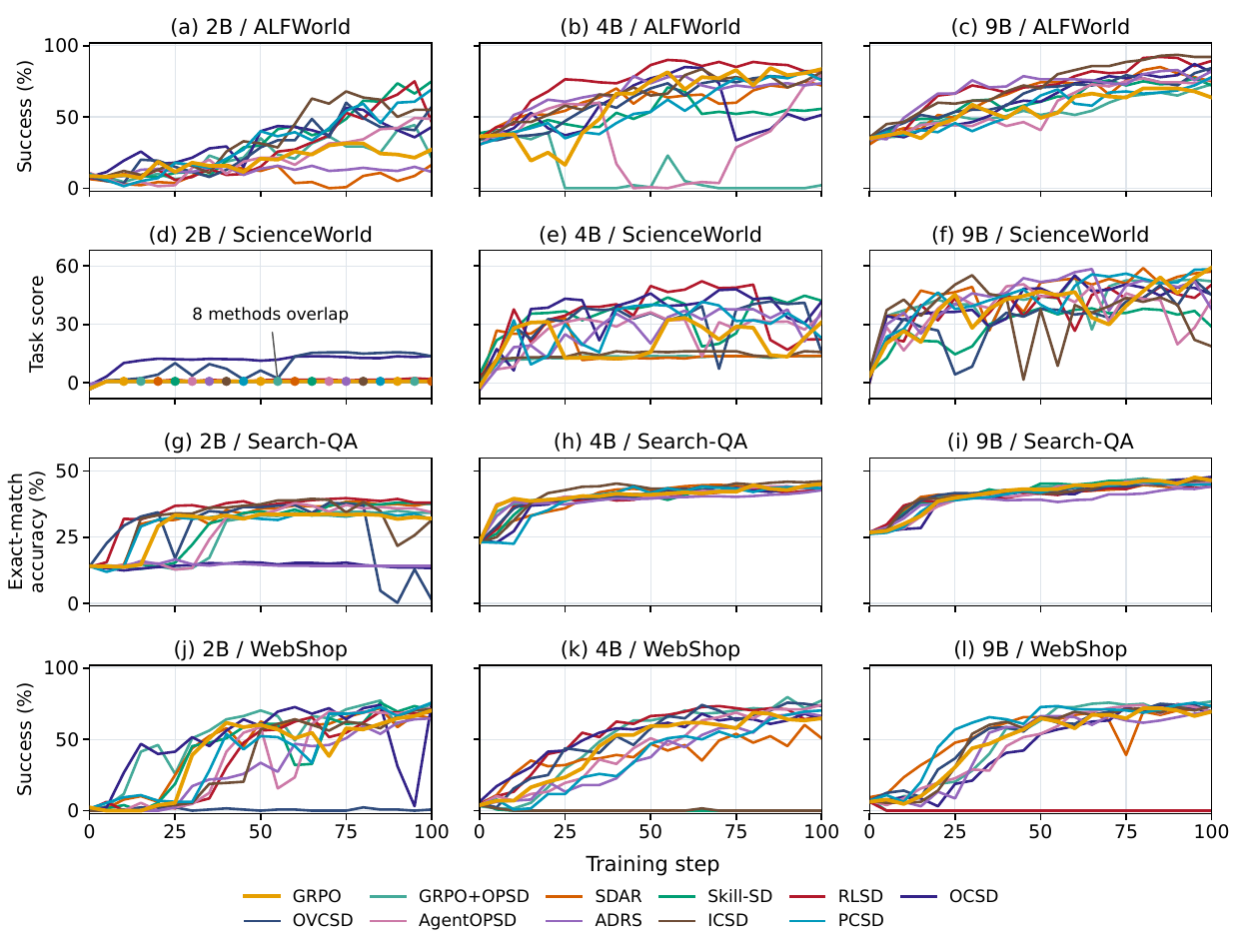}
\caption{Validation trajectories of eleven methods across four benchmarks and three model sizes, covering \PaperCompletedRuns{} training runs and \PaperCompletedValidations{} validation sweeps. Curves show validation scores recorded throughout training. Colored markers distinguish overlapping curves on 2B ScienceWorld.}
\label{fig:validation-curves}
\end{figure*}

%% file: figure/cost_performance_settings.tex
\begin{figure*}[!t]
\centering
\includegraphics[width=\textwidth]{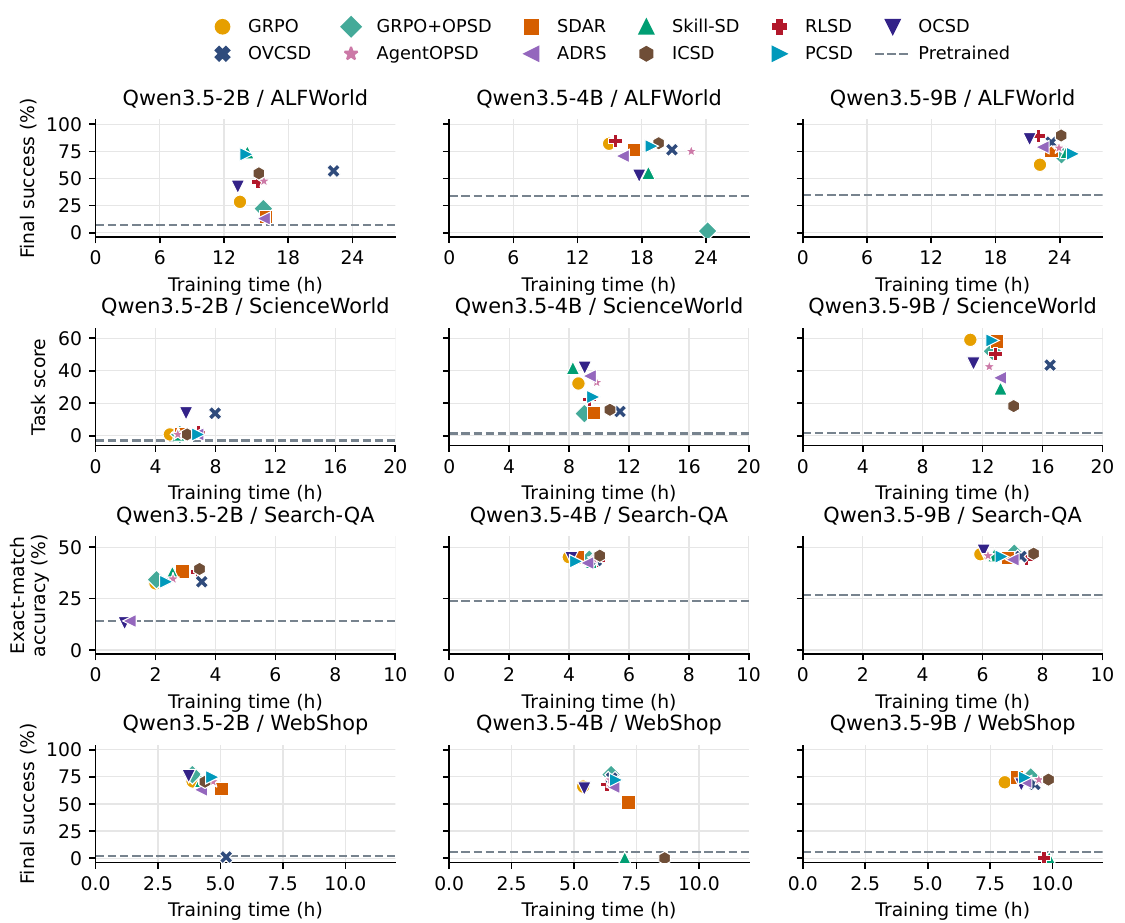}
\caption{Final performance versus training time across twelve model--task settings. Points show mean performance over eight rollout seeds; dashed lines mark pretrained means. Values for all points are listed in Table~\ref{tab:cost-performance-settings} in Appendix~\ref{app:performance-time-budgets}.}
\label{fig:cost-performance-settings}
\end{figure*}

%% file: sec/06_training_progress_stability.tex
\section{Training Progress and Stability}
\label{sec:analysis}
\label{sec:training-failures}

Final performance does not establish whether learning was sustained or earlier gains were retained.
Figure~\ref{fig:validation-curves} presents all \PaperCompletedRuns{} runs across twelve settings, measured every five steps for \PaperCompletedValidations{} validation sweeps.
We summarize all four benchmarks using the area under each validation curve, normalized by the training-step horizon and averaged equally across the twelve settings.
Intermediate declines and peak-to-final gaps reveal aspects of stability that this aggregate measure and the final score can each miss.
Appendix~\ref{app:scienceworld-plateau} examines early stagnation on 2B ScienceWorld, where most methods share a low-score plateau with overlapping validation curves and little further improvement.

\noindent \textbf{Final rankings do not capture sustained performance.}
Across all twelve settings, PCSD has the highest mean final performance (\PaperTwelvePCSDMean{}), but its validation AUC of \PaperAllPCSDAUC{} is lower than OCSD's \PaperAllOCSDAUC{}.
OCSD instead ranks first by AUC and fourth by final performance (\PaperTwelveOCSDMean{}).
AgentOPSD shows a larger discrepancy: it ranks second in final performance (\PaperTwelveAgentOPSDMean{}) but ninth by AUC (\PaperAllAgentAUC{}), below GRPO's \PaperAllGRPOAUC{}.
A strong final model can therefore emerge from a substantially weaker training trajectory; final rankings alone do not characterize performance throughout optimization.
Appendix~\ref{app:learning-curves} compares aggregate trajectories, rankings, and stage-wise performance.

\noindent \textbf{Recovery can conceal severe intermediate collapse.}
On 4B ALFWorld, AgentOPSD reaches 60.0\% validation success at step 35, then scores no more than 4.3\% at each of the six evaluations from steps 45 to 70, before recovering to 79.3\% at step 100.
In the same setting, GRPO+OPSD reaches 39.3\% at step 20 but falls to zero at steps 25--40 and ends at 2.1\%.
Both methods undergo severe degradation, but only one ends with low performance.
An endpoint-only assessment would miss AgentOPSD's prolonged near-zero phase.
A strong final result therefore does not establish a stable training process.
Appendices~\ref{app:instability-cases} and~\ref{app:failure-evidence} provide trajectory comparisons and failure diagnostics, including decoding-specific failures.

\noindent \textbf{Late regression matters even without complete failure.}
Across the twelve settings, \PaperAllDropRuns{} runs finish at least 20 points below their observed validation peak.
OCSD on 4B ALFWorld reaches 85.0\% at step 60 but ends at 51.4\%; its final three validations average 50.5\%.
On 9B ScienceWorld, ICSD falls from a peak task score of 55.23 to 18.77, with its final three validations averaging 26.99.
In both cases, performance remains well below the peak over the final three validations, indicating sustained late regression rather than an isolated unfavorable final measurement.
Stable training therefore requires retaining acquired performance as well as avoiding collapse; a nonzero endpoint is insufficient when substantial gains have been lost.
Appendix~\ref{app:peak-endpoint} provides per-run summaries.

%% file: sec/07_cost_performance.tex
\section{The Cost of Performance Gains}
\label{sec:cost}

The practical value of a performance gain depends on the resources required to obtain it.
Equal training-step counts do not imply equal costs, since rollout lengths and teacher computation vary across methods.
For all eleven methods across the twelve model--task settings, Figure~\ref{fig:cost-performance-settings} pairs mean final performance over eight rollout seeds with measured training time.
All runs use eight GPUs, allowing us to examine both the cost of the final result and the performance attainable within GRPO's training-time budget.
Appendix~\ref{app:compute-breakdown} details the timing components, GPU-hour totals, and stage-level costs.

\noindent \textbf{Additional computation does not guarantee better performance.}
Of the 120 comparisons with GRPO across methods and settings, 112 require more net training time; 58 of these also achieve lower eight-seed mean performance, and seven tie GRPO.
Even PCSD and AgentOPSD, the only alternatives that improve mean final performance, require mean paired training-time ratios of 1.17 and 1.19 relative to GRPO.
Additional training effort is therefore not consistently rewarded by higher performance, and the methods that do improve must be assessed together with their overhead.
Measured cost is essential to determining whether a reported gain represents a practically useful improvement.

\noindent \textbf{Fixed-step advantages may disappear under a time budget.}
Using each GRPO run's net training time as the budget, we select the latest validation result reached by each method within that budget and compare it with its step-100 score.
Scores are averaged over the three model sizes within each benchmark.
On ALFWorld, AgentOPSD's mean success falls from 68.10\% at step 100 to 37.63\% within budget, reversing its advantage over GRPO's 58.10\%.
PCSD shows the same reversal on WebShop: 73.17\% at step 100 becomes 63.83\% within budget, below GRPO's 68.20\%.
These reversals show that an advantage at a fixed step count need not be available at a comparable computational cost.
Method comparisons should therefore assess not only the final gains, but also whether those gains can be achieved within the resources available to the baseline.
Appendix~\ref{app:performance-time-budgets} gives the time-budget rule and complete results.

%% file: sec/08_discussion.tex
\section{Discussion}
\label{sec:discussion}

The faster a field produces new methods, the more it needs a shared basis for judging progress.
The publication pace documented in our survey makes this need particularly pressing for agentic self-distillation.
Without direct, comprehensive comparisons, a growing collection of reported gains need not yield a clearer understanding of which ideas are worth building on.

This is why we chose to invest in independent evaluation rather than propose another training method.
We aim to clarify where methods improve on strong baselines, where gains break down, and what risks and costs they entail.
A method need not win every setting to be useful, but its value can only be understood when both its strengths and its limitations are visible.

We hope this benchmark helps researchers choose baselines, investigate unresolved failures, and pursue reliable, cost-effective improvements.
We therefore call for the community to value systematic evaluation alongside algorithmic innovation.
As the literature expands, the community's capacity for comparison and analysis must grow with it, so that new methods contribute to cumulative understanding and sustained progress.

%% file: sec/09_conclusion.tex
\section{Conclusion}
\label{sec:conclusion}

We benchmark ten representative self-distillation methods against GRPO across four agent benchmarks and three model sizes.
Gains over GRPO remain setting-dependent, with no method uniformly superior; strong final results can conceal unstable training, and additional computation does not consistently improve performance.
These findings show that progress must be assessed through the breadth of gains, the stability of learning, and the cost of improvement.
By documenting both gains and failures under a common protocol, our study helps researchers select strong baselines, identify unresolved training challenges, and pursue more reliable, cost-effective agent training.

%% file: sec/10_limitations.tex
\section*{Limitations}

Limited computational resources and researcher time prevent an exhaustive evaluation of the rapidly growing literature on agentic self-distillation.
We therefore compare ten representative methods with GRPO, covering different approaches to incorporating teacher guidance.
Although this selection does not encompass every relevant method, the observed setting-dependent gains, training instability, and performance--cost trade-offs substantiate the need for the comprehensive evaluation advocated in this work.
Extending the benchmark to additional methods would help establish how broadly these patterns hold.

%% file: sec/11_ethics.tex
\section*{Ethical Considerations}

This study compares existing agent-training methods using publicly available models and benchmark datasets.
Experiments are confined to simulated environments and retrieval-based question answering, without human-participant studies or deployment on behalf of real users.
We identify no significant ethical concerns specific to this study.

%% file: sec/12_appendix.tex
\section{Appendix Overview}
\label{app:overview}

\begin{itemize}
    \item \textbf{Appendix~\ref{app:method-scope}: Method Descriptions and Adaptations}
    \item \textbf{Appendix~\ref{app:training-settings}: Experimental Details}
    \begin{itemize}
        \item Appendix~\ref{app:training-configuration}: Training Configuration
        \item Appendix~\ref{app:teacher-contexts}: Teacher Models and Contexts
        \item Appendix~\ref{app:numerical-settings}: Numerical Precision and Stability
        \item Appendix~\ref{app:evaluation-settings}: Validation and Repeated Evaluation
        \item Appendix~\ref{app:reproducibility}: Experimental Records
    \end{itemize}
    \item \textbf{Appendix~\ref{app:implemented-algorithms}: Training Objectives and Algorithms}
    \begin{itemize}
        \item Appendix~\ref{app:alg-shared}: Shared Training Procedure
        \item Appendix~\ref{app:alg-auxiliary}: Auxiliary Distillation Objectives
        \item Appendix~\ref{app:alg-reweighting}: Advantage Reweighting
        \item Appendix~\ref{app:alg-continuation}: Verified Teacher Continuations
        \item Appendix~\ref{app:alg-reward-shaping}: Teacher-Based Reward Shaping
    \end{itemize}
    \item \textbf{Appendix~\ref{app:endpoint-moments}: Cross-Setting Gains and Regressions}
    \begin{itemize}
        \item Appendix~\ref{app:all-setting-results}: All Twelve Settings
        \item Appendix~\ref{app:low-performance}: Low-Performance Outcomes
    \end{itemize}
    \item \textbf{Appendix~\ref{app:logged-diagnostics}: Training Dynamics and Stability}
    \begin{itemize}
        \item Appendix~\ref{app:learning-curves}: Aggregate Progress and Method Rankings
        \item Appendix~\ref{app:instability-cases}: Intermediate Instability and Late Regression
        \item Appendix~\ref{app:scienceworld-plateau}: Early Stagnation on ScienceWorld
        \item Appendix~\ref{app:peak-endpoint}: Per-Run Training Profiles
        \item Appendix~\ref{app:failure-evidence}: Task Failure and Numerical Diagnostics
    \end{itemize}
    \item \textbf{Appendix~\ref{app:logged-compute}: Computational Cost Analysis}
    \begin{itemize}
        \item Appendix~\ref{app:performance-time-budgets}: Performance and Time Budgets
        \item Appendix~\ref{app:compute-breakdown}: Total Compute and Stage-Level Costs
    \end{itemize}
    \item \textbf{Appendix~\ref{app:literature-inventory}: Literature Survey}
    \begin{itemize}
        \item Appendix~\ref{app:survey-method}: Search Strategy and Selection Criteria
        \item Appendix~\ref{app:survey-statistics}: Publication Statistics
    \end{itemize}
\end{itemize}

\input{sec/13_method_scope}

\input{sec/14_training_settings}
\input{sec/15_implemented_algorithms}
\input{sec/16_cross_setting_results}
\input{sec/17_training_dynamics}
\input{sec/18_computational_cost}
\input{sec/19_literature_survey}

%% file: sec/13_method_scope.tex
\section{Method Descriptions and Adaptations}
\label{app:method-scope}

The eleven methods span outcome-based policy optimization, auxiliary distillation, token- or turn-level advantage reweighting, reward shaping, and verified teacher continuations.
Teachers use the same model backbone as the student, with privileged information available only during training.
Table~\ref{tab:method-scope} summarizes these mechanisms.

We implement the methods in a shared text-agent framework.
For Skill-SD and RLSD, we follow the SDAR repository's skill-bank adaptations, using the pre-update actor to score student responses under skill-conditioned prompts~\citep{lu2026sdar}.
GRPO+OPSD uses SDAR's constant-gate auxiliary objective, whose gradient is independent of teacher scores.
OVCSD collects additional, environment-verified teacher continuations.

For AgentOPSD and ADRS, we follow their released implementations of return-based statistics~\citep{wang2026agentopsd,zhang2026adrs}.
AgentOPSD initializes its belief from the clipped mean of trajectory-deduplicated training returns.
ADRS computes L2 teacher utility over interaction rows grouped by task.

Our ICSD implementation uses an additional gradient-free student-scoring pass to calibrate weights within each local actor minibatch; this computation is included in the reported training cost.
Following \citet{lv2026pcsd}, PCSD uses a frozen, skill-conditioned teacher initialized from the pretrained backbone; we reuse the fixed reference model's weights for this branch.
The complete objectives and update procedures appear in Appendix~\ref{app:implemented-algorithms}.

\input{table/method_scope}

%% file: table/method_scope.tex
\begin{table*}[tp]
\centering
\small
\begin{tabular}{@{}p{0.17\textwidth}p{0.31\textwidth}p{0.46\textwidth}@{}}
\toprule
Method & Supervision source & Use in optimization \\
\midrule
GRPO~\citep{shao2024deepseekmath} & Task rewards only. & \textbf{Policy advantage:} group-normalized returns drive the clipped policy-gradient objective. \\
GRPO+OPSD~\citep{lu2026sdar} & Skill-conditioned teacher scores. & \textbf{Auxiliary loss:} constant-weight reinforcement of sampled student tokens; teacher scores do not affect its gradient. \\
SDAR~\citep{lu2026sdar} & Skill-conditioned teacher scores. & \textbf{Auxiliary loss:} a detached teacher--student score-gap gate weights sampled-token supervision. \\
Skill-SD~\citep{wang2026skillsd} & Skill-conditioned teacher scores. & \textbf{Auxiliary loss:} policy-ratio-weighted K3 distillation on sampled tokens. \\
RLSD~\citep{yang2026rlsd} & Skill-conditioned teacher scores. & \textbf{Token advantages:} clipped, sign-aware weighting from teacher--student log-probability gaps. \\
OCSD~\citep{yang2026ocsd} & Paired scores with and without future observations. & \textbf{Token advantages:} reweight selected turns using the score difference induced by future observations. \\
OVCSD~\citep{xia2026ovcsd} & Skill-conditioned, environment-verified continuations and teacher token distributions. & \textbf{Comparative advantages and suffix loss:} contrast state-aligned student/teacher actions and distill verified suffixes with top-$K$-plus-tail KL. \\
AgentOPSD~\citep{wang2026agentopsd} & Skill-conditioned teacher scores and group task returns. & \textbf{Turn advantages:} recursive success-belief increments yield bounded, sign-aware reweighting across a trajectory. \\
ADRS~\citep{zhang2026adrs} & Skill-conditioned teacher scores and task returns. & \textbf{Reward and advantage shaping:} gate teacher rewards by return-based utility, then recompute advantages and add a token-level correction. \\
ICSD~\citep{lan2026icsd} & Skill-conditioned teacher scores. & \textbf{Auxiliary loss:} calibrate teacher trust and advantage-dependent influence; redistribute token weights while preserving their per-response total. \\
PCSD~\citep{lv2026pcsd} & Skill-conditioned scores from a frozen reference model. & \textbf{Auxiliary loss:} gate supervision using variance-adaptive window averages and declining trends in the teacher--student score gap. \\
\bottomrule
\end{tabular}
\caption{Supervision sources and their use in the evaluated implementations. All methods use task rewards; teacher scores denote log-probabilities of student-generated tokens.}
\label{tab:method-scope}
\end{table*}

%% file: sec/14_training_settings.tex
\section{Experimental Details}
\label{app:training-settings}

\subsection{Training Configuration}
\label{app:training-configuration}

We train each of the eleven methods on ALFWorld, ScienceWorld, Search-QA, and WebShop with Qwen3.5-2B, 4B, and 9B, for \PaperCompletedRuns{} runs of 100 global steps each.
We combine shared optimization defaults with benchmark- and method-specific configurations drawn from the source papers and released code.
The sources of these configurations and the settings for our benchmark extensions are described below; method-specific algorithm hyperparameters remain fixed across the three model sizes within each benchmark.
Tables~\ref{tab:training-common}, \ref{tab:training-tasks}, and~\ref{tab:training-methods} summarize common settings, benchmark-specific settings and overrides, and algorithm hyperparameters, respectively, across all four benchmarks.

\input{table/training_common}

\noindent \textbf{Configuration sources.}
The base learning rate, rollout group size, and default reference KL coefficient follow Appendix~C of SDAR~\citep{lu2026sdar}.
For ALFWorld, WebShop, and Search-QA, prompt and response limits, interaction horizons, history windows, PPO minibatch sizes, and invalid-action penalties follow its released task-specific configurations.
ALFWorld and WebShop retain outcome rewards of 10 for success and 0 otherwise; Search-QA uses 1 for a correct answer and 0 otherwise, together with SDAR's top-three-passage retrieval configuration.
ScienceWorld uses the AgentGym task split~\citep{xi2025agentgym}, with 200 evaluation tasks and 2,059 training instances after removing 61 overlapping task IDs; its episode reward is the simulator's final task score divided by 100.
These rewards are assigned to interaction rows, with the invalid-action penalty applied to the corresponding row.
Table~\ref{tab:training-tasks} summarizes these benchmark configurations and their method-specific overrides.

\input{table/training_tasks}

\noindent \textbf{Baseline adaptations.}
SDAR and GRPO+OPSD retain SDAR's reported auxiliary coefficients and gate settings.
For Skill-SD and RLSD, we use the skill-bank adaptations in the SDAR repository: Skill-SD keeps a constant auxiliary weight of $0.001$, while RLSD follows the released Qwen3 scripts' 100-step linear decay with initial scale $\lambda_0=0.5$ and multiplier clipping of $0.2$.

\noindent \textbf{Method-specific settings from the source papers.}
AgentOPSD's algorithm hyperparameters follow Appendix~F of \citet{wang2026agentopsd}, including its upper PPO clipping bound of $0.24$.
OCSD's selection ratio and reweighting coefficient follow Table~4 of \citet{yang2026ocsd}; ICSD's calibration parameters follow Appendix~C.1 of \citet{lan2026icsd}; PCSD's window and gating parameters follow Table~5 of \citet{lv2026pcsd}.
OCSD's Search-QA warmup ratio of $0.1$ and reference KL coefficient of $0.001$ follow its released training configuration.
The algorithm-specific values are summarized in Table~\ref{tab:training-methods}.

\noindent \textbf{Benchmark-specific ADRS settings.}
We follow the primary GRPO/L2 configurations in Appendix~A.5 of \citet{zhang2026adrs}: ALFWorld uses $\eta=0.1$, batch-wide score normalization, and gate-input normalization; WebShop uses $\eta=0.02$, per-response score normalization, and no gate-input normalization.
The same source specifies ALFWorld's entropy coefficient of $0.01$ and upper PPO clipping bound of $0.28$.
For Search-QA, the teacher-reward scale $\eta=0.05$, batch-wide score normalization, and reference KL coefficient of $0.001$ likewise follow the reported Search-QA configuration.
These task-dependent choices preserve the source configurations; the paper's normalization ablation supports batch-wide scaling on ALFWorld and Search-QA, and per-response scaling on WebShop.
For the ScienceWorld extension, we adopt the Search-QA teacher-reward scale and normalization settings, including gate-input normalization, while retaining the common optimization defaults.
The benchmark-specific settings are listed in Table~\ref{tab:training-tasks}.

\noindent \textbf{Study-specific choices.}
The Qwen3.5 backbones, 100-step training horizon, eight-H800 execution, and fixed evaluation sets are specified by our benchmark.
Task batches retain SDAR's size of 16 for ALFWorld and WebShop; we use 16 for ScienceWorld and 64 for Search-QA.
Within each benchmark, task batches, environment settings, and evaluation sets are shared by all methods and model sizes (Table~\ref{tab:training-tasks}).
We follow OVCSD's outcome-verified intervention and suffix-distillation framework~\citep{xia2026ovcsd}.
For our implementation, we fix the continuation budget at two attempts per eligible group, the teacher-distribution approximation at top-32 plus tail, and the suffix-distillation weight at $0.1$ across all twelve settings.

\input{table/training_methods}

\noindent \textbf{Trajectory collection and updates.}
Each ALFWorld/WebShop/ScienceWorld global step samples 16 task instances and eight student trajectories per instance, giving 128 student episodes.
Search-QA instead samples 64 questions and eight trajectories each, giving 512 nominal student episodes per step; equal step counts therefore do not imply equal episode budgets across benchmarks.
Trajectories use separate environment replicas and are split into interaction turns for policy optimization.
The actor minibatch sizes are 256 turns for ALFWorld and Search-QA, and 64 for WebShop and ScienceWorld; one PPO epoch may therefore contain several optimizer updates.
OVCSD additionally allows up to two teacher continuation attempts per eligible task group, including prefix replay.

\subsection{Teacher Models and Contexts}
\label{app:teacher-contexts}

SDAR, GRPO+OPSD, Skill-SD, RLSD, AgentOPSD, ADRS, and ICSD score student responses with the pre-update actor conditioned on skills.
PCSD uses the fixed reference model as its teacher.
Following SDAR's skill-conditioning procedure~\citep{lu2026sdar}, we use benchmark-specific skill banks that combine general guidance with task-specific instructions selected from ALFWorld game identifiers or matching prompt keywords.
We prepend this context to the student prompt, retokenize, and apply left truncation at the task's prompt-length limit while keeping the student response unchanged.
OVCSD uses the same skill-conditioned input for continuations; OCSD instead compares replay contexts with and without future observations.

\subsection{Numerical Precision and Stability}
\label{app:numerical-settings}

Training uses FP32 actor weights with BF16 mixed-precision computation and FP32 reductions and buffers; reference and rollout weights use BF16.
Final-model evaluation uses BF16 weights for both the actor and inference engine.

\noindent \textbf{Numerical stability.}
We use $\epsilon_{\mathrm{num}}=10^{-8}$ for masked means and ADRS statistics, and $\epsilon_{\mathrm{norm}}=10^{-6}$ for GRPO and AgentOPSD standardization.
AgentOPSD clips prior and updated belief probabilities with $\epsilon_{\mathrm{prior}}=10^{-4}$ and $\epsilon_{\mathrm{bel}}=10^{-6}$, respectively.
OCSD selects turns with absolute advantage above $\epsilon_A=10^{-6}$.
OVCSD uses $\epsilon_R=10^{-6}$ as the upper threshold on raw episode-return standard deviation for intervention; ADRS uses the same value as the lower threshold on shaped row-return standard deviation for its additional advantage correction.
ICSD uses a calibration scale floor of $\epsilon_{\mathrm{cal}}=10^{-4}$; PCSD uses $\epsilon_{\mathrm{pc}}=10^{-8}$ for window statistics and slope normalization.
The reference K3 estimator is clipped to $[-c_{\mathrm{ref}},c_{\mathrm{ref}}]$, with $c_{\mathrm{ref}}=10$.

\subsection{Validation and Repeated Evaluation}
\label{app:evaluation-settings}

Training validation on ALFWorld, WebShop, and Search-QA follows SDAR's released settings~\citep{lu2026sdar}; ScienceWorld uses the same stochastic decoding rule as ALFWorld and WebShop.
Within each benchmark, all eleven methods use the validation configuration in Table~\ref{tab:training-tasks}.
Repeated final evaluation uses the same $T=0.4$ sampling protocol on all four benchmarks, with eight rollout seeds (0--7) indexed by task and interaction turn.
The 1,056 evaluations of \PaperCompletedRuns{} trained models and 96 evaluations of pretrained backbones total \PaperCompletedEvalSweeps{} evaluation sweeps over twelve settings.
Across all four benchmarks, validation and final evaluation use fixed task sets shared across methods, model sizes, and evaluation sweeps, retaining every instance, including incomplete final batches.
Overall performance averages the per-instance metric over the full evaluation set; task-category and source-specific results use the corresponding subsets.
This aggregation gives equal weight to instances rather than to batches or categories.

\subsection{Experimental Records}
\label{app:reproducibility}

For all four benchmarks, we retain training configurations, per-step metrics, and repeated-evaluation summaries for the trained and pretrained models.
The training records cover \PaperCompletedRuns{} runs and \PaperCompletedValidations{} validation sweeps, with each validation result linked to its original log.
Learning curves and computational costs are derived from these logs; final-evaluation means and sample standard deviations are taken from the repeated-evaluation summaries.
The analysis scripts generate the tables and figures from these records and verify metric aggregation against the source data.

%% file: table/training_common.tex
\begin{table*}[t]
\centering\small
\begin{tabular}{@{}p{0.42\textwidth}p{0.53\textwidth}@{}}
\toprule
Parameter & Configuration \\
\midrule
\multicolumn{2}{@{}l}{\textbf{Optimization}} \\
Trainable parameters & Full-parameter training \\
Optimizer & AdamW; $(\beta_1,\beta_2)=(0.9,0.999)$; weight decay $0.01$ \\
Learning rate & $10^{-6}$; constant after any warmup \\
Warmup ratio & $0$, except as listed in Table~\ref{tab:training-tasks} \\
Gradient norm cap $c_{\mathrm{grad}}$ & $1.0$ \\
PPO epochs and update microbatch & $N_{\mathrm{ep}}=1$; 1 interaction turn per GPU; fixed-size microbatches \\
Log-probability microbatch & 1 interaction turn per GPU for both actor and reference \\
Lower PPO clipping and dual clipping & $\epsilon_l=0.2$; $c_{\mathrm{dual}}=3.0$ \\
Upper PPO clipping $\epsilon_h$ & $0.2$, except as listed in Table~\ref{tab:training-tasks} \\
Reference policy & Frozen at the pretrained initialization \\
Reference KL regularization & Sampled-token K3 penalty added to the policy loss \\
Reference KL coefficient $c_{\mathrm{KL}}$ & $0.01$, except as listed in Table~\ref{tab:training-tasks} \\
Entropy coefficient $c_H$ & $0.001$, except as listed in Table~\ref{tab:training-tasks} \\
\midrule
\multicolumn{2}{@{}l}{\textbf{Trajectory Collection and Evaluation}} \\
Student trajectories per task & $G=8$ \\
Training sampling & Stochastic; temperature $\tau_{\mathrm{gen}}=1.0$ \\
Sampling cutoffs & Top-$p=1.0$; no top-$k$ restriction when sampling \\
Repeated final evaluation & Temperature $0.4$; eight rollout seeds (0--7) \\
Response length limit & 512 tokens per interaction turn \\
Training steps $S$ & 100 \\
Validation schedule & Before training and every 5 steps \\
\midrule
\multicolumn{2}{@{}l}{\textbf{Hardware and Execution}} \\
Training hardware & 1 node with 8 NVIDIA H800 GPUs \\
CPU allocation (Ray) & 64 logical CPUs per run \\
Rollout tensor parallelism & 1 \\
Distributed training & FSDP; activation checkpointing enabled \\
CPU offloading & Reference parameters only; actor parameters and optimizer states remain on GPU \\
Rollout inference & BF16; chunked prefill; Triton Gated DeltaNet prefill \\
Rollout execution & CUDA graphs with eager fallback \\
Thinking-mode chat formatting & Disabled \\
\bottomrule
\end{tabular}
\caption{Common training and evaluation settings for the eleven methods across all four benchmarks. Benchmark-specific settings and overrides appear in Table~\ref{tab:training-tasks}; algorithm hyperparameters appear in Table~\ref{tab:training-methods}.}
\label{tab:training-common}
\end{table*}

%% file: table/training_tasks.tex
\begin{table*}[t]
\centering\small
\setlength{\tabcolsep}{4pt}
\begin{tabular}{@{}>{\raggedright\arraybackslash}p{0.36\textwidth}*{4}{>{\raggedright\arraybackslash}p{0.14\textwidth}}@{}}
\toprule
Parameter & ALFWorld & ScienceWorld & Search-QA & WebShop \\
\midrule
\multicolumn{5}{@{}l}{\textbf{Benchmark Settings}} \\
Task batch size $B$ & 16 & 16 & 64 & 16 \\
Prompt limit (tokens per turn) & 2048 & 4096 & 4096 & 4096 \\
Maximum interaction turns $H$ & 50 & 20 & 4 & 15 \\
Retained history (turns) & 2 & 2 & 4 & 2 \\
PPO minibatch (turns, all GPUs) & 256 & 64 & 256 & 64 \\
Episode reward & Success: 10; otherwise: 0 & Final task score / 100 & Correct: 1; otherwise: 0 & Success: 10; otherwise: 0 \\
Invalid-action penalty $c_{\mathrm{inv}}$ & 0.1 & 0.1 & 0.01 & 0.1 \\
Retrieved passages per query & N/A & N/A & 3 & N/A \\
Retrieval timeout (s) & N/A & N/A & 60 & N/A \\
Validation set & All 140 \texttt{valid\_seen} games & 200 fixed tasks & 3125 fixed questions & 128 fixed goals \\
Validation batch size (tasks) & 140 & 32 & 512 & 128 \\
Validation decoding & Sampling, $T=0.4$ & Sampling, $T=0.4$ & Greedy, $T=0$ & Sampling, $T=0.4$ \\
\midrule
\multicolumn{5}{@{}l}{\textbf{Method-Specific Optimization Overrides}} \\
AgentOPSD: upper PPO clipping $\epsilon_h$ & 0.24 & 0.24 & 0.24 & 0.24 \\
ADRS: upper PPO clipping $\epsilon_h$ & 0.28 & 0.2 & 0.2 & 0.2 \\
ADRS: entropy coefficient $c_H$ & 0.01 & 0.001 & 0.001 & 0.001 \\
OCSD: warmup ratio & 0 & 0 & 0.1 & 0 \\
OCSD: reference KL $c_{\mathrm{KL}}$ & 0.01 & 0.01 & 0.001 & 0.01 \\
ADRS: reference KL $c_{\mathrm{KL}}$ & 0.01 & 0.01 & 0.001 & 0.01 \\
\midrule
\multicolumn{5}{@{}l}{\textbf{Benchmark-Specific ADRS Parameters}} \\
Teacher-reward scale $\eta$ & 0.1 & 0.05 & 0.05 & 0.02 \\
Teacher-signal normalization & Batch-wide SD & Batch-wide SD & Batch-wide SD & Per-response SD \\
Scale gate input by its SD & Enabled & Enabled & Enabled & Disabled \\
\bottomrule
\end{tabular}
\caption{Benchmark-specific settings and method-specific overrides across four benchmarks, shared across the three model sizes. Configuration sources and benchmark extensions are described in Appendix~\ref{app:training-configuration}. Unlisted optimization settings retain the defaults in Table~\ref{tab:training-common}. SD denotes standard deviation.}
\label{tab:training-tasks}
\end{table*}

%% file: table/training_methods.tex
\begin{table*}[t]
\centering\small
\begin{tabular}{@{}>{\raggedright\arraybackslash}p{0.15\textwidth}>{\raggedright\arraybackslash}p{0.25\textwidth}>{\raggedright\arraybackslash}p{0.54\textwidth}@{}}
\toprule
Method & Parameter group & Configuration \\
\midrule
GRPO & \multicolumn{2}{l}{No additional hyperparameters} \\
\addlinespace[2pt]
GRPO+OPSD & Auxiliary weight and gate & $\lambda_{\mathrm{sd}}=0.01$; $\beta=0$ (constant gate $1/2$) \\
\addlinespace[2pt]
SDAR & Auxiliary weight and gate & $\lambda_{\mathrm{sd}}=0.01$; sigmoid slope $\beta=5$; detached gate \\
\addlinespace[2pt]
Skill-SD & Auxiliary weight & $\lambda_{\mathrm{dist}}=0.001$, constant throughout training \\
& Exponent caps & $c_{\mathrm{K3}}=20$; $c_{\mathrm{IS}}=10$ \\
\addlinespace[2pt]
RLSD & Mixing schedule & $\lambda_s=\lambda_0(1-s/S)$; $\lambda_0=0.5$, $S=100$ \\
& Multiplier clipping & $\delta_{\mathrm{r}}=0.2$ \\
\addlinespace[2pt]
OCSD & Future context & $L_f=2$ future interaction turns \\
& Turn selection & Highest old-policy mean NLL among eligible turns within each trajectory; fraction $\alpha_{\mathrm{sel}}=0.2$ \\
& Reweighting & Residual scale $\lambda_{\mathrm{oc}}=0.5$; advantage threshold $\epsilon_A=10^{-6}$ \\
\addlinespace[2pt]
OVCSD & Eligibility and budget & Reward standard deviation threshold $\epsilon_R=10^{-6}$; $A_{\max}=2$ attempts per eligible group \\
& Suffix distillation & Top-$K$ plus tail with $K=32$; KL weight $\lambda_{\mathrm{suf}}=0.1$ \\
\addlinespace[2pt]
AgentOPSD & Advantage multiplier & Signed, turn-level; mixing $\lambda_{\mathrm{ag}}=0.5$; band $\delta_{\mathrm{ag}}=0.2$; no coefficient decay \\
& Belief update & Decay $\gamma=0.95$; prior from the clipped mean of trajectory-deduplicated training returns; fallback $0.5$ \\
\addlinespace[2pt]
ADRS & Teacher-utility gate & L2 teacher value advantage (TVA) over task-grouped interaction rows; alignment temperature $\tau_{\mathrm{align}}=1$; sigmoid slope $\beta_{\mathrm{TVA}}=2$ \\
& Reward and advantage shaping & Per-turn centering; additive correction clipped at $\delta_{\mathrm{ad}}=0.2$, sign $+1$ \\
& Schedule & Active throughout training, without warmup \\
\addlinespace[2pt]
ICSD & Auxiliary weight & $\lambda_{\mathrm{ic}}=0.01$ \\
& Calibration & CDF splits $\xi_{\mathrm{trust}}=0.4$, $\xi_{\mathrm{inf}}=0.5$; scale floor $\epsilon_{\mathrm{cal}}=10^{-4}$; minimum pool $n_{\mathrm{cal}}=8$ tokens \\
\addlinespace[2pt]
PCSD & Auxiliary weight & $\lambda_{\mathrm{pc}}=0.01$ \\
& Token windows & $N_{\min}=1$, $N_{\max}=8$; decay $\rho_{\mathrm{pc}}=0.8$ \\
& Variance and trend gate & Variance bounds $v_l=0.05$, $v_h=0.5$; trend strength $\gamma_{\mathrm{pc}}=0.3$; slope $\beta_{\mathrm{pc}}=5$ \\
\bottomrule
\end{tabular}
\caption{Algorithm-specific hyperparameters shared across all four benchmarks and three model sizes. Sources and implementation choices are described in Appendix~\ref{app:training-configuration}; benchmark-dependent values appear in Table~\ref{tab:training-tasks}. Symbols follow Appendix~\ref{app:implemented-algorithms}; $s$ denotes the global training step.}
\label{tab:training-methods}
\end{table*}

%% file: sec/15_implemented_algorithms.tex
\raggedbottom
\section{Training Objectives and Algorithms}
\label{app:implemented-algorithms}

Algorithms~\ref{alg:shared}--\ref{alg:adrs} specify the eleven methods as evaluated in this study.
The ten methods beyond GRPO modify its auxiliary objective, advantage weights, collected trajectories, or reward signal.
We define the shared operations once and expose the method-specific changes below.
For more detailed method descriptions and derivations, we refer readers to the original papers cited in Table~\ref{tab:method-scope}.

\noindent \textbf{Data and indices.}
A trajectory is one task attempt, comprising multiple model--environment interactions.
The training batch $D$ contains one row $b$ per active interaction turn; $t$ indexes tokens in that turn's response.
The binary mask $M_{bt}$ is one for valid response tokens and zero otherwise.
The maps $i(b)$ and $g(b)$ identify the trajectory and task group of row $b$; a task group contains all sampled attempts at the same task.
$B_i$ and $B_g$ denote their respective row sets, and $g(i)$ identifies the task group of trajectory $i$.

\noindent \textbf{Models and token scores.}
$\theta_0$ denotes the initial model parameters, $\theta$ the trainable student, and $\pi_\theta$ its token distribution.
The old policy $\theta_{\mathrm{old}}$ is refreshed at each training step and held fixed during the ensuing actor updates; the reference $\phi$ stays at $\theta_0$ throughout training.
For each sampled response token, $p_{bt}$, $o_{bt}$, and $f_{bt}$ are its log-probabilities under the current student, old policy, and reference, respectively.
The teacher score $q_{bt}$ is the log-probability of that same token given additional training-time context, such as skills or future observations.
These are sampled-token scores, not full-vocabulary distributions, unless stated otherwise.

\noindent \textbf{Operations and gradients.}
$\operatorname{sg}(x)$ keeps the value of $x$ but blocks its gradient.
Teacher, old-policy, and reference scores, rewards, selection decisions, and constructed advantages are detached; gradients flow through the current student scores $p$.
Products and divisions of token arrays are elementwise, and row quantities broadcast over tokens.
We use the sigmoid $\sigma(x)=(1+e^{-x})^{-1}$, log-odds $\operatorname{logit}(x)=\log[x/(1-x)]$, and $\operatorname{ReLU}(x)=\max(x,0)$.
$\operatorname{clip}(x,a,b)$ restricts $x$ to $[a,b]$; $C_\epsilon(x)=\operatorname{clip}(x,\epsilon,1-\epsilon)$ keeps beliefs away from zero and one.
Other intermediate quantities are local to each method. Subscripted $\epsilon$ terms denote numerical stabilizers, clipping widths, or thresholds, as specified below; hyperparameter values are listed in Appendix~\ref{app:training-settings}.

\subsection{Shared Training Procedure}
\label{app:alg-shared}

The shared loop in Algorithm~\ref{alg:shared} selects a task batch $X$ of size $B$ and collects $G$ trajectories per task, forming $\mathcal T$.
It runs for $S$ training steps, indexed by $s$, with at most $H$ interaction turns per trajectory and sampling temperature $\tau_{\mathrm{gen}}$.
$\mathcal M$ denotes the selected method, not a token mask.

\noindent \textbf{Rewards and advantages.}
We assign the complete episode return to the final response token of \emph{each interaction turn}.
The invalid-action penalty subtracts $c_{\mathrm{inv}}$ at that token for an invalid turn.
Let $r_{bt}$ be this sparse reward and $R_b=\sum_t r_{bt}$.
$A_b$ is the scalar GRPO advantage of row $b$, broadcast over its valid tokens as $A_{bt}=A_bM_{bt}$.
Group normalization pools the row returns within each task group:
\begin{equation}
 A_{bt}=\frac{R_b-\mu_{g(b)}}{s_{g(b)}+\epsilon_{\mathrm{norm}}}M_{bt}.
 \label{eq:code-grpo}
\end{equation}
Here $\mu_g$ and $s_g$ are the mean and sample standard deviation (SD) of $\{R_b:b\in B_g\}$; each trajectory contributes one observation per turn.
For a singleton group, we set $\mu_g=0,s_g=1$.
We denote this operation by $\mathcal G(r)$ and retain groups with constant rewards; $\widehat A$ denotes any method-specific replacement for $A$.

\noindent \textbf{Trajectory collection.}
$\operatorname{SampleTasks}$ draws training tasks, $\operatorname{Rollout}$ generates skill-free trajectories, and $\operatorname{TurnRows}$ unfolds them into $D$, retaining prompts, responses, masks, and task, trajectory, and turn IDs.
$\operatorname{Reward}$ assigns the rewards defined above.
$\operatorname{Score}(\theta,D)$ returns detached log-probabilities for the responses already in $D$; it does not generate new responses.
The batch $D$ carries the rewards $r$, row returns $R$, and scores $o,f$ computed in the shared loop.
$\operatorname{MethodStep}$ applies the selected algorithm to these quantities, updating $\theta$ while keeping $\theta_{\mathrm{old}}$ and $\phi$ fixed.
For OVCSD it calls $\operatorname{OVCSDUpdate}$ with both the eligible groups $\mathcal F$ and verified pairs $\mathcal V$ returned during collection.

\noindent \textbf{Common objective.}
The policy-gradient mask $M^{\mathrm{PG}}$ selects tokens for the policy loss; the regularization mask $M^{\mathrm{reg}}$ selects tokens for reference-KL and entropy terms.
Both are binary: one includes a token and zero excludes it from the corresponding term.
Except for OVCSD, whose masks are specified below, $M^{\mathrm{PG}}=M^{\mathrm{reg}}=M$.
For any such mask, $\langle X\rangle_M=\sum_{b,t}M_{bt}X_{bt}/(\sum_{b,t}M_{bt}+\epsilon_{\mathrm{num}})$ denotes its token average within a microbatch.

The current-to-old probability ratio is $\rho=\exp(p-o)$.
Let $u$ be the usual clipped surrogate, with lower and upper clipping widths $\epsilon_l,\epsilon_h$.
Dual clipping additionally bounds its negative-advantage branch using $c_{\mathrm{dual}}>1$:
\begin{align}
 u &= \max\{-A\rho,\,
       -A\,\operatorname{clip}(\rho,1-\epsilon_l,1+\epsilon_h)\}, \nonumber\\
 \ell_{\mathrm{PG}} &=
 \begin{cases}
 \min(u,-c_{\mathrm{dual}}A), & A<0,\\
 u, & A\geq 0.
 \end{cases} \label{eq:code-pg}
\end{align}
The common loss combines this policy-gradient term $\ell_{\mathrm{PG}}$, a reference penalty $k_{\mathrm{ref}}$ weighted by $c_{\mathrm{KL}}$, and token-distribution entropy $\mathcal H(\pi_\theta)$ weighted by $c_H$.
The constant $c_{\mathrm{ref}}$ caps the sampled-token KL penalty:
\begin{align}
 \mathcal L_0(A)&=\langle\ell_{\mathrm{PG}}\rangle_{M^{\mathrm{PG}}}
       +c_{\mathrm{KL}}\langle k_{\mathrm{ref}}\rangle_{M^{\mathrm{reg}}} \nonumber\\
       &\quad-c_H\langle\mathcal H(\pi_\theta)\rangle_{M^{\mathrm{reg}}}, \label{eq:code-base}\\
 k_{\mathrm{ref}}&=\operatorname{clip}
       \bigl(e^{f-p}-(f-p)-1,-c_{\mathrm{ref}},c_{\mathrm{ref}}\bigr). \nonumber
\end{align}
Here $k_{\mathrm{ref}}$ is the clipped sampled-token K3 estimator of reference KL; $\mathcal H$ denotes entropy, whereas $H$ is the interaction horizon.
The actor accumulates microbatch losses, clips the gradient norm at $c_{\mathrm{grad}}$, and skips an optimizer update if the norm is nonfinite.
$\operatorname{Update}_D(A,L_{\mathrm{aux}})$ minimizes $\mathcal L_0(A)+L_{\mathrm{aux}}$ on $D$ using AdamW over $N_{\mathrm{ep}}$ PPO epochs.
The auxiliary loss $L_{\mathrm{aux}}$ is recomputed from the current student at each forward pass.
GRPO (Algorithm~\ref{alg:grpo}) sets this term to zero, retaining the reference-KL and entropy terms.
The update changes $\theta$ in place; minibatching and gradient accumulation follow the common settings.

\noindent \textbf{Skill-conditioned scoring.}
$\operatorname{SkillReplay}(D,\psi)$ returns teacher scores $q$: it augments prompts with retrieved skills, keeps the sampled student responses, and scores them without gradients under model parameters $\psi$.
Its outputs remain fixed throughout the subsequent actor update. Most skill-replay methods use the pre-update actor, $\psi=\theta_{\mathrm{old}}$; PCSD instead uses the fixed reference, $\psi=\phi$.
Teacher prompts are left-truncated as described in Appendix~\ref{app:training-settings}.

\input{algorithm/shared}
\input{algorithm/grpo}

\subsection{Auxiliary Distillation Objectives}
\label{app:alg-auxiliary}
\noindent \textbf{GRPO+OPSD's constant-gate objective.}
GRPO+OPSD and SDAR weight their auxiliary loss by $\lambda_{\mathrm{sd}}$ and control the sigmoid gate's slope with $\beta$.
In our comparison, GRPO+OPSD (Algorithm~\ref{alg:grpo-opsd}) sets $\beta=0$, yielding a detached gate of $1/2$.
For finite log-probabilities its auxiliary gradient is therefore exactly that of $-(\lambda_{\mathrm{sd}}/2)\langle p\rangle_M$; teacher values affect the loss value but not this gradient.
The common reference penalty in Eq.~\ref{eq:code-base} is present in both GRPO and GRPO+OPSD and is separate from this auxiliary term.

\input{algorithm/grpo_opsd}

\noindent \textbf{SDAR and Skill-SD supervision.}
In SDAR (Algorithm~\ref{alg:sdar}), the gate $g(p)$ weights each valid response token according to the teacher--current-student log-probability gap, including tokens in unsuccessful trajectories; the gate is detached.
Skill-SD (Algorithm~\ref{alg:skillsd}) instead uses the log-probability difference $d(p)=p-q$ to construct a K3 term $k(p)$, weighted by the importance ratio $w(p)$ and auxiliary coefficient $\lambda_{\mathrm{dist}}$.
The constants $c_{\mathrm{K3}}$ and $c_{\mathrm{IS}}$ cap the two exponent inputs.
Gradients pass through both $k(p)$ and $w(p)$, but not through teacher or old-policy scores.
Both objectives supervise the sampled response tokens.

\input{algorithm/sdar}
\input{algorithm/skillsd}

\noindent \textbf{ICSD calibration and gradient boundaries.}
ICSD (Algorithm~\ref{alg:icsd}) first scores the current student on a local actor minibatch $D'$ to obtain detached scores $p^\star$.
Its teacher gap $\Delta=q-p^\star$ and influence score $u=A\exp(p^\star-o)\Delta$ are calibrated into trust weights $g$ and influence weights $h$.
$\operatorname{CalCDF}_{\xi}$ performs this calibration using a cumulative distribution function (CDF), pooling valid tokens with the same interaction-turn index within $D'$.
For each pool it fits location $m$ and scales $a_-,a_+$, then applies
\[
 F_\xi(x)=
 \begin{cases}
 \xi e^{(x-m)/a_-}, & x<m,\\
 1-(1-\xi)e^{-(x-m)/a_+}, & x\geq m.
 \end{cases}
\]
The split $\xi\in(0,1)$ sets $F_\xi(m)=\xi$; $\xi_{\mathrm{trust}}$ and $\xi_{\mathrm{inf}}$ specify the trust and influence calibrations.
The location is the lower sample median; the scales are mean absolute deviations below and at or above $m$, floored at $\epsilon_{\mathrm{cal}}$ (also used for an empty side).
Pools with fewer than $n_{\mathrm{cal}}$ tokens or nonfinite values fall back to all valid minibatch tokens, then to location zero and unit scales if still invalid.
The final token weights $c$ combine $g$ and $h$ within the eligible-token set $C_b$, preserving each response's total trust weight; other tokens retain $g$.
The normalizer $z_b$ is the sum of $gh$ over $C_b$.
The coefficient $\lambda_{\mathrm{ic}}$ scales the auxiliary loss, and $c$ remains fixed during that minibatch's gradient computation.
$\operatorname{UpdateMiniBatch}_{D'}$ applies one accumulated-gradient update on $D'$; it does not repeat the outer PPO epochs.

\input{algorithm/icsd}

\noindent \textbf{PCSD windows and teacher.}
PCSD (Algorithm~\ref{alg:pcsd}) smooths the detached teacher gap $\Delta=q-p$ over short and long forward windows of lengths $N_{\min}$ and $N_{\max}$.
All windows start at the current token and stay within its response row, excluding padding via $M$.
$\operatorname{DecayMean}$ weights offset $j$ by $\rho_{\mathrm{pc}}^j$, yielding means $m_s,m_l$.
The long-window population variance $v$, computed by $\operatorname{WindowVar}$, sets the interpolation weight $a$ between variance bounds $v_l,v_h$; $\bar\Delta$ is the resulting smoothed gap.
$\operatorname{WindowSlope}$ fits an unweighted least-squares slope against token offsets, adds $\epsilon_{\mathrm{pc}}$ to its denominator, and returns zero for fewer than two valid tokens.
Its slope $\ell$ is scaled by the row's mean absolute gap $d_b$; the factor $h$ suppresses declining support with strength $\gamma_{\mathrm{pc}}$.
The final detached weight $w$ combines $h$ with a sigmoid of slope $\beta_{\mathrm{pc}}$, and $\lambda_{\mathrm{pc}}$ scales the auxiliary loss.
The skill-conditioned teacher uses fixed reference weights throughout training; only the current student receives gradients.

\input{algorithm/pcsd}

\subsection{Advantage Reweighting}
\label{app:alg-reweighting}
RLSD, OCSD, and AgentOPSD replace $A$ by detached advantages $\widehat A$ before the common update.
Their multipliers preserve advantage signs and leave zero advantages unchanged under the evaluated coefficient ranges.

\noindent \textbf{RLSD mixing.}
RLSD (Algorithm~\ref{alg:rlsd}) mixes the original advantage with a teacher-reweighted version using $\lambda_s$; $\lambda_0$ sets the schedule's initial scale and $\delta_{\mathrm r}$ bounds the token multiplier $w$ around one.
For the one-based step $s$, $\lambda_s=\lambda_0\max(1-s/S,0)$, so it is $\lambda_0(1-1/S)$ at the first step and zero at the last; teacher scores are still computed at that endpoint.

\input{algorithm/rlsd}

\noindent \textbf{OCSD evidence selection.}
In OCSD (Algorithm~\ref{alg:ocsd}), $F_b$ counts the future observations available to row $b$, and $\operatorname{Top}_k(E;n)$ selects the $k$ rows in $E$ with the largest mean negative log-likelihood (NLL) $n$.
For trajectory $i$, $E_i$ contains eligible turns with future observations and $|A_b|>\epsilon_A$; $J_i$ contains the highest-NLL fraction $\alpha_{\mathrm{sel}}$, with at least one turn if $E_i$ is nonempty.
$\operatorname{EvidencePair}$ keeps the sampled response and constructs full/ablated prompts from up to $L_f$ future observations: intermediate action schemas are identical, but observations are withheld in the ablated prompt; both prompts are left-truncated.
$q^{\mathrm{full}}$ and $q^{\mathrm{abl}}$ score these two contexts; their bounded contrast $z$ modulates advantages with strength $\lambda_{\mathrm{oc}}$.

\input{algorithm/ocsd}

\noindent \textbf{AgentOPSD belief updates.}
In AgentOPSD (Algorithm~\ref{alg:agentopsd}), $V_{0,g}$ is a task group's prior belief, obtained by averaging trajectory-deduplicated returns and clipping to $(0,1)$.
$\operatorname{FirstRows}$ supplies one representative row per trajectory by retaining its first occurrence in the batch.
The accumulator $c$ sums turn-level teacher--old-policy log-ratios with decay $\gamma$; it updates the success belief $V_b$, whose increment is $v_b=V_b-V_{\mathrm{prev}}$.
The mean $\bar v_i$ and population SD over $B_i$ standardize these increments into $z_b$; a one-turn trajectory has zero SD.
The bounded multiplier $m_b$ uses band $\delta_{\mathrm{ag}}$, and $\lambda_{\mathrm{ag}}$ mixes it with the original advantage.
The prior deduplicates trajectories, while the base GRPO advantage still uses turn-row statistics.

\input{algorithm/agentopsd}

\subsection{Verified Teacher Continuations}
\label{app:alg-continuation}

\input{algorithm/ovcsd}

\noindent \textbf{Teacher continuations.}
OVCSD (Algorithm~\ref{alg:ovcsd}) makes at most $A_{\max}$ teacher-continuation attempts per eligible group; $\mathcal F$ collects these groups and $\mathcal V$ stores the verified comparison pairs.
$\operatorname{EligibleGroups}$ selects groups with no successful trajectory, raw episode-return population SD at most $\epsilon_R$, and at least one candidate prefix.
$\operatorname{Prefixes}$ builds a candidate set $C$ by grouping trajectories with the same action prefix and current observation, retaining states with a subsequent student action; $c$ denotes the selected candidate.
$U$ contains uncovered trajectories; $W$ records attempted candidate keys, distinguishing shared candidates from individual fallbacks.
$\operatorname{Members}$ returns the trajectory IDs represented by a group or candidate.
$\operatorname{Deepest}$ prefers an untried shared non-root prefix, then the deepest individual prefix.
$\operatorname{ReplayContinue}$ restores the candidate state and verifies its observation before sampling a skill-conditioned continuation $\tau^+$ within the remaining horizon; its indicator $v$ is one only if replay and task completion both succeed.
$\operatorname{AlignPairs}$ returns pairs $\mathcal P$ at the first action divergence while observations remain aligned.
Each pair $j$ matches one teacher turn to $m_j\geq1$ failed student turns, receiving comparative advantages $a_j^+$ and $a_j^-$, respectively.
After success, $\operatorname{Next}$ selects the deepest uncovered candidate; after failure it prefers the nearest untried ancestor, falling back to the deepest candidate.
Candidate members are removed from $U$ on success even if no aligned pair survives.

\noindent \textbf{OVCSD optimization masks.}
$\operatorname{VerifiedBatch}$ augments the original batch with verified teacher responses, producing $\widehat D$, masks $M^{\mathrm{PG}},M^{\mathrm{reg}}$, and advantages $\widehat A$.
Non-intervened groups keep their ordinary GRPO update.
For intervened failed groups, the original policy-gradient mask is cleared except at accepted divergent student turns; the original reference-KL and entropy masks remain active.
Appended skill-free teacher divergence rows receive the positive comparative advantage and both policy-gradient and regularization masks.
Later verified teacher rows receive only the suffix distillation loss, not policy-gradient, reference-KL, or entropy terms.
Old-policy and reference probabilities on appended rows are recomputed using their skill-free prompts.

\noindent \textbf{Suffix distillation.}
The suffix consists of teacher turns strictly after the divergence; $S_j$ is pair $j$'s set of valid suffix tokens, and $\mathcal J$ indexes pairs with nonempty suffixes.
$\widehat D^+$ contains their privileged teacher prompts, whereas $\widehat D$ uses skill-free student prompts for the same responses.
$\operatorname{TopTail}_{K}$ retains the teacher's $K$ most probable token IDs $I$ at each position and combines the remaining vocabulary into one tail bin, yielding a fixed $(K+1)$-category distribution $Q$.
$P_{\theta,bt}^{I}$ is the current student's distribution on that same partition under the skill-free prompt, recomputed at every actor forward pass.
The suffix objective sums per-pair token-averaged $\mathrm{KL}(Q\|P_\theta^{I})$, weighted by $\lambda_{\mathrm{suf}}$; its denominators $|S_j|$ are fixed over the complete pairs when accumulating microbatch contributions.
Empty suffixes contribute zero; their divergence updates remain active.

\subsection{Teacher-Based Reward Shaping}
\label{app:alg-reward-shaping}

\input{algorithm/adrs}

\noindent \textbf{Teacher signals and gating.}
ADRS (Algorithm~\ref{alg:adrs}) turns teacher scores into token rewards $h$ and an additional advantage correction $w$.
The row mean $\bar q_b$ measures teacher alignment with the sampled response; $\alpha_b\in(0,1)$ expresses this alignment relative to the batch, using temperature $\tau_{\mathrm{align}}$.
The group-level teacher value advantage (TVA), $T_g$, contrasts returns weighted by $\alpha$ and $1-\alpha$ using
\[
 \operatorname{WM}_g(R;\omega)=
 \frac{\sum_{b\in B_g}\omega_bR_b}{\sum_{b\in B_g}\omega_b+\epsilon_{\mathrm{num}}}.
\]
Here $\omega$ denotes the chosen weights for the task-group mean.
The coefficient $\eta$ sets the teacher-signal strength, and $\beta_{\mathrm{TVA}}$ controls the sigmoid gate's slope.

\noindent \textbf{Normalization and advantage correction.}
The centered teacher scores are normalized into $u=N_{\mathrm{task}}((q-\bar q)M)$.
$N_{\mathrm{task}}$ uses valid tokens to compute the batch-wide sample SD on ALFWorld, ScienceWorld, and Search-QA, or per-row population SD on WebShop, then divides by that SD plus $\epsilon_{\mathrm{num}}$.
TVA is broadcast to rows as $\widetilde T_b=T_{g(b)}$ and rescaled by $\Gamma_{\mathrm{task}}$: on ALFWorld, ScienceWorld, and Search-QA, divide by its sample SD plus $\epsilon_{\mathrm{num}}$ when that SD exceeds $\epsilon_{\mathrm{num}}$; otherwise, and on WebShop, leave it unchanged.

ADRS first adds the gated reward $h$ to $r$ and recomputes GRPO advantages, then adds the within-group correction $w$, clipped to $[-\delta_{\mathrm{ad}},\delta_{\mathrm{ad}}]$.
$\operatorname{Moments}$ returns the masked mean $\bar h_g$ and unbiased variance $v_g$ of the teacher rewards; $s_g(r+h)$ is the sample SD of the shaped row returns and determines whether the correction is applied.
All reward, TVA, and advantage calculations are detached.
The benchmark-specific normalization choices are listed in Table~\ref{tab:training-tasks}.

\flushbottom

%% file: algorithm/shared.tex
\begin{algorithm}[tbp]
\caption{Shared training loop}
\label{alg:shared}
\small
\begin{algorithmic}[1]
\Require $\theta_0,\mathcal M$; $B,G,S,H,\tau_{\mathrm{gen}}$
\Ensure Trained student parameters $\theta$
\State $\theta,\phi\gets\theta_0,\theta_0$
\For{$s=1,\ldots,S$}
  \State $\theta_{\mathrm{old}}\gets\theta$; $\mathcal F,\mathcal V\gets\varnothing,\varnothing$
  \State $X\gets\operatorname{SampleTasks}(B)$
  \State $\mathcal T\gets\operatorname{Rollout}(\theta_{\mathrm{old}},X;G,H,\tau_{\mathrm{gen}})$
  \If{$\mathcal M=\mathrm{OVCSD}$}
    \State $(\mathcal F,\mathcal V)\gets\operatorname{VerifyCont}(\mathcal T)$
  \EndIf
  \State $D\gets\operatorname{TurnRows}(\mathcal T)$; $r\gets\operatorname{Reward}(D)$
  \State $o\gets\operatorname{Score}(\theta_{\mathrm{old}},D)$
  \State $f\gets\operatorname{Score}(\phi,D)$; $A\gets\mathcal G(r)$
  \State $\operatorname{MethodStep}_{\mathcal M}(D,A,s,\mathcal F,\mathcal V)$
\EndFor
\State \Return $\theta$
\end{algorithmic}
\end{algorithm}

%% file: algorithm/grpo.tex
\begin{algorithm}[tbp]
\caption{GRPO}
\label{alg:grpo}
\small
\begin{algorithmic}[1]
\Require $D,A$
\State $\operatorname{Update}_D(A,0)$
\end{algorithmic}
\end{algorithm}

%% file: algorithm/grpo_opsd.tex
\begin{algorithm}[tbp]
\caption{GRPO+OPSD: constant gate}
\label{alg:grpo-opsd}
\small
\begin{algorithmic}[1]
\Require $D,A$; $\lambda_{\mathrm{sd}}$
\State $q\gets\operatorname{SkillReplay}(D,\theta_{\mathrm{old}})$
\State $L_{\mathrm{aux}}(p)\gets\frac{\lambda_{\mathrm{sd}}}{2}\langle q-p\rangle_M$
\State $\operatorname{Update}_D(A,L_{\mathrm{aux}})$
\end{algorithmic}
\end{algorithm}

%% file: algorithm/sdar.tex
\begin{algorithm}[tbp]
\caption{SDAR: detached confidence gate}
\label{alg:sdar}
\small
\begin{algorithmic}[1]
\Require $D,A$; $\lambda_{\mathrm{sd}},\beta$
\State $q\gets\operatorname{SkillReplay}(D,\theta_{\mathrm{old}})$
\State $g(p)\gets\operatorname{sg}\{\sigma[\beta(q-p)]\}$
\State $L_{\mathrm{aux}}(p)\gets\lambda_{\mathrm{sd}}\langle g(p)(q-p)\rangle_M$
\State $\operatorname{Update}_D(A,L_{\mathrm{aux}})$
\end{algorithmic}
\end{algorithm}

%% file: algorithm/skillsd.tex
\begin{algorithm}[tbp]
\caption{Skill-SD: importance-weighted K3}
\label{alg:skillsd}
\small
\begin{algorithmic}[1]
\Require $D,A$; $\lambda_{\mathrm{dist}},c_{\mathrm{K3}},c_{\mathrm{IS}}$
\State $q\gets\operatorname{SkillReplay}(D,\theta_{\mathrm{old}})$
\State $d(p)\gets p-q$
\State $k(p)\gets\exp(\min(-d(p),c_{\mathrm{K3}}))-1+d(p)$
\State $w(p)\gets\exp(\min(p-o,c_{\mathrm{IS}}))$
\State $L_{\mathrm{aux}}(p)\gets\lambda_{\mathrm{dist}}\langle w(p)k(p)\rangle_M$
\State $\operatorname{Update}_D(A,L_{\mathrm{aux}})$
\end{algorithmic}
\end{algorithm}

%% file: algorithm/icsd.tex
\begin{algorithm}[tbp]
\caption{ICSD: calibrated token weights}
\label{alg:icsd}
\small
\begin{algorithmic}[1]
\Require $D,A$; $\lambda_{\mathrm{ic}},\xi_{\mathrm{trust}},\xi_{\mathrm{inf}},\epsilon_{\mathrm{cal}},n_{\mathrm{cal}}$
\State $q\gets\operatorname{SkillReplay}(D,\theta_{\mathrm{old}})$
\For{each local actor minibatch $D'$ in the PPO epochs}
  \State Use the $q,o,A,M$ slices for $D'$
  \State $p^\star\gets\operatorname{Score}(\theta,D')$
  \State $\Delta\gets q-p^\star$; $u\gets A\exp(p^\star-o)\Delta$
  \State $g\gets\operatorname{CalCDF}_{\xi_{\mathrm{trust}}}(\Delta)$
  \State $h\gets\operatorname{CalCDF}_{\xi_{\mathrm{inf}}}(u)$
  \State $c\gets gM$
  \For{each response row $b$}
    \State $C_b\gets\{t:M_{bt}=1,\ \neg(A_{bt}<0\land\Delta_{bt}<0)\}$
    \State $z_b\gets\sum_{t\in C_b}g_{bt}h_{bt}$
    \If{$z_b>\epsilon_{\mathrm{cal}}$}
      \State $c_{bt}\gets g_{bt}h_{bt}\sum_{j\in C_b}g_{bj}/z_b,\quad t\in C_b$
    \EndIf
  \EndFor
  \State $L_{\mathrm{aux}}(p)\gets\lambda_{\mathrm{ic}}\langle\operatorname{sg}(c)(q-p)\rangle_M$
  \State $\operatorname{UpdateMiniBatch}_{D'}(A,L_{\mathrm{aux}})$
\EndFor
\end{algorithmic}
\end{algorithm}

%% file: algorithm/pcsd.tex
\begin{algorithm}[tbp]
\caption{PCSD: persistent teacher-support weighting}
\label{alg:pcsd}
\small
\begin{algorithmic}[1]
\Require $D,A$; $\lambda_{\mathrm{pc}},N_{\min},N_{\max},\rho_{\mathrm{pc}},v_l,v_h,\gamma_{\mathrm{pc}},\beta_{\mathrm{pc}}$
\State $q\gets\operatorname{SkillReplay}(D,\phi)$
\Function{$L_{\mathrm{aux}}$}{$p$}
  \State $\Delta\gets\operatorname{sg}(q-p)M$
  \State $m_s\gets\operatorname{DecayMean}(\Delta,M;N_{\min},\rho_{\mathrm{pc}})$
  \State $m_l\gets\operatorname{DecayMean}(\Delta,M;N_{\max},\rho_{\mathrm{pc}})$
  \State $v\gets\operatorname{WindowVar}(\Delta,M;N_{\max})$
  \State $a\gets\operatorname{clip}((v-v_l)/(v_h-v_l),0,1)$
  \State $\bar\Delta\gets(1-a)m_s+am_l$
  \State $\ell\gets\operatorname{WindowSlope}(\Delta,M;N_{\max})$
  \State $d_b\gets\sum_t M_{bt}|\Delta_{bt}|/\max(1,\sum_tM_{bt})+\epsilon_{\mathrm{pc}}$
  \State $h\gets\operatorname{clip}(1-\gamma_{\mathrm{pc}}\operatorname{ReLU}(-\ell/d),0,1)$
  \State $w\gets\operatorname{sg}\{\sigma(\beta_{\mathrm{pc}}\bar\Delta)hM\}$
  \State \Return $\lambda_{\mathrm{pc}}\langle w(q-p)\rangle_M$
\EndFunction
\State $\operatorname{Update}_D(A,L_{\mathrm{aux}})$
\end{algorithmic}
\end{algorithm}

%% file: algorithm/rlsd.tex
\begin{algorithm}[tbp]
\caption{RLSD: sign-aware advantage reweighting}
\label{alg:rlsd}
\small
\begin{algorithmic}[1]
\Require $D,A,s$; $\lambda_0,S,\delta_{\mathrm{r}}$
\State $q\gets\operatorname{SkillReplay}(D,\theta_{\mathrm{old}})$
\State $\lambda_s\gets\lambda_0\max(1-s/S,0)$
\State $w\gets\operatorname{clip}(e^{\operatorname{sign}(A)(q-o)},1-\delta_{\mathrm{r}},1+\delta_{\mathrm{r}})$
\State $\widehat A\gets\operatorname{sg}\{A[(1-\lambda_s)+\lambda_s w]\}$
\State $\operatorname{Update}_D(\widehat A,0)$
\end{algorithmic}
\end{algorithm}

%% file: algorithm/ocsd.tex
\begin{algorithm}[tbp]
\caption{OCSD: full/ablated future-evidence replay}
\label{alg:ocsd}
\small
\begin{algorithmic}[1]
\Require $D,A,o,F_b$; $\alpha_{\mathrm{sel}},L_f,\lambda_{\mathrm{oc}}$
\State $z\gets0$
\For{trajectory $i$}
  \State $E_i\gets\{b\in B_i:F_b>0,\ |A_b|>\epsilon_A\}$
  \If{$E_i=\varnothing$} \State \textbf{continue} \EndIf
  \State $n_b\gets-\dfrac{\sum_t M_{bt}o_{bt}}{\max(1,\sum_t M_{bt})}$ for $b\in E_i$
  \State $k_i\gets\min(|E_i|,\max(1,\lceil\alpha_{\mathrm{sel}}|E_i|\rceil))$
  \State $J_i\gets\operatorname{Top}_{k_i}(E_i;n)$
  \For{$b\in J_i$}
    \State $(D_b^{+},D_b^{-})\gets\operatorname{EvidencePair}(D_b,L_f)$
    \State $q_b^{\mathrm{full}}\gets\operatorname{Score}(\theta_{\mathrm{old}},D_b^{+})$
    \State $q_b^{\mathrm{abl}}\gets\operatorname{Score}(\theta_{\mathrm{old}},D_b^{-})$
    \State $z_{bt}\gets\tanh[(q^{\mathrm{full}}_{bt}-q^{\mathrm{abl}}_{bt})/2]$
  \EndFor
\EndFor
\State $\widehat A\gets\operatorname{sg}\{A[1+\lambda_{\mathrm{oc}}\operatorname{sign}(A)z]\}$
\State $\operatorname{Update}_D(\widehat A,0)$
\end{algorithmic}
\end{algorithm}

%% file: algorithm/agentopsd.tex
\begin{algorithm}[tbp]
\caption{AgentOPSD: signed turn credit}
\label{alg:agentopsd}
\small
\begin{algorithmic}[1]
\Require $D,A,R$; $\gamma,\lambda_{\mathrm{ag}},\delta_{\mathrm{ag}}$
\State $q\gets\operatorname{SkillReplay}(D,\theta_{\mathrm{old}})$
\For{task group $g$}
  \State $I_g\gets\operatorname{FirstRows}(B_g,i)$
  \State $V_{0,g}\gets C_{\epsilon_{\mathrm{prior}}}(\operatorname{mean}_{b\in I_g}R_b)$
\EndFor
\For{trajectory $i$}
  \State $c\gets0$; $V_{\mathrm{prev}}\gets V_{0,g(i)}$
  \For{$b\in B_i$ in turn order}
    \State $c\gets\gamma c+\sum_t M_{bt}(q_{bt}-o_{bt})$
    \State $V_b\gets C_{\epsilon_{\mathrm{bel}}}(\sigma[\operatorname{logit}(V_{0,g(i)})+c])$
    \State $v_b\gets V_b-V_{\mathrm{prev}}$; $V_{\mathrm{prev}}\gets V_b$
  \EndFor
  \For{$b\in B_i$}
    \State $z_b\gets(v_b-\bar v_i)/(\operatorname{SD}_{\mathrm{pop}}(v_{B_i})+\epsilon_{\mathrm{norm}})$
    \State $m_b\gets1+\delta_{\mathrm{ag}}\operatorname{clip}(\operatorname{sign}(A_b)z_b,-1,1)$
    \State $\widehat A_{bt}\gets A_b[(1-\lambda_{\mathrm{ag}})+\lambda_{\mathrm{ag}}m_b]M_{bt}$
  \EndFor
\EndFor
\State $\operatorname{Update}_D(\operatorname{sg}(\widehat A),0)$
\end{algorithmic}
\end{algorithm}

%% file: algorithm/ovcsd.tex
\begin{algorithm}[tb]
\caption{OVCSD: verified continuations}
\label{alg:ovcsd}
\small
\begin{algorithmic}[1]
\Require $A_{\max},\epsilon_R,K,\lambda_{\mathrm{suf}}$
\Function{VerifyCont}{$\mathcal T$}
  \State $\mathcal F\gets\operatorname{EligibleGroups}(\mathcal T;\epsilon_R)$
  \State $\mathcal V\gets\varnothing$
  \For{$g\in\mathcal F$}
    \State $U\gets\operatorname{Members}(\mathcal T_g)$
    \State $C\gets\operatorname{Prefixes}(\mathcal T_g)$; $W\gets\varnothing$
    \State $c\gets\operatorname{Deepest}(C,U,W)$
    \For{$a=1,\ldots,A_{\max}$}
      \If{$c=\varnothing$ or $U=\varnothing$}
        \State \textbf{break}
      \EndIf
      \State $(\tau^+,v)\gets\operatorname{ReplayContinue}(\theta_{\mathrm{old}},c,H)$
      \If{$v=1$}
        \State $\mathcal P\gets\operatorname{AlignPairs}(\mathcal T_g,\tau^+,c)$
        \State $(a_j^+,a_j^-)\gets(\sqrt{m_j},-1/\sqrt{m_j})$, $j\in\mathcal P$
        \State $\mathcal V\gets\mathcal V\cup\mathcal P$; $U\gets U\setminus\operatorname{Members}(c)$
      \EndIf
      \State $W\gets W\cup\{\operatorname{key}(c)\}$
      \State $c\gets\operatorname{Next}(C,U,W,c,v)$
    \EndFor
  \EndFor
  \State \Return $(\mathcal F,\mathcal V)$
\EndFunction
\Procedure{OVCSDUpdate}{$D,A,\mathcal F,\mathcal V$}
  \State $(\widehat D,\widehat A)\gets\operatorname{VerifiedBatch}(D,A,\mathcal F,\mathcal V)$
  \State $(I,Q)\gets\operatorname{TopTail}_{K}(\theta_{\mathrm{old}},\widehat D^{+})$
  \State $\mathcal J\gets\{j\in\mathcal V:|S_j|>0\}$
  \State $L_{\mathrm{suf}}(\theta)\gets\lambda_{\mathrm{suf}}\displaystyle\sum_{j\in\mathcal J}\frac{1}{|S_j|}\!\sum_{(b,t)\in S_j}\!\mathrm{KL}(Q_{bt}\|P_{\theta,bt}^{I})$
  \State $\operatorname{Update}_{\widehat D}(\widehat A,L_{\mathrm{suf}})$
\EndProcedure
\end{algorithmic}
\end{algorithm}

%% file: algorithm/adrs.tex
\begin{algorithm}[tbp]
\caption{ADRS: teacher-gated reward and advantage shaping}
\label{alg:adrs}
\small
\begin{algorithmic}[1]
\Require $D,r$; $\eta,\tau_{\mathrm{align}},\beta_{\mathrm{TVA}},\delta_{\mathrm{ad}}$
\State $q\gets\operatorname{SkillReplay}(D,\theta_{\mathrm{old}})$
\State $\bar q_b\gets\operatorname{mean}_{t:M_{bt}=1}q_{bt}$
\State $u\gets N_{\mathrm{task}}((q-\bar q)M)$; $T_g\gets0$ for all $g$
\State $\alpha_b\gets\sigma[(\bar q_b-\operatorname{mean}_b\bar q_b)/(\tau_{\mathrm{align}}+\epsilon_{\mathrm{num}})]$
\For{$g:|B_g|>1$}
  \State $T_g\gets\operatorname{WM}_g(R;\alpha)-\operatorname{WM}_g(R;1-\alpha)$
\EndFor
\State $\widetilde T_b\gets T_{g(b)}$; $\widetilde T\gets\Gamma_{\mathrm{task}}(\widetilde T)$
\State $h_{bt}\gets\eta\,\sigma(\beta_{\mathrm{TVA}}\widetilde T_b)u_{bt}$
\State $\widehat A\gets\mathcal G(r+h)$
\For{$g:|B_g|>1,\ s_g(r+h)\geq\epsilon_R$}
  \State $(\bar h_g,v_g)\gets\operatorname{Moments}(h_{B_g};M_{B_g})$
  \State $w_{bt}\gets\operatorname{clip}\!\left(\eta\,\dfrac{h_{bt}-\bar h_g}{\sqrt{v_g+\epsilon_{\mathrm{num}}}},-\delta_{\mathrm{ad}},\delta_{\mathrm{ad}}\right),\ b\in B_g$
  \State $\widehat A_{bt}\gets\widehat A_{bt}+w_{bt}M_{bt}$
\EndFor
\State $\operatorname{Update}_D(\operatorname{sg}(\widehat A),0)$
\end{algorithmic}
\end{algorithm}

%% file: sec/16_cross_setting_results.tex
\section{Cross-Setting Gains and Regressions}
\label{app:endpoint-moments}

\input{table/generated_twelve_setting_summary}
\input{figure/evidence_rank_flow}
\input{figure/stability_cases}
\input{table/generated_full_training_summary}
\input{table/generated_full_horizon_outcomes}

\subsection{All Twelve Settings}
\label{app:all-setting-results}

Table~\ref{tab:twelve-setting-summary} combines the sampled final evaluations on all four benchmarks, assigning equal weight to each model--benchmark setting.
It uses success for ALFWorld/WebShop, exact-match accuracy for Search-QA, and task score for ScienceWorld.
Only PCSD and AgentOPSD improve the twelve-setting mean over GRPO, by 4.58 and 1.66 points, respectively.
ICSD wins seven settings, including all three Search-QA scales, but its 66.02-point deficit on 4B WebShop and 40.73-point deficit on 9B ScienceWorld leave its overall mean 4.93 points below GRPO.
OCSD likewise illustrates scale-dependent gains: it improves on GRPO in 2B and 4B ScienceWorld but trails it by 14.69 points at 9B, and its twelve-setting mean remains 0.18 points below GRPO.
Win counts and gaps are descriptive comparisons of means; rollout-seed SDs do not support conclusions about variability across independent training runs.

\subsection{Low-Performance Outcomes}
\label{app:low-performance}

Across the twelve settings, \PaperAllLowFinalRuns{} trained models have a repeated final-evaluation mean below 5 on their benchmark's reported metric.
One ALFWorld model and five WebShop models also underperform their pretrained backbones; nine 2B ScienceWorld models have task scores below 2 despite improving over their initially negative scores.
No Search-QA model falls below this threshold in repeated final evaluation.
The same low endpoint can therefore reflect either lost pretrained capability or limited task progress.
Reporting these outcomes alongside aggregate gains distinguishes severe regressions from settings where the training methods have largely failed to learn useful behavior.
Tables~\ref{tab:main_results} and~\ref{tab:main_results_search_webshop} provide the model-specific comparisons; Appendix~\ref{app:failure-evidence} relates low validation scores to training trajectories and numerical diagnostics.

%% file: table/generated_twelve_setting_summary.tex
\begin{table}[t]
\centering\small
\begin{tabular}{lrrcr}
\toprule
Method & Mean & $\Delta$ & W/L/T & Worst \\
\midrule
PCSD & 54.26 & +4.58 & 6/5/1 & -8.31 \\
AgentOPSD & 51.34 & +1.66 & 6/5/1 & -16.51 \\
GRPO & 49.68 & +0.00 & 0/0/12 & +0.00 \\
OCSD & 49.50 & -0.18 & 6/6/0 & -29.73 \\
RLSD & 46.54 & -3.14 & 6/6/0 & -70.02 \\
SDAR & 46.33 & -3.35 & 3/8/1 & -18.30 \\
OVCSD & 46.17 & -3.51 & 5/7/0 & -69.73 \\
ICSD & 44.75 & -4.93 & 7/4/1 & -66.02 \\
ADRS & 44.61 & -5.07 & 3/8/1 & -23.36 \\
GRPO+OPSD & 43.05 & -6.63 & 6/5/1 & -80.53 \\
Skill-SD & 39.45 & -10.23 & 5/6/1 & -70.02 \\
\bottomrule
\end{tabular}
\caption{Equal-weight means across all twelve model--benchmark settings, using ALFWorld/WebShop success (\%), Search-QA exact-match accuracy (\%), and ScienceWorld task score. $\Delta$ and Worst are the mean and worst paired differences from GRPO; W/L/T counts wins/losses/ties of reported means.}
\label{tab:twelve-setting-summary}
\end{table}

%% file: figure/evidence_rank_flow.tex
\begin{figure*}[!t]
\centering
\includegraphics[width=\textwidth]{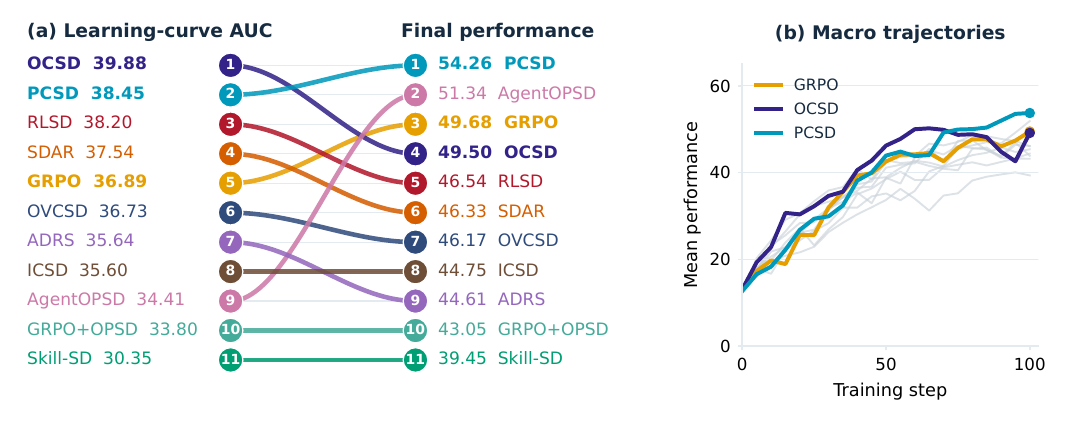}
\caption{(a) Rankings across all twelve settings by mean training-curve AUC and eight-seed final performance from Tables~\ref{tab:main_results} and~\ref{tab:main_results_search_webshop}. AUC integrates each benchmark's validation metric and normalizes by the training horizon. (b) Equally weighted mean validation trajectories, highlighting GRPO and the AUC and final-performance leaders, OCSD and PCSD; the other eight methods are shown in gray.}
\label{fig:evidence-rank-flow}
\end{figure*}

%% file: figure/stability_cases.tex
\begin{figure*}[!t]
\centering
\includegraphics[width=\textwidth]{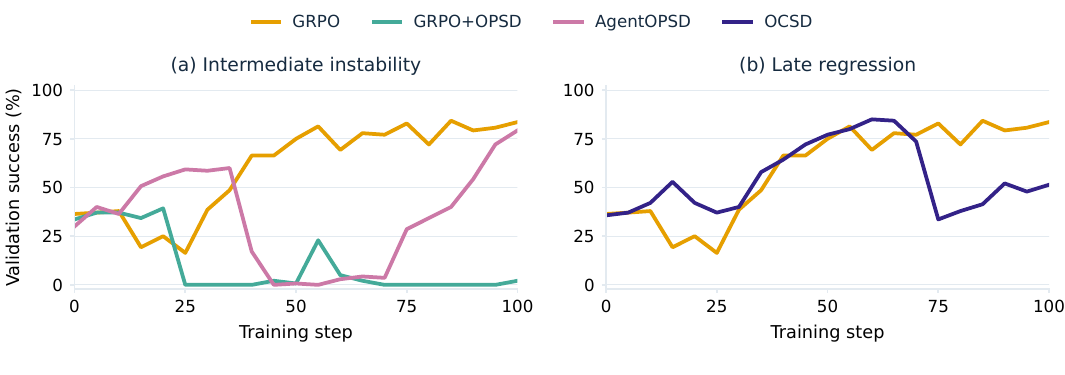}
\caption{Training instability on 4B ALFWorld. (a) AgentOPSD recovers from a prolonged near-zero phase, whereas GRPO+OPSD finishes near zero after losing its early gains. (b) OCSD declines from a later peak. Both panels retain GRPO as a reference and show raw validation success every five steps. These curves are selected from Figure~\ref{fig:validation-curves}.}
\label{fig:stability-cases}
\end{figure*}

%% file: table/generated_full_training_summary.tex
\begin{table}[!t]
\centering\small
\setlength{\tabcolsep}{3pt}
\begin{tabular}{@{}lrrrrrr@{}}
\toprule
Method & AUC & \shortstack{AUC\\rank} & Final & \shortstack{Final\\rank} & \shortstack{Low\\/12} & \shortstack{Drop\\/12} \\
\midrule
GRPO & 36.89 & 5 & 49.68 & 3 & 1 & 0 \\
GRPO+OPSD & 33.80 & 10 & 43.05 & 10 & 2 & 2 \\
SDAR & 37.54 & 4 & 46.33 & 6 & 1 & 0 \\
Skill-SD & 30.35 & 11 & 39.45 & 11 & 3 & 0 \\
RLSD & 38.20 & 3 & 46.54 & 5 & 2 & 2 \\
OCSD & 39.88 & 1 & 49.50 & 4 & 0 & 1 \\
OVCSD & 36.73 & 6 & 46.17 & 7 & 2 & 2 \\
AgentOPSD & 34.41 & 9 & 51.34 & 2 & 1 & 0 \\
ADRS & 35.64 & 7 & 44.61 & 9 & 1 & 1 \\
ICSD & 35.60 & 8 & 44.75 & 8 & 2 & 1 \\
PCSD & 38.45 & 2 & 54.26 & 1 & 1 & 0 \\
\bottomrule
\end{tabular}
\caption{Training progress and stability across all twelve equally weighted model--benchmark settings. AUC is the area under the raw validation curve divided by 100 training steps; Final is the repeated final-evaluation mean. Low counts final validation scores below 5; Drop counts peak-to-final declines of at least 20 points. These thresholds describe observed performance, not failure probabilities.}
\label{tab:full-training-summary}
\end{table}

%% file: table/generated_full_horizon_outcomes.tex
\begin{table}[tbp]
\centering\small
\setlength{\tabcolsep}{3pt}
\begin{tabular}{@{}lrrrrr@{}}
\toprule
Method & \shortstack{Step\\20} & \shortstack{Step\\40} & \shortstack{Step\\60} & \shortstack{Step\\80} & \shortstack{Step\\100} \\
\midrule
GRPO & 25.65 & 39.34 & 44.22 & 47.68 & 49.47 \\
GRPO+OPSD & 27.17 & 35.02 & 38.31 & 42.38 & 43.19 \\
SDAR & \textbf{30.63} & \textbf{40.81} & 42.50 & 45.52 & 46.15 \\
Skill-SD & 21.64 & 30.40 & 33.99 & 38.19 & 39.36 \\
RLSD & 30.37 & 37.29 & 43.80 & 48.65 & 46.10 \\
OCSD & 30.39 & 40.61 & \textbf{50.00} & 48.85 & 49.15 \\
OVCSD & 28.38 & 36.53 & 42.13 & 47.44 & 43.89 \\
AgentOPSD & 24.41 & 31.95 & 35.80 & 46.27 & 51.88 \\
ADRS & 26.16 & 35.84 & 42.75 & 44.03 & 45.38 \\
ICSD & 26.09 & 35.42 & 43.50 & 45.66 & 44.39 \\
PCSD & 26.82 & 38.16 & 43.82 & \textbf{50.03} & \textbf{53.73} \\
\bottomrule
\end{tabular}
\caption{Mean validation performance at successive training stages, with equal weight on all twelve settings. ALFWorld/WebShop use success (\%), Search-QA uses exact-match accuracy (\%), and ScienceWorld uses task score. Bold marks the highest mean at each stage.}
\label{tab:full-horizon-outcomes}
\end{table}

%% file: sec/17_training_dynamics.tex
\section{Training Dynamics and Stability}
\label{app:logged-diagnostics}
\label{app:training-curves}

\input{table/generated_full_training_runs_alfworld}
\input{table/generated_full_training_runs_scienceworld}
\input{table/generated_full_training_runs_search}
\input{table/generated_full_training_runs_webshop}
\input{table/generated_full_failure_cases}
\input{figure/phase_cost_breakdown}
\input{table/generated_cost_performance_settings}
\input{table/generated_full_performance_cost}
\input{table/generated_full_budget_aligned}
\input{table/generated_full_compute_ledger}

We apply the same trajectory summaries and diagnostic criteria to all \PaperCompletedRuns{} runs across four benchmarks.
ALFWorld and WebShop use success (\%), Search-QA uses exact-match accuracy (\%), and ScienceWorld uses task score, retaining negative values.
Each model--benchmark setting receives equal weight in aggregate AUC, stage-wise performance, and final-score comparisons.
These aggregate values summarize the reported metrics of the four benchmarks rather than a pooled task-success probability.
The analyses below connect these summaries to intermediate instability, late regression, and prolonged low-score plateaus.

\subsection{Aggregate Progress and Method Rankings}
\label{app:learning-curves}

Method rankings depend on whether performance is assessed throughout training or only at its endpoint.
Figure~\ref{fig:evidence-rank-flow} contrasts training-curve AUC with repeated final evaluation and shows the twelve-setting mean validation trajectories.
OCSD leads by AUC, whereas PCSD leads at the endpoint; the latter's advantage emerges later in training.
The stage-wise comparison makes this change explicit: SDAR leads at steps 20 and 40, OCSD at step 60, and PCSD at steps 80 and 100.
An endpoint advantage therefore need not reflect superior performance sustained throughout optimization.
Tables~\ref{tab:full-training-summary} and~\ref{tab:full-horizon-outcomes} report the aggregate rankings, low-performance counts, and stage-wise means for all eleven methods.

\subsection{Intermediate Instability and Late Regression}
\label{app:instability-cases}
\label{app:searchqa-decoding}

The 4B ALFWorld trajectories illustrate why instability cannot be reduced to a low final score.
AgentOPSD recovers after six consecutive validations at or below 4.3\%, whereas GRPO+OPSD loses its early gains and finishes near zero.
OCSD reaches 85.0\% before declining to 51.4\%, with its final three validations averaging 50.5\%.
A criterion based only on near-zero final success would identify GRPO+OPSD but miss both AgentOPSD's temporary collapse and OCSD's sustained late regression.
Assessing stability therefore requires examining the depth and persistence of intermediate declines, even when the final model remains capable.
Figure~\ref{fig:stability-cases} isolates these histories from Figure~\ref{fig:validation-curves}, retaining GRPO as a common reference.

\noindent \textbf{Decoding-specific collapse.}
Search-QA reveals that training instability can also depend on the decoding strategy: a model may retain effective sampled performance despite a collapse under greedy decoding.
Training-time validation uses greedy decoding ($T=0$), whereas final evaluation samples at $T=0.4$ over eight rollout seeds; both report exact-match accuracy on the same 3,125 questions.
The Search-QA panels in Figure~\ref{fig:validation-curves} show these training trajectories; Tables~\ref{tab:full-training-runs-search} and~\ref{tab:main_results_search_webshop} report their trajectory summaries and sampled final results, respectively.

For 2B OVCSD, greedy accuracy reaches 38.02\% at step 70 and remains at 37.09\% at step 80, before falling to 4.77\%, 0.22\%, 12.86\%, and 1.34\% at steps 85, 90, 95, and 100.
The final-three-validation mean is only 4.81\%, with a 36.67-point peak-to-final drop under the unchanged greedy protocol.
Yet the same final checkpoint scores $33.21\pm0.33$\% with sampling, compared with GRPO's $32.38\pm0.28$\%.
This is a decoding-specific collapse: previously acquired greedy performance is lost even though sampling at $T=0.4$ still produces effective task behavior.
A sampled endpoint alone would leave this failure mode unobserved.

ICSD shows a milder greedy regression rather than near-zero collapse.
Its 2B greedy accuracy peaks at 39.52\% at step 65, falls to 21.73\% at step 90, and partially recovers to 31.49\% at step 100; the final-three mean is 26.31\%.
Nevertheless, its sampled final accuracy of $39.39\pm0.19$\% leads the 2B Search-QA comparison.
Together, these cases extend the collapse and regression analysis beyond failures visible under sampled evaluation: effective nonzero-temperature sampling does not rule out a collapse or substantial loss of performance under greedy decoding.

\subsection{Early Stagnation on ScienceWorld}
\label{app:scienceworld-plateau}

\noindent \textbf{Low-score plateaus are informative training outcomes.}
At 2B, eight methods including GRPO improve over their initially negative scores but remain at 0.83 in every validation from steps 10 to 100, with repeated final scores of $0.83\pm0.00$.
Flat task scores need not imply inactive optimization: GRPO records finite, nonzero actor gradients throughout training.
For an initially weak policy, updates may change action probabilities without improving measured task outcomes if more effective behavior is rarely discovered.
This provides a plausible interpretation of the stagnation; matching aggregate scores do not require identical policies.
Such settings are valuable because they test whether teacher guidance can help agents move beyond limited initial task competence, rather than only improve agents that already learn effectively.
The overlapping traces in Figure~\ref{fig:validation-curves} reflect these matching recorded scores, with colored markers distinguishing the methods.

\noindent \textbf{Escaping the plateau remains method- and scale-dependent.}
OCSD and OVCSD reach final 2B task scores of $13.78\pm0.03$ and $13.85\pm0.37$, whereas RLSD reaches $1.89\pm0.07$.
Their separation from the common plateau shows that this setting is not uniformly unlearnable under the available training budget.
Yet the leading method changes with scale: OCSD reaches 41.77 at 4B, while GRPO leads at 9B with 58.99.
These results motivate examining whether a method enables sustained task progress at a given backbone scale, rather than treating either a flat curve as stability or a small-model gain as a general advantage.
Table~\ref{tab:full-training-runs-scienceworld} reports peak, late-training, and final task scores for all three scales.

\subsection{Per-Run Training Profiles}
\label{app:peak-endpoint}
\label{app:per-setting-cost}

Aggregate curves can conceal instability confined to a particular task or model size.
For every benchmark, we report each run's peak, final, and late-training performance alongside its AUC and computational cost.
Peak is the highest validation metric at steps $0,5,\ldots,100$; Step records its first occurrence, Tail-3 averages steps 90, 95, and 100, and Drop is Peak minus Final in the corresponding metric's points.
AUC integrates the raw validation curve by the trapezoidal rule and divides by the 100-step horizon; it is not ROC-AUC.
These summaries capture different aspects of the trajectory: a low Tail-3 indicates that weak performance extends beyond the final measurement, whereas a small Drop does not exclude earlier instability.
For example, AgentOPSD on 4B ALFWorld finishes at its own peak despite the prolonged near-zero phase described above.
Tables~\ref{tab:full-training-runs-alfworld}--\ref{tab:full-training-runs-webshop} report all \PaperCompletedRuns{} profiles with the same columns for each benchmark.

Net time includes rollout generation, teacher computation, and policy updates; total time additionally includes in-loop validation and checkpoint saving.
All runs use eight H800 GPUs, so the corresponding GPU-hours equal elapsed hours multiplied by eight.
These per-run measurements support the paired cost comparisons in Appendix~\ref{app:performance-time-budgets}.

\subsection{Task Failure and Numerical Diagnostics}
\label{app:failure-evidence}

We use a final validation metric below 5 as a descriptive low-performance criterion across all four benchmarks.
It identifies \PaperAllLowValidationRuns{} runs: one ALFWorld run, five WebShop runs, one Search-QA run, and nine ScienceWorld runs.
The ALFWorld and WebShop runs also regress below their pretrained final-evaluation baselines, whereas the ScienceWorld cases largely reflect low-score stagnation rather than loss from a capable starting point.
Search-QA OVCSD illustrates the decoding-dependent case discussed above: its low validation endpoint is not reproduced under sampled final evaluation.
Thus, the common threshold identifies cases for analysis without treating every low score as a training collapse.
Table~\ref{tab:full-failure-cases} summarizes these sixteen runs, including their training trajectories, repeated final results, late-training interaction lengths, and nonfinite-gradient counts.

\noindent \textbf{Persistent failure near the action limit.}
In all five low-performing WebShop runs, student rollout success is zero throughout the final 20 training steps, with mean episode lengths of 14.99--15.00 against the 15-action limit.
Over the same interval, GRPO+OPSD on 4B ALFWorld averages 7.1\% training success and 49.0 of the allowed 50 actions, compared with 81.5\% success and 18.4 actions for GRPO.
By contrast, low-performing ScienceWorld runs average 12.88--20.00 of the allowed 20 turns, and Search-QA OVCSD averages 2.48 of four turns.
Near-limit interaction length therefore characterizes the persistent WebShop failures but is not a universal signature of low performance.
Joint inspection of task progress and interaction length distinguishes wasted interaction budgets from other forms of ineffective learning.

\noindent \textbf{Gradient diagnostics alone do not identify task failure.}
Only \PaperAllLowAndNonfiniteRuns{} of the \PaperAllLowValidationRuns{} low-performing runs record nonfinite actor gradients; the other \PaperAllLowWithoutNonfiniteRuns{} do not.
Across all four benchmarks, \PaperAllNonfiniteWithoutLowRuns{} of the \PaperAllNonfiniteRuns{} runs with nonfinite gradients finish with validation performance at or above 5.
All \PaperAllFiniteDropRuns{} runs with peak-to-final declines of at least 20 points also have finite recorded gradients throughout.
Finite recorded gradients therefore do not establish successful learning, just as a nonfinite-gradient event does not by itself imply a near-zero endpoint.
Numerical checks and behavioral evaluation are complementary requirements for assessing training stability.

%% file: table/generated_full_training_runs_alfworld.tex
\begin{table*}[t]
\centering\small
\setlength{\tabcolsep}{4pt}
\begin{tabular}{@{}lrrrrrrrr@{}}
\toprule
Method & Peak & Step & Final & Tail-3 & Drop & AUC & Train h & Total h \\
\midrule
\multicolumn{9}{l}{\textbf{Qwen3.5-2B}} \\
GRPO & 31.40 & 75 & 27.10 & 24.03 & 4.30 & 19.25 & 13.49 & 14.37 \\
GRPO+OPSD & 44.30 & 95 & 21.40 & 35.47 & 22.90 & 21.90 & 15.68 & 16.62 \\
SDAR & 16.40 & 40 & 16.40 & 10.00 & 0.00 & 7.58 & 15.90 & 16.71 \\
Skill-SD & 75.00 & 100 & 75.00 & 71.67 & 0.00 & 32.84 & 14.18 & 15.08 \\
RLSD & 75.00 & 95 & 46.40 & 62.37 & 28.60 & 27.52 & 15.18 & 16.11 \\
OCSD & 56.40 & 75 & 42.90 & 39.77 & 13.50 & 32.45 & 13.28 & 14.20 \\
OVCSD & 60.00 & 75 & 57.10 & 46.63 & 2.90 & 28.14 & 22.23 & 23.26 \\
AgentOPSD & 49.30 & 95 & 48.60 & 46.67 & 0.70 & 21.15 & 15.73 & 16.70 \\
ADRS & 15.70 & 55 & 11.40 & 12.83 & 4.30 & 12.56 & 15.69 & 16.38 \\
ICSD & 67.90 & 75 & 55.00 & 53.33 & 12.90 & 35.15 & 15.26 & 16.13 \\
PCSD & 69.30 & 100 & 69.30 & 63.33 & 0.00 & 31.33 & 14.12 & 15.03 \\
\midrule
\multicolumn{9}{l}{\textbf{Qwen3.5-4B}} \\
GRPO & 84.30 & 85 & 83.60 & 81.20 & 0.70 & 59.78 & 14.94 & 16.22 \\
GRPO+OPSD & 39.30 & 20 & 2.10 & 0.70 & 37.20 & 9.92 & 24.14 & 25.55 \\
SDAR & 75.00 & 95 & 72.10 & 72.60 & 2.90 & 61.18 & 17.27 & 18.71 \\
Skill-SD & 70.70 & 55 & 55.70 & 55.00 & 15.00 & 50.73 & 18.61 & 20.12 \\
RLSD & 90.00 & 55 & 82.10 & 83.53 & 7.90 & 76.14 & 15.54 & 17.15 \\
OCSD & 85.00 & 60 & 51.40 & 50.47 & 33.60 & 55.10 & 17.75 & 19.34 \\
OVCSD & 82.10 & 95 & 76.40 & 78.80 & 5.70 & 61.57 & 20.81 & 22.40 \\
AgentOPSD & 79.30 & 100 & 79.30 & 68.57 & 0.00 & 33.66 & 22.62 & 24.43 \\
ADRS & 78.60 & 45 & 73.60 & 73.10 & 5.00 & 66.25 & 16.20 & 17.68 \\
ICSD & 83.60 & 65 & 81.40 & 75.70 & 2.20 & 65.21 & 19.57 & 21.08 \\
PCSD & 81.40 & 95 & 75.70 & 78.07 & 5.70 & 55.76 & 18.92 & 20.62 \\
\midrule
\multicolumn{9}{l}{\textbf{Qwen3.5-9B}} \\
GRPO & 70.00 & 80 & 63.60 & 67.17 & 6.40 & 55.55 & 22.16 & 24.51 \\
GRPO+OPSD & 72.90 & 100 & 72.90 & 69.07 & 0.00 & 53.60 & 24.16 & 26.47 \\
SDAR & 85.00 & 85 & 75.00 & 77.87 & 10.00 & 64.78 & 23.21 & 25.48 \\
Skill-SD & 77.10 & 95 & 72.10 & 74.27 & 5.00 & 61.77 & 24.65 & 27.08 \\
RLSD & 92.10 & 85 & 89.30 & 88.80 & 2.80 & 71.56 & 22.00 & 24.48 \\
OCSD & 87.10 & 95 & 81.40 & 82.60 & 5.70 & 61.77 & 21.19 & 23.37 \\
OVCSD & 84.30 & 100 & 84.30 & 79.03 & 0.00 & 64.03 & 23.22 & 25.63 \\
AgentOPSD & 77.10 & 80 & 76.40 & 74.77 & 0.70 & 56.79 & 23.94 & 26.17 \\
ADRS & 82.90 & 90 & 82.10 & 80.23 & 0.80 & 70.08 & 22.39 & 24.83 \\
ICSD & 93.60 & 90 & 92.10 & 92.60 & 1.50 & 72.62 & 24.13 & 26.45 \\
PCSD & 77.90 & 100 & 77.90 & 72.87 & 0.00 & 55.50 & 25.25 & 27.62 \\
\bottomrule
\end{tabular}
\caption{ALFWorld training profiles for all 33 runs, using success (\%). Peak includes initialization; Step gives its first occurrence; Final is step-100 validation; Tail-3 averages steps 90, 95, and 100; Drop is Peak minus Final. AUC is normalized by the training horizon. Train time excludes validation and saving; Total includes both.}
\label{tab:full-training-runs-alfworld}
\end{table*}

%% file: table/generated_full_training_runs_scienceworld.tex
\begin{table*}[t]
\centering\small
\setlength{\tabcolsep}{4pt}
\begin{tabular}{@{}lrrrrrrrr@{}}
\toprule
Method & Peak & Step & Final & Tail-3 & Drop & AUC & Train h & Total h \\
\midrule
\multicolumn{9}{l}{\textbf{Qwen3.5-2B}} \\
GRPO & 0.83 & 5 & 0.83 & 0.83 & 0.00 & 0.74 & 4.95 & 6.34 \\
GRPO+OPSD & 0.83 & 5 & 0.83 & 0.83 & 0.00 & 0.79 & 5.62 & 7.02 \\
SDAR & 1.08 & 5 & 0.83 & 0.83 & 0.25 & 0.75 & 5.74 & 7.24 \\
Skill-SD & 0.83 & 5 & 0.83 & 0.83 & 0.00 & 0.76 & 5.53 & 6.95 \\
RLSD & 2.39 & 95 & 1.89 & 2.02 & 0.51 & 1.35 & 6.87 & 8.09 \\
OCSD & 13.86 & 100 & 13.86 & 13.60 & 0.00 & 11.72 & 6.04 & 7.35 \\
OVCSD & 15.77 & 90 & 13.71 & 14.92 & 2.05 & 8.90 & 7.98 & 9.36 \\
AgentOPSD & 0.83 & 5 & 0.83 & 0.83 & 0.00 & 0.77 & 5.48 & 6.66 \\
ADRS & 0.83 & 5 & 0.83 & 0.83 & 0.00 & 0.80 & 6.77 & 8.01 \\
ICSD & 0.83 & 5 & 0.83 & 0.83 & 0.00 & 0.73 & 6.11 & 7.43 \\
PCSD & 0.83 & 5 & 0.83 & 0.83 & 0.00 & 0.75 & 6.86 & 8.16 \\
\midrule
\multicolumn{9}{l}{\textbf{Qwen3.5-4B}} \\
GRPO & 33.37 & 60 & 31.11 & 22.59 & 2.25 & 21.07 & 8.62 & 11.24 \\
GRPO+OPSD & 13.86 & 90 & 13.66 & 13.71 & 0.20 & 12.87 & 9.02 & 10.87 \\
SDAR & 33.40 & 25 & 13.86 & 13.79 & 19.54 & 16.55 & 9.66 & 12.14 \\
Skill-SD & 44.73 & 95 & 42.15 & 42.64 & 2.59 & 34.28 & 8.25 & 10.19 \\
RLSD & 52.12 & 65 & 22.43 & 20.56 & 29.69 & 35.43 & 9.38 & 11.34 \\
OCSD & 48.02 & 75 & 41.67 & 35.25 & 6.34 & 36.12 & 9.04 & 11.29 \\
OVCSD & 41.73 & 85 & 15.54 & 32.36 & 26.20 & 31.54 & 11.39 & 14.04 \\
AgentOPSD & 36.20 & 50 & 35.01 & 31.28 & 1.20 & 26.77 & 9.84 & 11.72 \\
ADRS & 37.90 & 70 & 36.95 & 31.97 & 0.95 & 27.12 & 9.37 & 11.21 \\
ICSD & 16.27 & 75 & 15.86 & 15.29 & 0.41 & 14.46 & 10.72 & 12.79 \\
PCSD & 41.63 & 65 & 23.00 & 29.79 & 18.63 & 28.74 & 9.63 & 11.55 \\
\midrule
\multicolumn{9}{l}{\textbf{Qwen3.5-9B}} \\
GRPO & 59.15 & 100 & 59.15 & 53.03 & 0.00 & 38.90 & 11.18 & 14.44 \\
GRPO+OPSD & 54.91 & 60 & 52.11 & 52.59 & 2.80 & 42.98 & 12.65 & 15.94 \\
SDAR & 58.90 & 80 & 57.13 & 55.90 & 1.77 & 46.40 & 12.95 & 16.94 \\
Skill-SD & 38.34 & 50 & 28.84 & 33.60 & 9.50 & 30.04 & 13.19 & 16.84 \\
RLSD & 50.36 & 100 & 50.36 & 46.48 & 0.00 & 38.30 & 12.86 & 16.52 \\
OCSD & 55.11 & 60 & 45.36 & 47.74 & 9.75 & 41.52 & 11.39 & 14.29 \\
OVCSD & 52.73 & 95 & 45.09 & 45.82 & 7.63 & 32.70 & 16.50 & 20.26 \\
AgentOPSD & 50.61 & 45 & 42.58 & 30.22 & 8.03 & 37.53 & 12.44 & 15.56 \\
ADRS & 58.41 & 65 & 37.83 & 46.33 & 20.58 & 44.49 & 13.16 & 16.44 \\
ICSD & 55.23 & 30 & 18.77 & 26.99 & 36.46 & 36.00 & 14.07 & 17.15 \\
PCSD & 58.33 & 100 & 58.33 & 55.95 & 0.00 & 44.33 & 12.67 & 15.99 \\
\bottomrule
\end{tabular}
\caption{ScienceWorld training profiles for all 33 runs, using task score. Peak includes initialization; Step gives its first occurrence; Final is step-100 validation; Tail-3 averages steps 90, 95, and 100; Drop is Peak minus Final. AUC is normalized by the training horizon. Train time excludes validation and saving; Total includes both.}
\label{tab:full-training-runs-scienceworld}
\end{table*}

%% file: table/generated_full_training_runs_search.tex
\begin{table*}[t]
\centering\small
\setlength{\tabcolsep}{4pt}
\begin{tabular}{@{}lrrrrrrrr@{}}
\toprule
Method & Peak & Step & Final & Tail-3 & Drop & AUC & Train h & Total h \\
\midrule
\multicolumn{9}{l}{\textbf{Qwen3.5-2B}} \\
GRPO & 33.73 & 50 & 32.03 & 32.17 & 1.70 & 29.56 & 2.00 & 2.64 \\
GRPO+OPSD & 35.01 & 90 & 34.24 & 34.51 & 0.77 & 26.34 & 2.03 & 2.54 \\
SDAR & 38.40 & 75 & 38.02 & 37.46 & 0.38 & 32.49 & 2.90 & 3.77 \\
Skill-SD & 38.08 & 90 & 37.57 & 37.86 & 0.51 & 28.90 & 2.56 & 3.43 \\
RLSD & 39.74 & 75 & 38.02 & 38.50 & 1.73 & 35.58 & 3.40 & 4.48 \\
OCSD & 15.58 & 45 & 13.31 & 13.49 & 2.27 & 14.38 & 0.97 & 1.14 \\
OVCSD & 38.02 & 70 & 1.34 & 4.81 & 36.67 & 27.84 & 3.54 & 4.60 \\
AgentOPSD & 37.92 & 60 & 34.43 & 35.40 & 3.49 & 27.87 & 2.58 & 3.37 \\
ADRS & 16.61 & 25 & 14.18 & 14.18 & 2.43 & 14.55 & 1.13 & 1.34 \\
ICSD & 39.52 & 65 & 31.49 & 26.31 & 8.03 & 32.10 & 3.47 & 4.45 \\
PCSD & 34.05 & 85 & 31.71 & 32.92 & 2.34 & 30.17 & 2.36 & 2.99 \\
\midrule
\multicolumn{9}{l}{\textbf{Qwen3.5-4B}} \\
GRPO & 45.12 & 100 & 45.12 & 44.11 & 0.00 & 40.86 & 3.99 & 5.41 \\
GRPO+OPSD & 44.42 & 75 & 43.81 & 43.84 & 0.61 & 40.69 & 4.67 & 6.14 \\
SDAR & 45.57 & 100 & 45.57 & 44.05 & 0.00 & 40.14 & 4.29 & 5.45 \\
Skill-SD & 43.74 & 95 & 43.30 & 43.45 & 0.45 & 39.62 & 4.76 & 6.28 \\
RLSD & 43.23 & 100 & 43.23 & 42.46 & 0.00 & 39.35 & 5.00 & 6.56 \\
OCSD & 44.35 & 95 & 44.19 & 43.85 & 0.16 & 39.27 & 4.08 & 5.55 \\
OVCSD & 44.16 & 80 & 43.65 & 43.52 & 0.51 & 40.44 & 4.91 & 6.46 \\
AgentOPSD & 44.42 & 100 & 44.42 & 43.99 & 0.00 & 40.90 & 4.82 & 6.30 \\
ADRS & 42.75 & 100 & 42.75 & 41.97 & 0.00 & 38.96 & 4.60 & 6.06 \\
ICSD & 46.14 & 100 & 46.14 & 45.94 & 0.00 & 41.83 & 5.02 & 6.25 \\
PCSD & 44.16 & 65 & 43.55 & 43.73 & 0.61 & 39.28 & 4.24 & 5.57 \\
\midrule
\multicolumn{9}{l}{\textbf{Qwen3.5-9B}} \\
GRPO & 47.62 & 95 & 46.50 & 46.46 & 1.12 & 41.28 & 5.93 & 8.22 \\
GRPO+OPSD & 47.55 & 100 & 47.55 & 46.93 & 0.00 & 42.33 & 7.05 & 9.38 \\
SDAR & 45.15 & 95 & 45.15 & 44.64 & 0.00 & 41.11 & 6.85 & 9.06 \\
Skill-SD & 46.50 & 100 & 46.50 & 45.69 & 0.00 & 41.87 & 6.39 & 8.09 \\
RLSD & 45.25 & 80 & 44.22 & 44.04 & 1.02 & 40.60 & 7.46 & 9.83 \\
OCSD & 47.78 & 100 & 47.78 & 47.18 & 0.00 & 40.37 & 6.04 & 8.26 \\
OVCSD & 45.79 & 100 & 45.79 & 45.00 & 0.00 & 41.40 & 7.28 & 9.75 \\
AgentOPSD & 45.98 & 75 & 45.41 & 45.06 & 0.58 & 40.50 & 6.18 & 7.96 \\
ADRS & 44.19 & 100 & 44.19 & 43.61 & 0.00 & 38.85 & 6.99 & 9.27 \\
ICSD & 46.59 & 100 & 46.59 & 46.10 & 0.00 & 41.44 & 7.70 & 9.96 \\
PCSD & 45.76 & 80 & 44.99 & 44.92 & 0.77 & 40.57 & 6.64 & 8.60 \\
\bottomrule
\end{tabular}
\caption{Search-QA training profiles for all 33 runs, using exact-match accuracy (\%). Peak includes initialization; Step gives its first occurrence; Final is step-100 validation; Tail-3 averages steps 90, 95, and 100; Drop is Peak minus Final. AUC is normalized by the training horizon. Train time excludes validation and saving; Total includes both.}
\label{tab:full-training-runs-search}
\end{table*}

%% file: table/generated_full_training_runs_webshop.tex
\begin{table*}[t]
\centering\small
\setlength{\tabcolsep}{4pt}
\begin{tabular}{@{}lrrrrrrrr@{}}
\toprule
Method & Peak & Step & Final & Tail-3 & Drop & AUC & Train h & Total h \\
\midrule
\multicolumn{9}{l}{\textbf{Qwen3.5-2B}} \\
GRPO & 70.30 & 100 & 70.30 & 67.17 & 0.00 & 41.19 & 3.89 & 4.23 \\
GRPO+OPSD & 77.30 & 85 & 75.80 & 67.20 & 1.50 & 53.47 & 3.87 & 4.15 \\
SDAR & 71.90 & 85 & 64.80 & 63.53 & 7.10 & 46.06 & 5.05 & 5.36 \\
Skill-SD & 75.80 & 85 & 70.30 & 71.07 & 5.50 & 43.04 & 4.22 & 4.52 \\
RLSD & 69.50 & 85 & 68.80 & 68.53 & 0.70 & 37.68 & 4.39 & 4.69 \\
OCSD & 75.00 & 100 & 75.00 & 36.43 & 0.00 & 49.91 & 3.73 & 4.07 \\
OVCSD & 2.30 & 0 & 0.80 & 0.53 & 1.50 & 0.86 & 5.23 & 5.62 \\
AgentOPSD & 71.90 & 100 & 71.90 & 70.07 & 0.00 & 36.50 & 4.70 & 5.03 \\
ADRS & 64.80 & 100 & 64.80 & 63.53 & 0.00 & 30.63 & 4.22 & 4.55 \\
ICSD & 71.10 & 100 & 71.10 & 69.27 & 0.00 & 36.15 & 4.38 & 4.70 \\
PCSD & 75.80 & 100 & 75.80 & 71.10 & 0.00 & 40.34 & 4.68 & 5.01 \\
\midrule
\multicolumn{9}{l}{\textbf{Qwen3.5-4B}} \\
GRPO & 68.80 & 80 & 64.80 & 64.07 & 4.00 & 45.70 & 5.36 & 5.88 \\
GRPO+OPSD & 79.70 & 90 & 77.30 & 76.30 & 2.40 & 50.06 & 6.48 & 7.00 \\
SDAR & 60.20 & 95 & 50.80 & 52.10 & 9.40 & 39.55 & 7.18 & 7.76 \\
Skill-SD & 5.50 & 0 & 0.00 & 0.00 & 5.50 & 0.18 & 7.03 & 7.40 \\
RLSD & 73.40 & 70 & 66.40 & 67.20 & 7.00 & 54.71 & 6.32 & 6.87 \\
OCSD & 68.00 & 65 & 64.10 & 65.13 & 3.90 & 52.07 & 5.41 & 5.95 \\
OVCSD & 75.80 & 90 & 73.40 & 74.73 & 2.40 & 50.76 & 6.50 & 7.09 \\
AgentOPSD & 74.20 & 100 & 74.20 & 72.13 & 0.00 & 43.05 & 6.45 & 7.02 \\
ADRS & 68.80 & 95 & 66.40 & 67.47 & 2.40 & 37.03 & 6.56 & 7.11 \\
ICSD & 3.90 & 0 & 0.00 & 0.00 & 3.90 & 0.26 & 8.63 & 9.05 \\
PCSD & 70.30 & 100 & 70.30 & 69.00 & 0.00 & 37.05 & 6.68 & 7.17 \\
\midrule
\multicolumn{9}{l}{\textbf{Qwen3.5-9B}} \\
GRPO & 71.90 & 80 & 69.50 & 68.73 & 2.40 & 48.77 & 8.09 & 8.89 \\
GRPO+OPSD & 76.60 & 100 & 76.60 & 74.50 & 0.00 & 50.69 & 9.13 & 9.94 \\
SDAR & 74.20 & 100 & 74.20 & 71.87 & 0.00 & 53.89 & 8.57 & 9.38 \\
Skill-SD & 7.80 & 0 & 0.00 & 0.00 & 7.80 & 0.20 & 9.87 & 10.44 \\
RLSD & 6.20 & 0 & 0.00 & 0.00 & 6.20 & 0.15 & 9.66 & 10.25 \\
OCSD & 74.20 & 85 & 68.80 & 70.87 & 5.40 & 43.90 & 8.75 & 9.55 \\
OVCSD & 75.00 & 85 & 69.50 & 72.63 & 5.50 & 52.61 & 9.29 & 10.14 \\
AgentOPSD & 75.00 & 90 & 69.50 & 72.40 & 5.50 & 47.46 & 9.45 & 10.27 \\
ADRS & 69.50 & 100 & 69.50 & 67.70 & 0.00 & 46.36 & 8.90 & 9.63 \\
ICSD & 73.40 & 100 & 73.40 & 72.13 & 0.00 & 51.31 & 9.84 & 10.56 \\
PCSD & 75.80 & 95 & 73.40 & 74.47 & 2.40 & 57.55 & 8.91 & 9.74 \\
\bottomrule
\end{tabular}
\caption{WebShop training profiles for all 33 runs, using success (\%). Peak includes initialization; Step gives its first occurrence; Final is step-100 validation; Tail-3 averages steps 90, 95, and 100; Drop is Peak minus Final. AUC is normalized by the training horizon. Train time excludes validation and saving; Total includes both.}
\label{tab:full-training-runs-webshop}
\end{table*}

%% file: table/generated_full_failure_cases.tex
\begin{table*}[t]
\centering\small
\setlength{\tabcolsep}{4pt}
\begin{tabular}{@{}lrrrrrrr@{}}
\toprule
Model / method & Initial & Peak & Step & Val. & Eval. & Turns/limit & NF \\
\midrule
\multicolumn{8}{l}{\textbf{ALFWorld}} \\
4B GRPO+OPSD & 33.60 & 39.30 & 20 & 2.10 & 1.52 & 49.01/50 & 0 \\
\midrule
\multicolumn{8}{l}{\textbf{ScienceWorld}} \\
2B GRPO & -2.94 & 0.83 & 5 & 0.83 & 0.83 & 17.38/20 & 0 \\
2B GRPO+OPSD & -0.94 & 0.83 & 5 & 0.83 & 0.83 & 15.60/20 & 0 \\
2B SDAR & -2.77 & 1.08 & 5 & 0.83 & 0.83 & 16.71/20 & 0 \\
2B Skill-SD & -1.97 & 0.83 & 5 & 0.83 & 0.83 & 12.88/20 & 39 \\
2B RLSD & -2.51 & 2.39 & 95 & 1.89 & 1.89 & 20.00/20 & 0 \\
2B AgentOPSD & -1.73 & 0.83 & 5 & 0.83 & 0.83 & 19.52/20 & 0 \\
2B ADRS & -0.42 & 0.83 & 5 & 0.83 & 0.83 & 19.99/20 & 0 \\
2B ICSD & -3.23 & 0.83 & 5 & 0.83 & 0.83 & 14.28/20 & 0 \\
2B PCSD & -2.51 & 0.83 & 5 & 0.83 & 0.83 & 20.00/20 & 0 \\
\midrule
\multicolumn{8}{l}{\textbf{Search-QA}} \\
2B OVCSD & 13.95 & 38.02 & 70 & 1.34 & 33.21 & 2.48/4 & 0 \\
\midrule
\multicolumn{8}{l}{\textbf{WebShop}} \\
2B OVCSD & 2.30 & 2.30 & 0 & 0.80 & 1.07 & 14.99/15 & 0 \\
4B Skill-SD & 5.50 & 5.50 & 0 & 0.00 & 0.00 & 15.00/15 & 1 \\
4B ICSD & 3.90 & 3.90 & 0 & 0.00 & 0.00 & 15.00/15 & 0 \\
9B Skill-SD & 7.80 & 7.80 & 0 & 0.00 & 0.00 & 15.00/15 & 68 \\
9B RLSD & 6.20 & 6.20 & 0 & 0.00 & 0.00 & 15.00/15 & 0 \\
\bottomrule
\end{tabular}
\caption{All runs with final validation performance below 5 across the four benchmarks. Initial, Peak, and Val. use the training-validation metric; Eval. is the repeated final-evaluation mean. Turns/limit averages training steps 81--100 relative to the interaction limit; NF counts steps with nonfinite actor gradients. Low scores include stagnation as well as lost performance.}
\label{tab:full-failure-cases}
\end{table*}

%% file: figure/phase_cost_breakdown.tex
\begin{figure*}[!t]
\centering
\includegraphics[width=\textwidth]{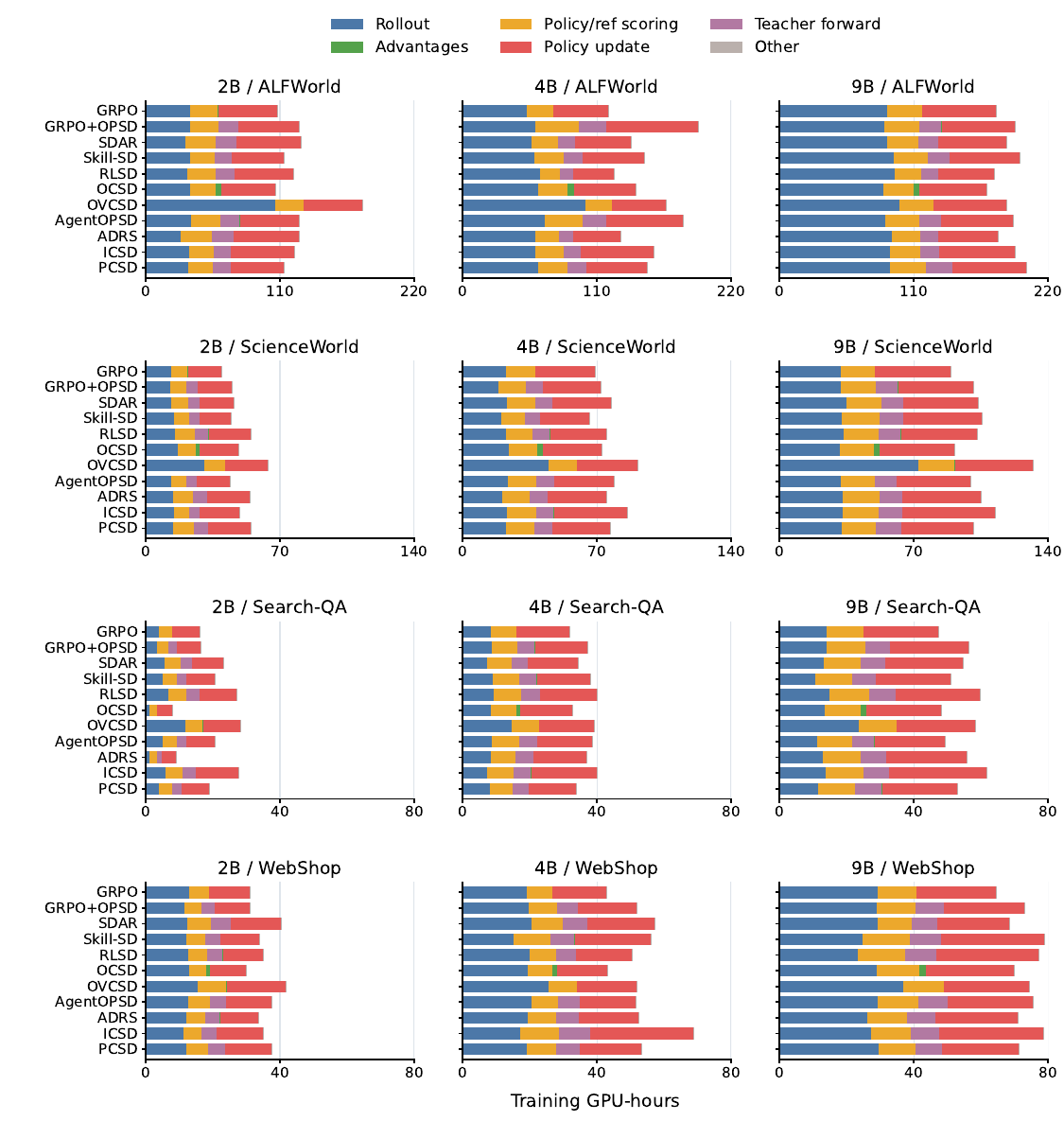}
\caption{Training cost by stage across all eleven methods and twelve settings. Policy/ref scoring combines old-policy and reference log-probability computation; Other combines reward computation and the remaining training time. Teacher forward denotes separately timed teacher passes; teacher-related work may also occur in other stages. Axis ranges are shared within each benchmark. Validation and checkpoint saving are excluded.}
\label{fig:phase-cost}
\end{figure*}

%% file: table/generated_cost_performance_settings.tex
\begin{table*}[!t]
\centering\small
\setlength{\tabcolsep}{4pt}
\begin{tabular}{@{}l*{4}{rr}@{}}
\toprule
Method & \multicolumn{2}{c}{ALFWorld} & \multicolumn{2}{c}{ScienceWorld} & \multicolumn{2}{c}{Search-QA} & \multicolumn{2}{c}{WebShop} \\
\cmidrule(lr){2-3}\cmidrule(lr){4-5}\cmidrule(lr){6-7}\cmidrule(lr){8-9}
 & Perf. & Time (h) & Perf. & Time (h) & Perf. & Time (h) & Perf. & Time (h) \\
\midrule
\multicolumn{9}{l}{\textbf{Qwen3.5-2B}} \\
GRPO & 28.48 & 13.49 & 0.83 & 4.95 & 32.38 & 2.00 & 70.80 & 3.89 \\
GRPO+OPSD & 22.23 & 15.68 & 0.83 & 5.62 & 34.22 & 2.03 & 77.05 & 3.87 \\
SDAR & 14.46 & 15.90 & 0.83 & 5.74 & 38.15 & 2.90 & 63.67 & 5.05 \\
Skill-SD & 74.29 & 14.18 & 0.83 & 5.53 & 37.56 & 2.56 & 71.19 & 4.22 \\
RLSD & 46.61 & 15.18 & 1.89 & 6.87 & 37.93 & 3.40 & 69.73 & 4.39 \\
OCSD & 42.32 & 13.28 & 13.78 & 6.04 & 12.96 & 0.97 & 75.59 & 3.73 \\
OVCSD & 56.96 & 22.23 & 13.85 & 7.98 & 33.21 & 3.54 & 1.07 & 5.23 \\
AgentOPSD & 47.50 & 15.73 & 0.83 & 5.48 & 34.49 & 2.58 & 70.51 & 4.70 \\
ADRS & 13.04 & 15.69 & 0.83 & 6.77 & 14.10 & 1.13 & 63.09 & 4.22 \\
ICSD & 54.73 & 15.26 & 0.83 & 6.11 & 39.39 & 3.47 & 70.51 & 4.38 \\
PCSD & 72.32 & 14.12 & 0.83 & 6.86 & 33.18 & 2.36 & 74.71 & 4.68 \\
\midrule
\multicolumn{9}{l}{\textbf{Qwen3.5-4B}} \\
GRPO & 82.05 & 14.94 & 32.16 & 8.62 & 45.15 & 3.99 & 66.02 & 5.36 \\
GRPO+OPSD & 1.52 & 24.14 & 13.65 & 9.02 & 44.16 & 4.67 & 77.15 & 6.48 \\
SDAR & 76.16 & 17.27 & 13.86 & 9.66 & 44.96 & 4.29 & 51.17 & 7.18 \\
Skill-SD & 55.27 & 18.61 & 41.67 & 8.25 & 42.81 & 4.76 & 0.00 & 7.03 \\
RLSD & 84.38 & 15.54 & 22.41 & 9.38 & 43.60 & 5.00 & 68.16 & 6.32 \\
OCSD & 52.32 & 17.75 & 41.77 & 9.04 & 44.73 & 4.08 & 64.16 & 5.41 \\
OVCSD & 76.43 & 20.81 & 14.93 & 11.39 & 43.51 & 4.91 & 73.73 & 6.50 \\
AgentOPSD & 75.00 & 22.62 & 32.80 & 9.84 & 44.25 & 4.82 & 72.07 & 6.45 \\
ADRS & 70.71 & 16.20 & 36.73 & 9.37 & 42.31 & 4.60 & 65.72 & 6.56 \\
ICSD & 82.50 & 19.57 & 15.99 & 10.72 & 45.87 & 5.02 & 0.00 & 8.63 \\
PCSD & 79.82 & 18.92 & 23.85 & 9.63 & 43.11 & 4.24 & 72.17 & 6.68 \\
\midrule
\multicolumn{9}{l}{\textbf{Qwen3.5-9B}} \\
GRPO & 62.68 & 22.16 & 58.99 & 11.18 & 46.59 & 5.93 & 70.02 & 8.09 \\
GRPO+OPSD & 71.79 & 24.16 & 51.98 & 12.65 & 47.08 & 7.05 & 74.90 & 9.13 \\
SDAR & 75.62 & 23.21 & 58.00 & 12.95 & 44.75 & 6.85 & 74.32 & 8.57 \\
Skill-SD & 74.55 & 24.65 & 29.08 & 13.19 & 46.19 & 6.39 & 0.00 & 9.87 \\
RLSD & 89.20 & 22.00 & 50.27 & 12.86 & 44.31 & 7.46 & 0.00 & 9.66 \\
OCSD & 86.16 & 21.19 & 44.30 & 11.39 & 48.16 & 6.04 & 67.77 & 8.75 \\
OVCSD & 83.39 & 23.22 & 43.45 & 16.50 & 45.43 & 7.28 & 68.07 & 9.29 \\
AgentOPSD & 78.04 & 23.94 & 42.48 & 12.44 & 45.81 & 6.18 & 72.27 & 9.45 \\
ADRS & 78.93 & 22.39 & 35.63 & 13.16 & 43.93 & 6.99 & 70.31 & 8.90 \\
ICSD & 89.64 & 24.13 & 18.26 & 14.07 & 46.84 & 7.70 & 72.46 & 9.84 \\
PCSD & 72.77 & 25.25 & 58.60 & 12.67 & 45.42 & 6.64 & 74.32 & 8.91 \\
\bottomrule
\end{tabular}
\caption{Numerical values for the points in Figure~\ref{fig:cost-performance-settings}. Perf. is mean final performance over eight rollout seeds, using the benchmark metrics in Tables~\ref{tab:main_results} and~\ref{tab:main_results_search_webshop}; Time is net training time in hours.}
\label{tab:cost-performance-settings}
\end{table*}

%% file: table/generated_full_performance_cost.tex
\begin{table}[!htbp]
\centering\small
\begin{tabular}{lrrr}
\toprule
Method & Mean & Time ($\times$) & GPU-h \\
\midrule
GRPO & 49.68 & 1.000 & 836.7 \\
GRPO+OPSD & 43.05 & 1.158 & 996.1 \\
SDAR & 46.33 & 1.183 & 956.5 \\
Skill-SD & 39.45 & 1.153 & 954.0 \\
RLSD & 46.54 & 1.208 & 944.5 \\
OCSD & 49.50 & 0.999 & 861.2 \\
OVCSD & 46.17 & 1.370 & 1111.1 \\
AgentOPSD & 51.34 & 1.187 & 993.9 \\
ADRS & 44.61 & 1.100 & 927.8 \\
ICSD & 44.75 & 1.293 & 1031.2 \\
PCSD & 54.26 & 1.167 & 967.6 \\
\bottomrule
\end{tabular}
\caption{Final performance and training cost across twelve settings. Mean uses the benchmark metrics in Tables~\ref{tab:main_results} and~\ref{tab:main_results_search_webshop}; Time averages paired training-time ratios to GRPO. GPU-h sums twelve training runs, excluding validation and saving.}
\label{tab:performance_cost}
\end{table}

%% file: table/generated_full_budget_aligned.tex
\begin{table*}[!t]
\centering\small
\begin{tabular}{lrrrrrrrr}
\toprule
Method & \multicolumn{2}{c}{ALFWorld} & \multicolumn{2}{c}{ScienceWorld} & \multicolumn{2}{c}{Search-QA} & \multicolumn{2}{c}{WebShop} \\
\cmidrule(lr){2-3}\cmidrule(lr){4-5}\cmidrule(lr){6-7}\cmidrule(lr){8-9}
 & Step 100 & Budget & Step 100 & Budget & Step 100 & Budget & Step 100 & Budget \\
\midrule
GRPO & 58.10 & 58.10 & 30.36 & 30.36 & 41.22 & 41.22 & 68.20 & 68.20 \\
GRPO+OPSD & 32.13 & 31.67 & 22.20 & 21.74 & 41.87 & 41.85 & 76.57 & 72.13 \\
SDAR & 54.50 & 54.50 & 23.94 & 21.82 & 42.91 & 41.15 & 63.27 & 61.70 \\
Skill-SD & 67.60 & 66.70 & 23.94 & 27.00 & 42.45 & 41.61 & 23.43 & 23.17 \\
RLSD & 72.60 & 77.37 & 24.89 & 22.67 & 41.82 & 41.50 & 45.07 & 47.13 \\
OCSD & 58.57 & 54.07 & 33.63 & 30.71 & 35.09 & 34.91 & 69.30 & 71.63 \\
OVCSD & 72.60 & 59.77 & 24.78 & 18.99 & 30.26 & 42.04 & 47.90 & 44.80 \\
AgentOPSD & 68.10 & 37.63 & 26.14 & 24.87 & 41.42 & 41.83 & 71.87 & 70.83 \\
ADRS & 55.70 & 54.07 & 25.20 & 30.37 & 33.71 & 32.59 & 66.90 & 61.20 \\
ICSD & 76.17 & 77.37 & 11.82 & 19.51 & 41.41 & 43.07 & 48.17 & 42.97 \\
PCSD & 74.30 & 68.07 & 27.39 & 27.46 & 40.09 & 40.96 & 73.17 & 63.83 \\
\bottomrule
\end{tabular}
\caption{Validation performance at step 100 and within the matched GRPO training-time budget, averaged over the three model sizes separately for each benchmark. Budget selects the latest logged validation reached within each setting\textquotesingle s budget. ScienceWorld reports task score; the other benchmarks report success or exact-match accuracy (\%). Each comparison retains its benchmark\textquotesingle s validation decoder.}
\label{tab:grpo-time-budget}
\end{table*}

%% file: table/generated_full_compute_ledger.tex
\begin{table*}[!t]
\centering\small
\begin{tabular}{lrrrrr}
\toprule
Group & Runs & Training & Validation & Saving & Total \\
\midrule
All runs & 132 & 10580.7 & 1515.3 & 28.4 & 12124.4 \\
\midrule
Qwen3.5-2B & 44 & 2511.6 & 282.0 & 5.3 & 2798.9 \\
Qwen3.5-4B & 44 & 3474.4 & 484.7 & 9.2 & 3968.4 \\
Qwen3.5-9B & 44 & 4594.6 & 748.6 & 13.8 & 5357.1 \\
\midrule
ALFWorld & 33 & 5067.3 & 408.3 & 12.2 & 5487.8 \\
ScienceWorld & 33 & 2527.5 & 602.6 & 1.0 & 3131.1 \\
Search-QA & 33 & 1214.8 & 377.5 & 1.4 & 1593.6 \\
WebShop & 33 & 1771.2 & 126.9 & 13.8 & 1911.9 \\
\bottomrule
\end{tabular}
\caption{Recorded H800 GPU-hours for all 132 training runs across four benchmarks. Model-size and benchmark blocks separately partition the same total. Training includes rollout and teacher computation; Total additionally includes in-loop validation and saving.}
\label{tab:compute_ledger}
\end{table*}

%% file: sec/18_computational_cost.tex
\section{Computational Cost Analysis}
\label{app:logged-compute}


This appendix analyzes computational cost for all \PaperCompletedRuns{} runs across four benchmarks, complementing the per-setting comparisons in Figure~\ref{fig:cost-performance-settings}.
Final-performance comparisons use the repeated evaluations; time-budget comparisons select observed points from each benchmark's training-validation trajectories.

\subsection{Performance and Time Budgets}
\label{app:performance-time-budgets}

Computational cost changes the practical value of a performance gain.
We assess both the resources used to obtain the final agent and the performance reached within a fixed training-time budget.
For the former, we pair repeated-evaluation performance with net training time, average the within-setting time ratios to GRPO, and sum GPU-hours across each method's twelve runs.
Nine alternatives have higher mean paired training time than GRPO; OCSD is approximately tied at 0.999$\times$.
Only PCSD and AgentOPSD improve mean final performance, with time ratios of 1.17 and 1.19.
PCSD's larger aggregate gain comes at a smaller average relative overhead, whereas OVCSD has the largest time ratio (1.37) without improving aggregate performance.
Table~\ref{tab:cost-performance-settings} lists the performance and training-time values for every point in Figure~\ref{fig:cost-performance-settings}, allowing direct comparisons where markers overlap; Table~\ref{tab:performance_cost} summarizes these trade-offs across settings.

\noindent \textbf{Performance within GRPO's training-time budget.}
We use the net training time of GRPO as the budget for each model--task setting.
For every method, we select the latest evaluated checkpoint reached within this budget; methods that finish earlier use their final checkpoint.
The fixed-step and time-budget scores are drawn from the same validation trajectory.
For each of the four benchmarks, we report the mean across the three model sizes using that benchmark's performance metric.
AgentOPSD loses its ALFWorld advantage and PCSD loses its WebShop advantage under this rule, while OCSD retains a much smaller ScienceWorld advantage.
Earlier checkpoints can also outperform the endpoint: 2B OVCSD on Search-QA reaches step 55 within budget, scoring 36.93\% before its late greedy-decoding collapse.
Time-budget assessment therefore reflects both the rate of acquiring performance and the ability to retain it.
Table~\ref{tab:grpo-time-budget} reports both scores for every method and benchmark.

\subsection{Total Compute and Stage-Level Costs}
\label{app:compute-breakdown}

\noindent \textbf{GPU-hours and training workload.}
Across \PaperCompletedRuns{} runs, the study records \PaperCompletedTrainSteps{} training steps and \PaperCompletedValidations{} validation sweeps, including initial validation.
The training loops consume \PaperAllRecordedGPUHours{} H800 GPU-hours, computed as elapsed hours multiplied by the eight GPUs assigned to each run and summed across runs.
Training accounts for \PaperAllTrainingGPUHours{} GPU-hours, in-loop validation for \PaperAllValidationGPUHours{}, and checkpoint saving for \PaperAllCheckpointGPUHours{}.
This total separates training, in-loop validation, and checkpoint saving; it excludes initialization, initial validation, and subsequent repeated evaluation.
Table~\ref{tab:compute_ledger} reports the breakdown by model size and benchmark, distinguishing the experimental expenditure from the net training cost used in method comparisons.

\noindent \textbf{Distillation overhead extends beyond teacher forward passes.}
Teacher supervision introduces work at different stages of training, depending on how each method uses it.
Across all twelve settings, OVCSD's verified continuations contribute to rollout generation, whose mean time is 218.0 seconds per step versus GRPO's 124.9; OCSD's paired teacher replay contributes to advantage computation, averaging 10.6 seconds versus 0.2 for GRPO.
These measurements show why separately timed teacher passes cannot represent the full cost of incorporating teacher information.
Efficiency should be judged using end-to-end training time, with stage-level measurements used to locate the additional work.
Figure~\ref{fig:phase-cost} decomposes the absolute GPU-hours into rollout generation, policy/reference scoring, teacher forward passes, advantage computation, policy updates, and other processing.

%% file: sec/19_literature_survey.tex
\section{Literature Survey}
\label{app:literature-inventory}

\subsection{Search Strategy and Selection Criteria}
\label{app:survey-method}

\noindent \textbf{Scope.}
Our survey characterizes recent research on distillation for training multi-turn language-model agents that interact through tools or environments.
It covers text and GUI agents trained with privileged-context supervision, other on-policy teacher distillation, or offline and iterative trajectory distillation, together with empirical analyses of these approaches.
The common focus is the use of distillation to update the agent's action policy.
Within this broader landscape, our benchmark compares GRPO with ten representative methods whose main contributions target specific components of text-agent training, including auxiliary objectives, credit assignment, reward shaping, and verified teacher continuations (Appendix~\ref{app:method-scope}).

\noindent \textbf{Search and coverage.}
We examine first submissions from 25 June to 22 September 2026 (UTC), a 90-day period, using arXiv metadata retrieved on 23 September.
Two title-and-abstract searches pair agent-related terms with distillation, and self-distillation with reinforcement learning.
Deduplication and date filtering yield 635 records, from which the agent--distillation query and interactive-agent keywords identify 84 candidates.
Reviewing these 84 candidates' titles and abstracts, with full-text inspection where needed, yields \SurveyPaperCount{} papers that meet the survey scope and form the final inventory.
As a coverage check, we verify that it includes every source paper of the evaluated methods first submitted within this period.

\subsection{Publication Statistics}
\label{app:survey-statistics}

Counting each unique arXiv paper once by its first-submission date gives \SurveyPaperCount{} papers over \SurveyDayCount{} days, or approximately one paper every two days ($\SurveyDailyRatePrecise$ per day).
The inventory includes \SurveyPrivilegedCount{} papers on privileged-context policy supervision (P), 10 on other on-policy distillation and related empirical frameworks (O), and 5 on trajectory-based policy distillation (T).
The full inventory, with titles, first-submission dates, supervision categories, and arXiv links, appears in Tables~\ref{tab:arxiv-inventory-1}--\ref{tab:arxiv-inventory-4}.

\input{table/arxiv_inventory}

%% file: table/arxiv_inventory.tex
\begin{table*}[tp]
\centering
\small
\begin{tabular}{@{}p{0.105\textwidth}p{0.118\textwidth}p{0.64\textwidth}l@{}}
\toprule
Submitted & arXiv ID & Title & Route \\
\midrule
2026-06-25 & \href{https://arxiv.org/abs/2606.26790}{2606.26790} & OPID: On-Policy Skill Distillation for Agentic Reinforcement Learning & P \\
2026-06-26 & \href{https://arxiv.org/abs/2606.27814}{2606.27814} & ATOD: Annealed Turn-Aware On-Policy Distillation for Multi-Turn Agentic Tasks & O \\
2026-06-28 & \href{https://arxiv.org/abs/2606.29476}{2606.29476} & CRAFT: Counterfactual Credit Assignment from Free Sibling Rollouts for Self-Distilled Agentic Reinforcement Learning & P \\
2026-06-28 & \href{https://arxiv.org/abs/2606.29502}{2606.29502} & UCOB: Learning to Utilize and Evolve Agentic Skills via Credit-Aware On-Policy Bidirectional Self-Distillation & P \\
2026-06-29 & \href{https://arxiv.org/abs/2606.29863}{2606.29863} & KbSD: Knowledge Boundary aware Self-Distillation for Behavioral Calibration in Agentic Search & P \\
2026-06-29 & \href{https://arxiv.org/abs/2606.29961}{2606.29961} & DuoMem: Towards Capable On-Device Memory Agents via Dual-Space Distillation & T \\
2026-06-29 & \href{https://arxiv.org/abs/2606.30044}{2606.30044} & Building Multi-Task Agentic LLMs via Two-Phase Distillation & O \\
2026-07-05 & \href{https://arxiv.org/abs/2607.04425}{2607.04425} & UI-MOPD: Multi-Platform On-Policy Distillation for Unified GUI Agents & O \\
2026-07-07 & \href{https://arxiv.org/abs/2607.05804}{2607.05804} & TurnOPD: Making On-Policy Distillation Turn-Aware for Efficient Long-Horizon Agent Training & O \\
2026-07-08 & \href{https://arxiv.org/abs/2607.07820}{2607.07820} & DeepSearch-World: Self-Distillation for Deep Search Agents in a Verifiable Environment & T \\
2026-07-13 & \href{https://arxiv.org/abs/2607.11012}{2607.11012} & EasyOPD: An Easy-to-use On-Policy Distillation Framework for Large Language Models & O \\
2026-07-16 & \href{https://arxiv.org/abs/2607.14777}{2607.14777} & SEED: Self-Evolving On-Policy Distillation for Agentic Reinforcement Learning & P \\
\bottomrule
\end{tabular}
\caption{Papers included in the literature survey (1/4). Dates indicate first submissions (UTC). P: privileged-context supervision; O: other on-policy distillation; T: trajectory distillation.}
\label{tab:arxiv-inventory-1}
\end{table*}

\begin{table*}[tp]
\centering
\small
\begin{tabular}{@{}p{0.105\textwidth}p{0.118\textwidth}p{0.64\textwidth}l@{}}
\toprule
Submitted & arXiv ID & Title & Route \\
\midrule
2026-07-20 & \href{https://arxiv.org/abs/2607.17558}{2607.17558} & Why Does Feedback-Augmented Self-Distillation Fail to Improve Retrieval-Interleaved Search Agents? & P \\
2026-07-27 & \href{https://arxiv.org/abs/2607.24280}{2607.24280} & From Proprietary to Open-Source: Bridging the Distribution Gap via Multi-Agent Protocol Distillation in Agentic Search & P \\
2026-07-27 & \href{https://arxiv.org/abs/2607.24720}{2607.24720} & The Physics of Multi-Turn Long-Horizon Planning: From Pre-training to Post-training via Single- and Multi-Teacher On-Policy Agentic Distillation & O \\
2026-07-30 & \href{https://arxiv.org/abs/2607.27937}{2607.27937} & From Scoring to Acting: Outcome-Verified Comparative Self-Distillation for LLM Agents & P \\
2026-07-30 & \href{https://arxiv.org/abs/2607.28026}{2607.28026} & Contrastive Reinforced Policy Optimization via Privileged Self-Distillation & P \\
2026-07-30 & \href{https://arxiv.org/abs/2607.28076}{2607.28076} & Group-Reflective Self-Distillation for Agentic Reinforcement Learning & P \\
2026-07-31 & \href{https://arxiv.org/abs/2607.29320}{2607.29320} & MAGA: Multi-Platform Self-Fusion of GUI Agents via Structured Action Distillation & O \\
2026-08-01 & \href{https://arxiv.org/abs/2608.00485}{2608.00485} & SERL-SQL: Selective Hindsight Distillation for Text-to-SQL Reinforcement Agentic Learning & P \\
2026-08-02 & \href{https://arxiv.org/abs/2608.01359}{2608.01359} & EviSD: Evidence-Conditioned Self-Distillation for Search-Augmented Agents & P \\
2026-08-03 & \href{https://arxiv.org/abs/2608.01597}{2608.01597} & HindSearch: Trajectory-Level Hindsight Critique for Search-Augmented Reinforcement Learning & P \\
2026-08-03 & \href{https://arxiv.org/abs/2608.01837}{2608.01837} & PCSD: Persistent Consistency for Self-Distillation in Agentic Reinforcement Learning & P \\
2026-08-03 & \href{https://arxiv.org/abs/2608.01953}{2608.01953} & Look Ahead Before You Distill: Future Trajectory Validation of Teacher Guidance for Agentic On-Policy Distillation & O \\
\bottomrule
\end{tabular}
\caption{Papers included in the literature survey (2/4). Dates indicate first submissions (UTC). P: privileged-context supervision; O: other on-policy distillation; T: trajectory distillation.}
\label{tab:arxiv-inventory-2}
\end{table*}

\begin{table*}[tp]
\centering
\small
\begin{tabular}{@{}p{0.105\textwidth}p{0.118\textwidth}p{0.64\textwidth}l@{}}
\toprule
Submitted & arXiv ID & Title & Route \\
\midrule
2026-08-04 & \href{https://arxiv.org/abs/2608.03223}{2608.03223} & Agentic Reinforcement Learning with Self-Distilled Reward Shaping & P \\
2026-08-04 & \href{https://arxiv.org/abs/2608.04007}{2608.04007} & TurnSight: Turn-Level Hindsight Self-Distillation for Tool-Integrated Reasoning & P \\
2026-08-05 & \href{https://arxiv.org/abs/2608.05219}{2608.05219} & When Privileged Guidance Misaligns: State-Matched Routing and Contextualized Self-Distillation for Multi-Turn Agents & P \\
2026-08-05 & \href{https://arxiv.org/abs/2608.04788}{2608.04788} & Agentic Reinforcement Learning with Observation-Calibrated Self-Distillation & P \\
2026-08-05 & \href{https://arxiv.org/abs/2608.04794}{2608.04794} & Privileged, but Biased: How PI-Conditioned Teachers Break Self-Distillation & P \\
2026-08-06 & \href{https://arxiv.org/abs/2608.05987}{2608.05987} & AgentOPSD: Recursive Self-Distillation for Agentic Reinforcement Learning & P \\
2026-08-06 & \href{https://arxiv.org/abs/2608.06065}{2608.06065} & The Next Screenshot Knows: Gated Hindsight Distillation for Mobile GUI Agents & P \\
2026-08-07 & \href{https://arxiv.org/abs/2608.07068}{2608.07068} & MemOPD: On-Policy Distillation through Memory State Alignment for Long-Horizon Agents & O \\
2026-08-07 & \href{https://arxiv.org/abs/2608.07371}{2608.07371} & Trajectory-Relative Hindsight Distillation for Agentic Reinforcement Learning & P \\
2026-08-09 & \href{https://arxiv.org/abs/2608.08570}{2608.08570} & FailForge: Distilling Procedural Competence from Persistent Failures into Code Agents & T \\
2026-08-09 & \href{https://arxiv.org/abs/2608.08960}{2608.08960} & Reading is not Reasoning: Bridging the Agentic Policy Gap in Vision-Text Compression & P \\
2026-08-10 & \href{https://arxiv.org/abs/2608.09555}{2608.09555} & Bidirectional Context Self-Distillation for Reinforcement Learning of Skill-Based LLM Agents & P \\
\bottomrule
\end{tabular}
\caption{Papers included in the literature survey (3/4). Dates indicate first submissions (UTC). P: privileged-context supervision; O: other on-policy distillation; T: trajectory distillation.}
\label{tab:arxiv-inventory-3}
\end{table*}

\begin{table*}[tp]
\centering
\small
\begin{tabular}{@{}p{0.105\textwidth}p{0.118\textwidth}p{0.64\textwidth}l@{}}
\toprule
Submitted & arXiv ID & Title & Route \\
\midrule
2026-08-12 & \href{https://arxiv.org/abs/2608.11967}{2608.11967} & LoongReflect: Boosting Long-Horizon Reflection in Search Agents via Global Perspective Distillation & P \\
2026-08-13 & \href{https://arxiv.org/abs/2608.12764}{2608.12764} & Beyond Outcome Rewards: Step-Level Self-Distilled Policy Optimization for Deep Search Agents & P \\
2026-08-14 & \href{https://arxiv.org/abs/2608.14945}{2608.14945} & Trust Is Not Enough: Influence Calibration for On-Policy Self-Distillation in Agentic RL & P \\
2026-08-19 & \href{https://arxiv.org/abs/2608.18524}{2608.18524} & DART-SD: Diamond-topology Aware Retrieval and Tuning for Self-Distillation of Multi-Turn Tool-Calling Agents & P \\
2026-08-22 & \href{https://arxiv.org/abs/2608.21946}{2608.21946} & EDGE: Experience-Distillation for Guided Exploration in Agentic Reinforcement Learning & P \\
2026-08-25 & \href{https://arxiv.org/abs/2608.24114}{2608.24114} & AHEAD: Adaptive Hindsight with Environment-Augmented Distillation for Agentic RL & P \\
2026-08-29 & \href{https://arxiv.org/abs/2608.29357}{2608.29357} & LiteSearch-VL: Small Multimodal Search Agents via Trajectory Distillation and Synthetic Step-DPO & T \\
2026-08-31 & \href{https://arxiv.org/abs/2608.31077}{2608.31077} & Reconciling Process Supervision with Outcome-Based Credit in Agentic Policy Optimization & P \\
2026-09-13 & \href{https://arxiv.org/abs/2609.14636}{2609.14636} & Know When to Stop, Where to Restart: Accelerating Multi-Turn Agentic On-Policy Distillation & O \\
2026-09-17 & \href{https://arxiv.org/abs/2609.20784}{2609.20784} & RetireOPD: Self-Retiring On-Policy Distillation for Agentic Reinforcement Learning & P \\
2026-09-21 & \href{https://arxiv.org/abs/2609.24974}{2609.24974} & Harness-Zero: Harness Distillation via Agent-as-Harness & T \\
\bottomrule
\end{tabular}
\caption{Papers included in the literature survey (4/4). Dates indicate first submissions (UTC). P: privileged-context supervision; O: other on-policy distillation; T: trajectory distillation.}
\label{tab:arxiv-inventory-4}
\end{table*}